\documentclass{article}
\usepackage{amsmath, amssymb, amscd, amsthm, amsfonts}
\usepackage{graphicx}
\usepackage{hyperref}
\usepackage[a4paper, total={6.5in, 10in}]{geometry} 
\usepackage[utf8]{inputenc} 
\usepackage[T1]{fontenc}    
\usepackage{hyperref}       
\usepackage{url}            
\usepackage{booktabs}       
\usepackage{amsfonts}       
\usepackage{nicefrac}       
\usepackage{microtype}      
\usepackage[dvipsnames]{xcolor}
\usepackage{graphicx,float} 
\usepackage{subfig}
\usepackage{amssymb}
\usepackage{amsmath,bm}
\usepackage{xcolor}
\usepackage{makecell}
\usepackage{comment}
\usepackage{soul}
\usepackage{pifont}
\usepackage{diagbox}
\usepackage{enumitem}
\usepackage[ruled, lined, linesnumbered, commentsnumbered, longend]{algorithm2e}
\usepackage{tikz}
\usepackage{ifthen}
\usepackage{soul}

\newtheorem{lemma}{Lemma}
\newtheorem{proposition}{Proposition}
\newtheorem{theorem}{Theorem}
\theoremstyle{definition}
\newtheorem{definition}{Definition}
\graphicspath{ {./arxiv_figures/} }

\newcommand{\pr}{\mathbb{P}}

\SetCommentSty{mycommfont}
\newcommand{\sR}{\mathtt{R}_{\lambda,\varepsilon}}
\newcommand{\sRn}{\mathtt{R}_{\lambda,\varepsilon}^{m+n}}

\newcommand{\R}{\mathtt{R}_{\lambda}}
\newcommand{\Rn}{\mathtt{R}^{m+n}}

\newcommand{\sRE}{\mathtt{sRE}_{\lambda,\varepsilon}}
\newcommand{\sREn}{\mathtt{sRE}_{\lambda, \varepsilon}^{m,n}}
\newcommand{\sre}{\mathtt{sRE}}

\newcommand{\RE}{\mathtt{RE}_{\lambda}}
\newcommand{\REn}{\mathtt{RE}_{m,n}}

\newcommand{\srmmd}{\mathtt{sRMMD}}
\newcommand{\sRMMD}{\mathtt{sRMMD}_{\lambda,\varepsilon, k}}
\newcommand{\sRMMDn}{\mathtt{sRMMD}_{\lambda,\varepsilon, k}^{m,n}}

\newcommand{\feps}{f_\varepsilon}
\newcommand{\geps}{g_\varepsilon}
\newcommand{\Teps}{T_\varepsilon}

\DeclareMathOperator*{\argmin}{arg\,min}

\newcommand{\norm}[1]{\left\lVert#1\right\rVert}

\usetikzlibrary{decorations.pathreplacing, calligraphy}
\usetikzlibrary{shapes,arrows}
\tikzset{%
  every neuron/.style={
    circle,
    draw,
    minimum size=0.5cm
  },
  neuron missing/.style={
    draw=none, 
    scale=1,
    text height=0.333cm,
    execute at begin node=\color{black}$\vdots$
  },
}
\tikzstyle{block} = [draw, fill=blue!20, rectangle, 
    minimum height=3em, minimum width=6em]
\tikzstyle{block1} = [draw,  ellipse, 
    minimum height=3em, minimum width=6em]
\tikzstyle{sum} = [draw, ellipse, node distance=1cm]
\tikzstyle{input} = [coordinate]
\tikzstyle{output} = [coordinate]
\tikzstyle{pinstyle} = [draw,  ellipse,, pin edge={to-,thin,black}]
\newif\ifcomments
 \commentstrue
\makeatletter
\title{Multivariate rank via entropic optimal transport: \\sample efficiency and generative modeling}

\author{Shoaib Bin Masud$^{\star}$, Matthew Werenski$^{\dagger}$, James M. Murphy$^{\ddagger}$, Shuchin Aeron$^{\star}$%
  \thanks{The first two authors contributed equally. \newline Emails and Affiliations: \newline $^{\star}$ Department of ECE, Tufts University. Emails:  shoaib\_bin.masud@tufts.edu, shuchin@ece.tufts.edu \newline  $^{\dagger}$ Department of Computer Science, Tufts University. Email: matthew.werenski@tufts.edu \newline $^{\ddagger}$ Department of Mathematics, Tufts University. Email: jm.murphy@tufts.edu } 
  } 

\newif\ifcomments
\commentstrue

\ifcomments
    \usepackage{ulem}
    \def\mwedit#1{{$\!$\color{magenta} [MW: #1]}}
    
\else 
    \def\mwedit#1{}
    
\fi

\ifcomments
    \usepackage{ulem}
    \def\smedit#1{{$\!$\color{orange} [SM: #1]}}
    
\else 
    \def\smedit#1{}
    
\fi

\ifcomments
    \usepackage{ulem}
    \def\saedit#1{{$\!$\color{blue} [SA: #1]}}
    
\else 
    \def\saedit#1{}
    
\fi

\ifcomments
    \usepackage{ulem}
    \def\jmedit#1{{$\!$\color{brown} [JM: #1]}}
    
\else 
    \def\jmedit#1{}
    
\fi

\newtheorem*{remark}{Remark}

\begin{document}
\maketitle
\begin{abstract}

The framework of optimal transport has been leveraged to extend the notion of rank to the multivariate setting while preserving desirable properties of the resulting goodness-of-fit (GoF) statistics. In particular, the rank energy (RE) and rank maximum mean discrepancy (RMMD) are distribution-free under the null, exhibit high power in statistical testing, and are robust to outliers. In this paper, we point to and alleviate some of the practical shortcomings of these proposed GoF statistics, namely their high computational cost, high statistical sample complexity, and lack of differentiability with respect to the data. We show that all these practically important issues are addressed by considering entropy-regularized optimal transport maps in place of the rank map, which we refer to as the \emph{soft rank}. We consequently propose two new statistics, the \emph{soft rank energy (sRE)} and \emph{soft rank maximum mean discrepancy (sRMMD)}, which exhibit several desirable properties. Given $n$ sample data points, we provide non-asymptotic convergence rates for the sample estimate of the entropic transport map to its population version that are essentially of the order $n^{-1/2}$ when the starting measure is subgaussian and the target measure has compact support. This result is novel compared to existing results which achieve a rate of $n^{-1}$ but crucially rely on both measures having compact support. We leverage this result to demonstrate fast convergence of sample sRE and sRMMD to their population version making them useful for high-dimensional GoF testing. Our statistics are differentiable and amenable to popular machine learning frameworks that rely on gradient methods. We leverage these properties towards showcasing the utility of the proposed statistics for generative modeling on two important problems: image generation and generating valid knockoffs for controlled feature selection.

\end{abstract}
\section{Introduction}
It is well-known that in one dimension, the notions of rank and quantile with respect to the distribution of the data are naturally defined via the cumulative distribution function (cdf) and its generalized inverse, respectively. This is because $\mathbb{R}$ has a canonical ordering, which is naturally captured by the cdf.  Based on these notions, a number of goodness-of-fit (GoF) statistics and tests have been proposed in the literature, such as the Kolmogorov-Smirnov test \cite{smirnov1939estimation}, Wilcoxon rank test \cite{wilcoxon1947probability}, and more recently rank-quantile tests \cite{ramdas2017wasserstein}. These rank- and quantile-based statistics are computationally feasible, non-parametric, and are distribution-free under the null, making them desirable in a number of applications, e.g. unsupervised change point detection \cite{OJSP_2020_Cheng}, tests of independence \cite{gretton2007kernel, heller2013consistent}, and hierarchical clustering \cite{szekely2005hierarchical}.

Recently, meaningful multivariate extensions of the notion of rank and quantile maps were proposed in the pioneering works \cite{hallin2017distribution, hallin2021distribution,chernozhukov2017monge}, and more recently in \cite{deb2021multivariate} based on the theory of optimal transportation \cite{villani2009optimal, santambrogio2015optimal}. For a detailed discussion, we refer the reader to a recent survey on the topic \cite{hallin2021measure} and references therein. Essentially, these ideas leverage the convex geometry of the optimal transport (OT) problem with the squared Euclidean metric as the ground cost, where under some mild conditions the optimal maps are gradients of strictly convex functions \cite{mccann1995existence,brenier1991polar}. These extensions and the corresponding high-dimensional analogues of the rank-based GoF statistics based on these extensions retain some of the useful properties of their one dimensional counterparts, namely they are computationally feasible for small sample sizes and are distribution-free under the null.

In this paper we focus on the multivariate rank-based GoF statistic proposed in \cite{deb2021multivariate}, namely the rank energy and rank maximum mean discrepancy (MMD), based on a particular choice of the reference measure when defining the multivariate ranks via optimal transport maps. These statistics are shown to be distribution-free in finite samples (under the null), consistent against alternatives, exhibit high power in statistical testing for heavy-tailed distributions, and are robust to outliers.

However, as we discuss in detail in Section \ref{subsec:IssuesWithRE}, practical use of the rank energy and rank MMD suffer from the well-known curse of dimensionality associated with the estimation of OT maps as well as high computational costs for large sample sizes. Furthermore, rank energy and rank MMD suffer from a lack of differentiabilty in the data, limiting a direct use of iterative gradient-based optimization methods.  This inhibits their use for learning a generative model by using these GoF statistics as a loss function, as has been successfully done with MMD and Wasserstein distances \cite{li2017mmd, arjovsky2017wasserstein}.

In this context, our paper makes the following main contributions.
\begin{itemize}
    \item[\textbf{(C1)}] We introduce the multivariate notion of the \textit{soft rank} that exploits the recent developments in computational optimal transport, namely entropic regularization \cite{peyre2019computational}.  We then propose the \textit{soft rank energy (sRE)}, a new multivariate GoF statistic, and the related \emph{soft rank maximum mean discrepancy (sRMMD)}.  In Proposition \ref{prop:properties} and Proposition \ref{prop:sRE_eps_convergence}, we establish the  properties of soft rank energy and its convergence to rank energy, which together suggest its use for measuring GoF in two-sample testing. 
    
    \item[\textbf{(C2)}]  We provide a novel result (Theorem \ref{thm:subg_conv_rate}) on the convergence rate of a sample-driven estimate for general entropic maps, from a subgaussian measure to a measure with a compact support, that enjoys a convergence rate of $n^{-1/2}$ to the population entropic map, even in high dimensions. We note that the subgaussian assumption on the starting measure is a significant weakening of the assumptions compared to a recent result of \cite{rigollet2022sample} that assumes compactness of both measures albeit providing a faster rate of $n^{-1}$. In this context we also note that our method of proof is different from \cite{rigollet2022sample}, which relies on strong concavity furnished by the compactness assumption on both measures. Specifically, we adapt the methodology developed in \cite{pooladian2021entropic} and use empirical process theory arguments to establish our result under the weakened assumptions.
    
\item [\textbf{(C3)}] Since the target measure in defining multivariate ranks has a compact support, we utilize the previous result in Theorems \ref{thm:sre_convergence_rate} and \ref{thm:srmmd_convergence_rate} to establish the statistical convergence of sample sRE and sample sRMMD, respectively, to their population versions with rate $n^{-1/2}$. This result is important when using these quantities are used for testing goodness-of-fit between two subgaussian measures. 

    \item[\textbf{(C4)}]  We show the practical advantages of our proposed methods on several real generative modeling problems.  First, we use sRE and sRMMD as the loss functions in a simple generative model architecture to produce MNIST-digits. Under an appropriate choice of the entropic regularization parameter, we show that using sRE and sRMMD  as the loss functions can generate all of the digits successfully, and does not suffer from mode collapse.  We then utilize the sRMMD in a deep generative model in order to produce valid knockoffs \cite{barber2015controlling}. We showcase improved tradeoffs between detection power versus false discovery rate (FDR) compared to other benchmarks of knockoff generation techniques on different Gaussian and non-Gaussian distributional settings.  We also test our approach for provable biomarker selection in metabolomics.\footnote{Codes to reproduce the results are available at \url{https://github.com/ShoaibBinMasud/soft-rank-energy-and-applications}} 
\end{itemize}
\paragraph{Paper outline:}  The paper is organized as follows. In Section \ref{sec:background}, we provide the required background on optimal transport theory and it's entropy-regularized variant as well as discuss the multivariate rank energy. In Section \ref{sec:MainDefinitions} we introduce the soft rank energy (sRE) and the soft rank maximum mean discrepancy (sRMMD), and sample versions thereof.  In Section \ref{sec:properties} we state our main theorems which establish the favorable properties of the soft rank energy. We also prove finite sample convergence rates for the sample sRE and sRMMD to their population versions.  In Section \ref{sec:applications}, we provide extensive simulations to establish soft rank energy as a GoF measure and include two applications where we use it as the loss function of a generative model.

\paragraph{Notation:}

We will let $X,Y$ denote random vectors in $\mathbb{R}^d$ and we will use superscripts to denote their entries $X=(X^1,\cdots, X^d)$. A bold $\mathbf{X}$ will denote a matrix. $\|\cdot\|$ will denote the Euclidean norm in $\mathbb R^d$. $\overset{d}{=}$ will denote equality in distribution.

Throughout the paper we consider the space of measures in two settings. First, for a bounded set $\Omega \subset \mathbb{R}^d$ we  let $\mathcal{P}(\Omega)$ denote the set of probability measures supported on $\Omega$. Second, we consider $\sigma^2$-subgaussian measures.  More precisely, a measure $P$ is said to be $\sigma^2$-subgaussian if the random  vector $X$ with law $P$ satisfies $\mathbb{E}\exp(\norm{X}^2/2d\sigma^2) \leq 2$.

We write $a\lesssim b$ if there exists a constant $C$ such that $a\leq Cb$. The rest of the notations is standard and clear from the context. We also include in Table \ref{tab:notation} a list of notations introduced later for easy reference.
\begin{table}[ht]
\centering
\begin{tabular}{|c|c|}
\hline
 Symbol & Meaning  \\ \hline \hline
 $B_2^d(0,r)$ & Euclidean ball of radius $r$ in $\mathbb{R}^d$ \\ \hline 
 $P_\lambda = \lambda P_X + (1-\lambda)P_Y$ & Mixture distribution, $\lambda \in (0,1)$. \\ \hline 
 $T$ & Optimal transport map \\ \hline
 $\Teps$ & Entropic map \\ \hline
 $\Teps^{n,n}$ & Two-sample entropic map \\ \hline
 $\R$ & Rank map. $T$ from $P_\lambda$ to $\text{Unif}([0,1]^d)$ \\ \hline
 $\sR$ & Soft rank map. $\Teps$ from $P_\lambda$ to $\text{Unif}([0,1]^d)$ \\ \hline
 $\sRn$ & Sample soft rank map \\ \hline
 $\RE$ & Rank energy \\ \hline
 $\sRE$ & Soft rank energy \\ \hline
 $\sREn$ & Sample soft rank energy \\ \hline 
 $\sRMMD$ & Soft rank MMD \\ \hline 
 $\sRMMDn$ & Sample soft rank MMD \\ \hline
\end{tabular}
\caption{Frequently used notation throughout the paper.} \label{tab:notation}
\end{table}

\section{Background on Optimal Transport and Rank Energy} \label{sec:background}

Throughout this work we consider measures contained in $\mathcal P_{ac}(\Omega)$, the space of absolutely continuous measure on $\Omega \subset \mathbb{R}^d$  i.e. those characterized by a density function with respect to the Lebesgue measure. We will assume throughout that $\Omega$ is bounded and contained in the ball $B_2^d(0,r)$.
\subsection{Optimal Transport}
Given two distributions $P, Q\in \mathcal P(\Omega)$, where $\mathcal P(\Omega)$ denotes the space of Borel probability measures on $\Omega$, the \emph{Monge problem} \cite{monge1781memoire} seeks a measurable map $T: \Omega \rightarrow \Omega$ that pushes $P$ to $Q$ with a minimal cost.  Precisely, it solves
\begin{equation}\label{eq:monge}
     \inf_{T}\int \frac{1}{2}\|x- T(x)\|^2 dP(x),\;\; \text{subject to}\;\;  T_\#P= Q,
\end{equation}
where $T_\#P$ denotes the \textit{push-forward measure}, which satisfies $(T_\#P)[A] = P[T^{-1}(A)]$ for all measurable sets $A$.  

Throughout we will make heavy use of the optimal map $T$ which minimizes (\ref{eq:monge}). It is therefore important to establish the existence, uniqueness, and important properties of $T$, which are established by the following celebrated theorem.
\begin{theorem}[Brenier-McCann \cite{mccann1995existence, brenier1991polar}]\label{th:mccann}
    Let $P\in \mathcal P_{ac}(\Omega)$ and $Q\in \mathcal P(\Omega)$. Then there exists a convex function $\phi:\Omega\rightarrow \mathbb R$ whose gradient $\nabla \phi:\Omega\rightarrow \Omega$ pushes $P$ forward to $Q$.  Moreover, if $P$ and $Q$ have finite second moments, then $\nabla \phi$ is the unique (up to sets of measure 0) solution to the Monge problem \eqref{eq:monge}.
\end{theorem}
\subsubsection{Entropic Optimal Transport}
Towards developing the notion of soft rank energy we first state a relaxation of the \emph{Monge problem}, where instead of a map, one seeks an optimal ``coupling'' $\pi$ between a source distribution $P$ and a target distribution $Q$. The \emph{Kantorovich relaxation} \cite{OTAM} solves 
\begin{align}\label{eq:kantoro_OT}
    \min_{\pi\in \Pi(P, Q)}\int \frac{1}{2}\| x -  y \|^2 d\pi( x, y),
\end{align}
where $\Pi(P, Q)$ is the set of joint probability measure on $\mathcal P (\Omega \otimes \Omega)$ with marginals $P$ and $Q$. When a solution to the Monge problem \ref{eq:monge} exists, then the solution to Kantorovich relaxation coincides with it in the sense that the optimal plan is concentrated on $\{(x, T(x)) : x\in \text{supp}(P)\}$ \cite{OTAM}. 

The statistic we propose relies on an \textit{entropy-regularized} version of (\ref{eq:kantoro_OT}). For $\varepsilon>0$, the primal formulation of the entropy-regularized optimal transport is given by
\begin{equation}\label{eq:entropicOT}
    \min_{ \pi\in \Pi(P, Q)} \int \frac{1}{2}\|x-y\|^2 \text d\pi( x,  y) + \varepsilon \text{KL}( \pi \  || \ P \otimes Q),
\end{equation}
where
\begin{equation*}
    \text{KL}(\pi|P \otimes Q) \triangleq \int \ln \left ( \frac{d\pi( x, y)}{ dP(x) dQ( y)} \right ) d\pi( x, y).
\end{equation*}
This problem has been extensively studied both for its theoretical properties, as well as for the efficient algorithms that are used to solve it (See \cite{cuturi2013sinkhorn,peyre2019computational,genevay2016stochastic} and references therein).
Importantly, (\ref{eq:entropicOT}) admits the following dual formulation that follows from the Fenchel-Rockafellar duality theorem \cite{clason2021entropic}, a derivation of which may be found in \cite{genevay2019entropy, peyre2019computational}
\begin{align}\label{eq:dual}
    \max_{f, g} \int f(x)dP(x) + \int  g(y)dQ(y) -\varepsilon \int \int \exp\left [ \frac{1}{\varepsilon}\left ( f(x)+g(y)-(1/2)\|x-y\|^2 \right ) \right ] dP(x)dQ(y) + \varepsilon,
\end{align}
where the maximization is over the pairs $f \in L^1(P), g \in L^1(Q)$ of functions over $\Omega$.  The optimal entropic potentials for $\varepsilon$ are the pair of functions $(\feps, \geps)$ which achieve the maximum in  \eqref{eq:dual}. Furthermore, there is an optimality relation between (\ref{eq:entropicOT}) and (\ref{eq:dual}) given by
\begin{equation} \label{eq:opt_relation}
    d\pi_\varepsilon(x, y) = \exp\left [ \frac{1}{\varepsilon}\left ( \feps(x)+\geps(y)-\frac{1}{2}\|x-y\|^2 \right ) \right ] dP(x)dQ(y),
\end{equation}
where $\pi_\varepsilon$ denotes the solution to \eqref{eq:entropicOT}. We emphasize the fact that $\pi_\varepsilon$ is not a map, but a diffused coupling, and that the degree of diffusion depends on the entropic regularization parameter $\varepsilon$ \cite{peyre2019computational}.  In the finite sample setting, entropic regularized optimal transport significantly reduces computational complexity \cite{cuturi2013sinkhorn} and also yields a differentiable loss function \cite{schmitz2018wasserstein}.
\subsection{Rank and Quantile Maps}
Let $P\in \mathcal{P}(\Omega)$ for $\Omega \subset \mathbb{R}^d$ and consider a random variable $X \sim P$.  When $d=1$, the rank map, or cumulative distribution function, is $F_X(t) = \mathbb{P}\{X \leq t\} $, and the quantile map is its generalized inverse, $F_X^{-1}(p) =\inf \{t\in \mathbb R: p\leq F_X(t)\}$. The rank and quantile maps are always monotonic increasing and are continuous when $P$ has a density. When $F_X$ is continuous one can show that the random variable $F_X(X)$ is distributed according to $\text{Unif}([0, 1])$. Similarly, when the quantile map is continuous $F_X^{-1}(U)$ is distributed according to $X$ where $ U \sim \text{Unif}([0, 1])$.

The key insight in using the theory of optimal transport to define multivariate rank and quantile maps comes from noticing that in one dimension, the optimal map betwwen two measures, $P$ and $Q$, is given by $T = F_Q^{-1} \circ F_P$, where $F_P$ is the rank map of $P$ and $F_Q^{-1}$ is the quantile map of $Q$ \cite{hallin2017distribution, hallin2021distribution,chernozhukov2017monge, deb2021multivariate}. When $Q = \text{Unif}([0,1])$, one has $F_Q^{-1} = \text{Id}$ and therefore $F_Q^{-1} \circ F_P = \text{Id} \circ F_P = F_P$. By the push-forward constraint we know that $F_P\#P = \text{Unif}([0,1])$ which is just one way of observing that $F_P(X) \sim \text{Unif}([0,1])$. The main intuition is that the rank map is exactly the optimal map from $P$ to $\text{Unif}([0,1])$. Analogously, with Theorem \ref{th:mccann} ensuring the existence of a unique optimal $T$ that is monotone (being a gradient of a convex function), \cite{deb2021multivariate} generalizes the notion of rank and quantile maps to dimensions $d\ge 2$ as optimal transport maps to and from $\text{Unif}([0,1]^d)$.  
\begin{definition}[Definition 2.1 \cite{deb2021multivariate}] \label{def:population_rank}
    Let $P\in \mathcal{P}_{ac}(\Omega)$ and let $Q = \text{Unif}([0,1]^d)$. The multivariate rank and quantile maps for $P$ are defined as $\mathtt{R} = \nabla \phi$ and $\mathtt{Q} = \nabla \phi^*$, respectively, where $\phi$ is the \emph{strictly} convex function as in Theorem \ref{th:mccann} such that $\nabla \phi$ optimally transports $P$ to $Q$. Here $\phi^*$ denotes the standard convex conjugate\footnote{for any proper function $f:\Omega\rightarrow R$, the convex conjugate $f^*:\Omega\rightarrow \mathbb R$ is defined as: $f^*(y)\triangleq \sup_{y} \langle x, y\rangle- f(y)$ for all $y\in \Omega$.} of the convex function $\phi$. 
\end{definition}
With the high-dimensional analogue of the rank defined, we can now state the candidate GoF statistics proposed in \cite{deb2021multivariate}.
\begin{definition}[Definition 3.2 \cite{deb2021multivariate}]\label{def:re}
    Let $P_X, P_Y\sim \mathcal{P}_{ac}(\Omega)$ and let $X,X' \sim P_X$ and $Y,Y'\sim P_Y$ be independent. Let $P_\lambda = \lambda P_X +(1-\lambda)P_Y$ denote the mixture distribution for any $\lambda \in (0,1)$ and let $\R$ be the multivariate rank map of $P_\lambda$ as in Definition \ref{def:population_rank}. The \textit{(population) rank energy} is defined
    \begin{align}\label{eq:population_rank}
        \RE(P_X, P_Y)^2 &\triangleq C_d \int_{\mathcal S^{d-1}}\int_{\mathbb R} \left (\mathbb P\big(a^\top \R(X)\leq t\big) - \mathbb P\big(a^\top \R(Y)\leq t\big)\right )^2 dt  d\kappa(a) \notag \\
        &= 2\mathbb{E}||\R(X) - \R(Y)|| - \mathbb{E}||\R(X) - \R(X')|| - \mathbb{E}||\R(Y) - \R(Y')||,
    \end{align}
    where $\mathcal{S}^{d-1} \triangleq \{x\in \mathbb R^d: \|x\| = 1\}$ is the unit sphere in $\mathbb{R}^d$, $\kappa(\cdot)$ is the uniform measure on $\mathcal S^{d-1}$, and $C_d = \left (2\Gamma(d/2) \right )^{-1} \sqrt{\pi}(d-1)\Gamma\big((d-1)/2\big)$ is an appropriate normalizing constant.  
\end{definition}
Note that $\RE^2$ closely resembles the definition of energy distance \cite{szekely2013energy},
\begin{equation*}
    \text{En}(P_X,P_Y)^2 \triangleq  C_d \int_{\mathcal S^{d-1}}\int_{\mathbb R} \left (\mathbb P\big(a^\top X\leq t\big) - \mathbb P\big(a^\top Y\leq t\big) \right)^2 dt  d\kappa(a)
\end{equation*}
a widely used GoF statistic in two-sample testing. This definition is motivated by the continuity and uniqueness of characteristic functions. Namely, $P_X = P_Y$ if and only if $a^\top X \overset{d}{=} a^\top Y$ (that is, equality in distribution) for $\kappa$-almost everywhere $a$; indeed, the integration over the sphere aggregates the discrepancy in characteristic functions in every direction. For a discussion showing the equivalence of the formulation in \eqref{eq:population_rank} using integrals over $\mathcal{S}^{d-1}$ and the formulation in terms of expectations of $\R$, see \cite{baringhaus2004new}.  One advantage of  $\RE^2$ over the energy distance is that $\RE^2$ is distribution-free under the null for all sample sizes \cite{deb2021multivariate}.

One can generalize the rank energy by replacing the pairwise distance with a kernel function \cite{phillips2011gentle}. Given $k:\mathbb{R}^d \times \mathbb{R}^d \rightarrow \mathbb{R}$ a characteristic kernel \footnote{a kernel $k$ is said to be characteristic if the map $P\rightarrow \int_{\mathcal X} k(\cdot, x)d P(x)$ is injective, where $P$ is a measure defined on the topological space $\mathcal X$.}, we can interpret this distance as \textit{rank maximum mean discrepancy},
\begin{align}\label{eq:rank_mmd}
        \mathtt{RMMD}_\lambda(P_X,P_Y)^2 \triangleq  \mathbb{E} [k(\R(X),\R(X')] + \mathbb{E} [k(\R(Y),\R(Y')]- 2\mathbb{E}[k(\R(X),\R(Y)].
\end{align}
Note that $\mathtt{RMMD}_\lambda(P_X,P_Y)^2$  closely follows the definition of maximum mean discrepancy (MMD) \cite{gretton2012kernel}, which is a widely used statistic in the framework of two-sample testing: $\mathtt{MMD}(P_X,P_Y)^2 \triangleq  \mathbb{E} [k(X,X')] + \mathbb{E} [k(Y,Y')]- 2\mathbb{E}[k(X,Y)].$

One can simply view $\RE^2$ and $\mathtt{RMMD}_\lambda^2$ as the ``rank-transformed'' energy distance and maximum mean discrepancy, respectively. 

In practice one can rarely evaluate the $\RE^2$ and $\mathtt{RMMD}_\lambda^2$ directly since the exact multivariate rank $\R$ often unavailable. Instead the statistic must be estimated from samples, a procedure outlined in the next subsection. 

\subsubsection{Sample Rank Map and Rank Energy}\label{subsubsec:sample_rank}
Given i.i.d. samples  $X_1, \dots,  X_m \sim P$, the empirical measure is defined as $P^m = \frac{1}{m}\sum_{i=1}^m \delta_{X_i}$ where $\delta_{X_i}$ is a Dirac distribution placed at $X_i$. Given two empirical measures $P^m = \frac{1}{m}\sum_{i=1}^m \delta_{X_i}, Q^m = \frac{1}{m}\sum_{j=1}^m \delta_{Y_j}$, where $Y_1, \cdots, Y_m\sim Q$, the plug-in estimate of the transport map $T^m$ between $P$ and $Qf$ is obtained by solving (\ref{eq:monge}) between $P^m, Q^m$ 
\begin{equation} \label{eq:OT_comb}
    T^m \triangleq \argmin_{T: T_\#P^m = Q^m} \int \frac{1}{2}||x - T(x)||^2 dP^m(x) = \argmin_{T: T_\#P^m = Q^m} \frac{1}{m} \sum_{i=1}^m \frac{1}{2} ||X_i - T(X_i)||^2.
\end{equation}
This problem can be converted to a standard linear program and solved either by tailored methods or general linear program solvers \cite{peyre2019computational}. To obtain the sample rank map for a measure $P$, one performs the procedure above with $Q = \text{Unif}([0,1]^d)$ and takes $\mathtt{R}^m$ to be the estimate $T^m$.  In practice, one may generate the samples from $\text{Unif}([0,1]^d)$ using a pseudo-random sequence of points. In \cite{deb2021multivariate}, Halton sequences \cite{hofer2009distribution} are used and we do the same in our experiments. Nevertheless, any sequence which weakly converges to $\text{Unif}([0,1]^d)$ can be used. 

With the sample rank map defined, one can immediately extend the definition of the rank energy. Given two sets of samples $X_1, \dots, X_m \sim P_X$ and $Y_1, \dots, Y_n \sim P_Y$, define $P^{m+n}$ as
\begin{equation*}
    P^{m+n} = \frac{1}{m+n}\left ( \sum_{i=1}^m \delta_{X_i} + \sum_{j=1}^n \delta_{Y_j} \right ),
\end{equation*} 
the empirical mixture of the two sets of samples. Next, let $Q^{n+m} =  \frac{1}{m+n}\sum_{i=1}^{n+m} \delta_{U_i}$ where $U_i \sim \text{Unif}([0,1]^d)$. Using these measures, one obtains $\Rn$ and then computes the sample rank energy as 
\begin{align} 
    \REn(P_X,P_Y)^2 \triangleq \frac{2}{mn}\sum_{i=1}^{m}\sum_{j=1}^n& \|\Rn(X_i)- \Rn(Y_j)\|- \frac{1}{m^2}\sum_{i,j=1}^{m}\|\Rn(X_i)- \Rn(X_j)\| \nonumber \\
    &- \frac{1}{n^2}\sum_{i,j=1}^{n}\|\Rn(Y_i)- \Rn(Y_j)\|. \label{eq:sre_mn}
\end{align}
The authors in \cite{deb2021multivariate} also define the sample version of rank maximum mean discrepancy as
\begin{align} 
        \mathtt{RMMD}_{m,n}(P_X,P_Y)^2 &\triangleq  \frac{1}{m^2} \sum_{i,j=1}^m k(\Rn(X_i),\Rn(X_j)) + \frac{1}{n^2} \sum_{i,j=1}^n k(\Rn(Y_i),\Rn(Y_j)) \nonumber \\
        & \hspace{3cm} - \frac{2}{mn} \sum_{i=1}^m\sum_{j=1}^n k(\Rn(X_i),\Rn(Y_j)), \label{eq:sample_rmmd}
\end{align}
for some characteristic kernel $k$.
Both $\REn^2$ and $\mathtt{RMMD}_{m,n}^2$ are distribution-free under the null for any fixed sample size \cite{deb2021multivariate}. 
\subsubsection{Practical Issues with Rank Energy}\label{subsec:IssuesWithRE}
While the rank energy enjoys certain desirable properties, it also suffers from important drawbacks.  We focus on two below.
\paragraph{Complexity, sample and computational:} In practice, to compute the test statistic proposed in \cite{deb2021multivariate}, one needs to solve the discrete version of the Monge problem. Given $n$ samples, solving this problem exactly using typical methods requires $O(n^3 \log n)$ computations \cite{peyre2019computational}, and the standard procedure is not amenable to parallelization. To make matters worse, when the samples are in $\mathbb{R}^d$, (\ref{eq:OT_comb})
has a large sample complexity in the sense that when $\lambda = m/(m+n)$ and in the absence of further assumptions, one only has
\begin{equation*}
    \frac{1}{m+n}\mathbb{E}\left [ \sum_{i=1}^m \|\Rn(X_i) - \R(X_i) \| + \sum_{j=1}^n \|\Rn(Y_j) - \R(Y_j) \| \right ] \lesssim (m+n)^{-1/d}.
\end{equation*}
Unfortunately, this rate which depends exponentially on $d$ can be tight \cite{dudley1969speed}, implying the need for extremely large sample sizes when working in high dimension to get a faithful estimate of the map $\R$ which is required to estimate $\RE(P_X, P_Y)^2$. Together, these observations mean the approach of \cite{deb2021multivariate} exhibits a statistical-computational bottleneck that precludes its use in high dimensions: one needs $n$ large to get a good estimate, but computing the estimate for large $n$ is computationally infeasible.
\paragraph{Gradient issues:} Another reason that one may be interested in obtaining the rank statistic is to use it as a loss function for generative modelling. For example, if $X_1,...,X_m$ are real samples and $Y_1,...,Y_n$ are from a standard model (e.g., $Y_i \sim N(0,I_d)$) one may seek to solve\footnote{Technically this should be written $\REn(P_X, (T_\theta)_\#P_Y)$, but for clarity we avoid this notation here.}
\begin{equation} \label{eq:re_opt_template}
    \min_{\theta} \REn(\{X_i\}_{i=1}^m, \{T_\theta(Y_j)\}_{j=1}^n)^2
\end{equation}
where $\REn^2$ is the (squared) sample rank energy (defined in (\ref{eq:sre_mn})) and $T_\theta:\Omega \rightarrow \Omega$ is a learnable transformation parameterized by $\theta$. In this setting, a small statistic indicates that $T_\theta$ is successfully transforming the samples $\{Y_i\}_{i=1}^n$ in such a way that they are difficult to distinguish from the $\{X_i\}_{i=1}^m$. To obtain $\theta^*$, a (near) minimizer of (\ref{eq:re_opt_template}), many standard procedures  start at a random $\theta_0$ and then use a gradient based procedure which requires access to $\nabla_\theta \REn(\{X_i\}_{i=1}^m,  \{T_\theta(Y_i)\}_{i=1}^n)^2$ to obtain a sequence of improving $\theta_i$ (for example, $\theta_{i+1} = \theta_i - \eta \nabla_{\theta_i} \REn(\{X_i\}_{i=1}^m,  \{T_\theta(Y_j)\}_{j=1}^n)^2$ for some learning rate $\eta > 0$) \cite{bottou2012stochastic, boyd2004convex}.

This gradient descent approach will not work for the rank energy in typical settings for two reasons. First, one needs to rely on special methods to back-propagate through the construction of the rank map which is the argmin of a convex optimization problem (one is forced to rely on so-called convex optimization layers \cite{agrawal2019differentiable}, for example). Second, when $n = m$, even if the gradient is computed, it will typically be exactly zero. This is because the rank energy is invariant to sufficiently small perturbations of the samples (see Section \ref{sec:lack_of_grad} for a precise description). In the absence of a non-zero gradient, the methods above will not be able to incrementally improve the setting of $\theta$, or even choose a direction in the parameter space to search along. In practice the first-order methods above are by far the most popular and are often the only feasible methods when $\theta$ is extremely high-dimensional (i.e. represents the weights of a deep neural network). The absence of a proper gradient severely restricts the utility of the rank energy in broader contexts. 

These two drawbacks are not isolated to the rank energy, but are present whenever an empirical rank map is involved. This  also limits the the utility of the rank maximum mean discrepancy. In contrast, many other GoF measures in statistics (e.g., maximum mean discrepancy (MMD) \cite{li2015generative}, energy distance \cite{bellemare2017cramer}), can be used as a loss function in a generative model to learn a new distribution as close as possible to a target distribution. Motivated by the use of a rank-based GoF statistic towards generative modeling, we propose \textit{soft rank energy} and \textit{soft rank maximum mean discrepancy}, which largely alleviate the defects of their rank counterparts.
\section{Soft Rank and Soft Rank Energy}\label{sec:MainDefinitions}
Based on the primal-dual relationship in equations \eqref{eq:entropicOT},\eqref{eq:dual}, we note the following definition of the entropic map.
\begin{definition}[Entropic map \cite{pooladian2021entropic}]\label{def:entropic_map}
Given an optimal entropic plan $\pi_\varepsilon$ or the optimal entropic potentials $(\feps, \geps)$ between $P\in \mathcal{P}_{ac}(\Omega)$ and $Q \in \mathcal{P}_{ac}(\Omega)$, the \emph{entropic map} is defined by
\begin{equation}\label{eq:entropic_map}
    \Teps(x)  \triangleq \displaystyle\int\pi_\varepsilon(y|x) = \mathbb E_{\pi_\varepsilon}[Y|X=x] = \frac{\displaystyle\int y \exp\left(\frac{1}{\varepsilon}\left (\geps(y) - \frac{1}{2}\|x-y\|^2  \right) \right ) dQ(y)}{\displaystyle\int \exp{\left(\frac{1}{\varepsilon}\left (\geps(y) - \frac{1}{2}\|x-y\|^2\right ) \right ) } dQ(y)}
\end{equation}
where $\pi_\varepsilon(y|x)$ denotes the conditional distribution of $\pi_\varepsilon$.
\end{definition}
In \cite{pooladian2021entropic}, the latter form is used to define a sample version of the entropic map. Given samples $X_1,...,X_n \sim P$ and $Y_1,...,Y_n \sim Q$ and optimal entropic potentials $(\feps^n, \geps^n)$ solving (\ref{eq:dual}) between $P^n, Q^n$, the sample entropic map is defined as
\begin{equation}\label{eq:entropic_map_sample}
    \Teps^{n,n}(x)  \triangleq \frac{\displaystyle\sum_{i=1}^n Y_i \exp\left(\frac{1}{\varepsilon}\left (\geps^n(Y_i) - \frac{1}{2}\|x - Y_i\|^2 \right )\right)}{\displaystyle\sum_{i=1}^n \exp\left(\frac{1}{\varepsilon}\left (\geps^n(Y_i) - \frac{1}{2}\|x - Y_i\|^2 \right )\right)}.
\end{equation}

We adopt these definitions in constructing the soft rank maps.
\begin{definition}[Soft Rank Map]\label{def:entropic_rank}
    The \emph{soft rank map} $\mathtt{R}_\varepsilon$ for a measure $P$ is the entropic map from $P$ to $Q = \text{Unif}([0,1]^d)$ as in (\ref{eq:entropic_map}). The sample soft rank map $\mathtt{R}_\varepsilon^n$ is defined as the sample entropic map from $P^n$ to $Q^n$ as in (\ref{eq:entropic_map_sample}).
\end{definition}
The name ``soft rank'' exists in the literature as a general term used  for a differentiable approximation to sorting \cite{taylor2008softrank, blondel2020fast} which can be seen as a special case of the soft rank map for one dimensional data. 

With these alternatives to the original rank map, the definition of the soft rank energy is as follows.
\begin{definition}[Soft Rank Energy]\label{def:sre}
Let $P_X,P_Y \in \mathcal{P}_{ac}(\Omega)$ and let $X,X' \overset{i.i.d.}{\sim} P_X, Y,Y' \overset{i.i.d.}{\sim} P_Y$.  Let $P_\lambda = \lambda P_X +(1-\lambda) P_Y$ for  $\lambda\in (0,1)$ and let $\sR$ be the soft rank map of $P_\lambda$. 
\begin{enumerate}
\item[(a)] \textit{soft rank energy} (sRE) is defined as:
    \begin{align}\label{eq:sre}
        \sRE(P_X,P_Y)^2 &\triangleq C_d \int_{\mathcal S^{d-1}}\int_{\mathbb R} \left (\mathbb P\big(a^\top \sR(X)\leq t\big) - \mathbb P\big(a^\top \sR(Y)\leq t) \right )^2 dt  d\kappa(a) \\
        &= 2\mathbb E\big\|\sR(X)-\sR(Y)\big\|- \mathbb E\big\|\sR(X)-\sR(X')\big\| - \mathbb E\big\|\sR(Y)-\sR(Y')\big\|,\nonumber
    \end{align}
    where $C_d$ and $\kappa$ are the same as in Definition \ref{def:re}. 
    \item[(b)] Let $X_1,...,X_m \overset{i.i.d.}{\sim} P_X, Y_1,...,Y_n \overset{i.i.d.}{\sim} P_Y$ and let $\sRn$ be an independently estimated soft rank map using $m+n$ samples from $P_\lambda$. The \emph{sample soft rank energy} is defined as
    \begin{align}
        \sREn(P_X,P_Y)^2 &\triangleq \frac{2}{mn} \sum_{i=1}^m\sum_{j=1}^n \| \sRn(X_i) - \sRn(Y_j)  \| - \frac{1}{m^2} \sum_{i,j=1}^n \| \sRn(X_i) - \sRn(X_j)  \| \nonumber \\
        & \hspace{3cm} - \frac{1}{n^2} \sum_{i,j=1}^n \| \sRn(Y_i) - \sRn(Y_j)  \|.  \label{eq:sample_sRE}
    \end{align}
\end{enumerate}
\end{definition}
The equivalence of the formulation in terms of integrals over $\mathcal{S}^{d-1}$ and the formulation in terms of expectations of $\sR$ is shown in Proposition \ref{prop:properties}.
Comparing the definition of the rank energy (Definition \ref{def:re}) to the soft rank energy, the only change is the use of the soft rank map instead of the rank map. In the subsequent sections we establish a convergence property of entropic maps which immediately implies a convergence property for the soft rank map (as the soft rank map is a special case of an entropic map). From the convergence of the soft rank map several favorable properties of the soft rank energy can be derived which motivate this choice of statistic both practically and theoretically.

\begin{remark} The reason for using separate batches of samples is a technical artifact and is required only because our analysis requires independence between the samples used to compute the estimate of the soft rank energy and those used to estimate the map. Imposing this condition only requires a doubling of the number of samples, and in fact the choice to use $m+n$ samples to estimate the map is somewhat arbitrary, and we use this convention to make the notation and statement of the results more compact. In practice, one may even choose not to use separate batches of samples for map estimation and calculating the statistic at all, and we adopt this strategy in our experiments in Section \ref{sec:applications}.
\end{remark}

Using a similar approach to the one taken in \eqref{eq:rank_mmd}, one can also generalize soft rank energy by using a kernel function $k:\mathbb R^d\times \mathbb R^d\rightarrow \mathbb R$ instead of pairwise Euclidean distances. 

\begin{definition}
    Let $k:\mathbb{R}^d \times \mathbb{R}^d \rightarrow \mathbb{R}$ be a characteristic kernel. Let $P_X$ and $P_Y$ be two probability measures and let $X,X' \overset{i.i.d.}{\sim} P_X, Y,Y' \overset{i.i.d.}{\sim} P_Y$.  Let $\sR$ denote the soft rank map of $P_\lambda$ for $\lambda \in (0,1)$. 
    
    \begin{enumerate}
    \item[(a)] The \textit{soft rank maximum mean discrepancy} (sRMMD) is defined as 
    \begin{align*}
        \sRMMD(P_X,P_Y)^2 &\triangleq  \mathbb{E} [k(\sR(X),\sR(X'))] + \mathbb{E} [k(\sR(Y),\sR(Y'))] \nonumber \\
        & - 2\mathbb{E}[k(\sR(X),\sR(Y))].
    \end{align*}
    \item[(b)] Let $X_1,...,X_m \overset{i.i.d.}{\sim} P_X, Y_1,...,Y_n \overset{i.i.d.}{\sim} P_Y$ and let $\sRn$ be an independently estimated rank map using $m+n$ samples from $P_\lambda$. The \emph{sample soft rank maximum mean discrepancy} is defined as
    \begin{align*} 
        \sRMMDn(P_X,P_Y)^2 &\triangleq  \frac{1}{m^2} \sum_{i,j=1}^m k(\sRn(X_i),\sRn(X_j)) + \frac{1}{n^2} \sum_{i,j=1}^n k(\sRn(Y_i),\sRn(Y_j)) \nonumber \\
        & \hspace{3cm} - \frac{2}{mn} \sum_{i=1}^m\sum_{j=1}^n k(\sRn(X_i),\sRn(Y_j)). 
    \end{align*}
    \end{enumerate}
\end{definition}
\section{Properties of the Soft Rank Energy} \label{sec:properties}

\subsection{Estimation of the Entropic Map}

As a first step for many of our results we give a convergence rate of the sample entropic map to the population entropic map.  In this work, we consider results in two regimes. First, when the measure $P$ is subgaussian and $Q$ is bounded. In this case we have

\begin{theorem} \label{thm:subg_conv_rate}
    Suppose that  $Q \in \mathcal{P}(B_2^d(0,r))$ and that $P$ is $\sigma^2$-subgaussian for some $\sigma^2 \geq \frac{r^2}{2d \log 2}$. Let $\Teps$ be the entropic map from $P$ to $Q$ and let $\Teps^{n,n}$ be the estimated entropic map from $P^n$ to $Q^n$. Then 
    \begin{align*}
        \mathbb{E}||\Teps^{n,n} - \Teps||_{L^2(P)}^2 &\leq b_1(r,d,\sigma^2, \varepsilon) n^{-1/2}
    \end{align*}
    for some function $b_1$ independent of $n$.
\end{theorem}
An exact expression for $b_1$ can be found in (\ref{eq:b1_expression}) in the appendix.  The proof of Theorem \ref{thm:subg_conv_rate} is deferred to Section \ref{sec:cr_proof} in the supplement. The argument builds on tools developed in \cite{pooladian2021entropic} and techniques from empirical process theory. 

An important point of comparison for Theorem \ref{thm:subg_conv_rate} is \cite{rigollet2022sample}, in which the following result is shown when $P,Q$ have bounded supported.
\begin{theorem}[\cite{rigollet2022sample}, Theorem 4, adapted] \label{thm:bounded_conv_rate}
    Let  $P,Q \in \mathcal{P}(B_2^d(0,r))$ let $\Teps$ be the entropic map from $P$ to $Q$ and let $\Teps^{n,n}$ be the estimated entropic map from $P^n$ to $Q^n$. Then 
    \begin{align*}
        \mathbb{E}||\Teps^{n,n} - \Teps||_{L^2(P)}^2 &\leq b_2(r,d,\varepsilon) n^{-1}.
    \end{align*}
    for some function $b_2$ independent of $n$.
\end{theorem}

\begin{remark} 
The analysis leading to this result requires a strong concavity property of the entropic dual problem which is only present when both $P$ and $Q$ have bounded support. As such, this result does not appear to be directly extendable to the case where $P$ has unbounded support.
\end{remark}

These results stand out against results for the non-regularized transport map in that the rate is either $n^{-1}$ or $n^{-1/2}$ and only requires the source $P$ to be bounded or subgaussian and the target $Q$ to be bounded. In contrast in \cite{Hutter2021_Minimax} the authors showed that under some technical conditions that the minimax optimal rate (up to log factors) for estimating the unregularized transport map is $n^{-\frac{2\alpha}{2\alpha - 2 + d}}$ where $\alpha$ is an assumed smoothness parameter of the optimal map. These rates can be incredibly slow when the dimension $d$ dominates the smoothness $\alpha$.  This suggests that entropic regularization helps break the curse of dimensionality for OT map estimation.  These theorems also plays a key role in the subsequent results we consider and in particular it allows one to control the convergence rate of the sample soft-rank energy to the rank energy.

\subsection{Fast Convergence of \texorpdfstring{$\sREn(P_X, P_Y)^2$}{sRE}}

We first establish a few preliminary facts about $\sRE(P_X,P_Y)^2$ which will be useful when combined with Theorem \ref{thm:subg_conv_rate} and also demonstrate some of the important properties of the soft rank energy as a GoF statistic.

\begin{proposition}\label{prop:properties} 
The soft rank energy satisfies the following properties:
\begin{enumerate}
    \item[(a)] Let $X, X'\overset{i.i.d.}{\sim} P_X$ and $Y, Y' \overset{i.i.d.}{\sim} P_Y$. Then the soft rank energy $\sRE^2$ can also be expressed as 
    \begin{align*}
        \sRE(P_X,P_Y)^2 &= 2\mathbb E\big\|\sR(X)-\sR(Y)\big\|- \mathbb E\big\|\sR(X)-\sR(X')\big\| \nonumber \\ 
        & - \mathbb E\big\|\sR(Y)-\sR(Y')\big\|.
    \end{align*}
    \item[(b)] $\sRE(P_X,P_Y)^2 =\sRE(P_Y,P_X)^2$.
    \item[(c)] $\sRE(P_X,P_Y)^2 = 0$ if $P_X = P_Y$.
\end{enumerate}
\end{proposition}

The proof is deferred to Section \ref{supp:properties} of the supplement. Property (a) is particularly useful because it allows one to use an easily computed formula instead of estimating the integrals and probabilities in (\ref{eq:sre}).
Properties (b) and (c) make $\sRE(P_X,P_Y)^2$ a good candidate for measuring the GoF between distributions.

One can also show that under some regularity conditions that the original rank energy is recovered by sending $\varepsilon$ to 0, which further motivates the notion of the soft rank energy.
\begin{proposition} \label{prop:sRE_eps_convergence} Let $ P_X,P_Y \in \mathcal{P}_{ac}([0,1]^d)$ and $\lambda \in (0,1)$ be such that for $\R = \nabla \phi$ we have $aI \preceq \nabla \phi \preceq bI$ for some $0 < a,b$, and $\R^{-1} \in \mathcal{C}^\alpha$ for some $\alpha \geq 2$. Then
\begin{equation*}
    \lim_{\varepsilon \rightarrow 0^+} \sRE(P_X,P_Y)^2 = \RE(P_X,P_Y)^2.
\end{equation*}
\end{proposition}
The proof appears in the supplement.  Implicit in the proof of Proposition \ref{prop:sRE_eps_convergence} is a result that also characterizes the degree of approximation of soft rank energy with rank energy as a function of the entropy regularization $\varepsilon$.

An important consequence of Theorem  \ref{thm:subg_conv_rate} is a matching convergence rate for the sample soft rank energy:
\begin{theorem} \label{thm:sre_convergence_rate}
    Let $\lambda \in (0,1)$. If $P_X,P_Y$ are $\sigma^2$-subgaussian for $\sigma^2 \geq \frac{1}{2\log 2}$  then 
    \begin{equation*}
        ||\sREn(P_X,P_Y)^2 - \sRE(P_X,P_Y)^2||_{L^2}^2 \lesssim  \frac{b_1(r,d,\sigma^2,\varepsilon)}{\min(\lambda, 1-\lambda)}(m+n)^{-1/2} + \frac{d(m+n)}{mn}.
    \end{equation*}
    If $P_X,P_Y \in \mathcal{P}(\Omega)$ for a bounded set $\Omega$, then
    \begin{equation*}
        ||\sREn(P_X,P_Y)^2 - \sRE(P_X,P_Y)^2||_{L^2}^2 \lesssim  \frac{b_2(r,d,\varepsilon)}{\min(\lambda, 1-\lambda)}(m+n)^{-1} + \frac{d(m+n)}{mn}.
    \end{equation*}
\end{theorem}
The proof, which is deferred to Section \ref{sec:sre_cr_proof} in the supplement, relies on four ingredients: The convergence of the entropic map using either Theorem \ref{thm:subg_conv_rate} or Theorem \ref{thm:bounded_conv_rate}, Proposition \ref{prop:properties} (a),  several applications of the triangle and reverse triangle inequalities, as well as the Efron-Stein inequality (\cite{boucheron2013concentration} Theorem 3.1). The first term in the bound can be thought of as the amount of error incurred from estimating $\sR$ using $m+n$ samples, while the second term bounds the error incurred from estimating an expectation by sampling. 

The factor of $\frac{1}{\min(\lambda, 1-\lambda)}$ comes from the fact that in $\sREn(P_X,P_Y)^2$ there is an equal weight given to both $\{X_1,...,X_m\}$  and $\{Y_1,...,Y_n\}$, even if $m\ll n$  or vice versa. In contrast, the estimate of $\sRn$ weights $P_X$ and $P_Y$ proportionally to $\lambda$ and $(1-\lambda)$ respectively. The intuition here is that if one pays little attention to constructing a good map on the support of $P_X$ then one should expect a large amount of error for $\R(X_i)$ which the weighting scheme only amplifies when computing the statistic.

In a similar flavor, one can show that sample version of sRMMD converges quickly in expectation to the population version. Again this is a consequence of Theorem \ref{thm:subg_conv_rate}.

\begin{theorem} \label{thm:srmmd_convergence_rate}
    Let $k$ be a characteristic kernel such that for all $x$, the function $k(x,\cdot)$ is $l$-Lipschitz with respect to the Euclidean norm. Let $\lambda \in (0,1)$. If $P_X,P_Y$ are $\sigma^2$-subgaussian for $\sigma^2 \geq \frac{1}{2\log 2}$  then 
    \begin{equation*}
        ||\sRMMDn(P_X,P_Y)^2 - \sRMMD(P_X,P_Y)^2||_{L^2}^2 \lesssim  \frac{l^2b_1(r,d,\sigma^2,\varepsilon)}{\min(\lambda, 1-\lambda)}(m+n)^{-1/2} + \frac{l^2d(m+n)}{mn}.
    \end{equation*}
    If $P_X,P_Y \in \mathcal{P}(\Omega)$ for a bounded set $\Omega$, then
    \begin{equation*}
        ||\sRMMDn(P_X,P_Y)^2 - \sRMMD(P_X,P_Y)^2||_{L^2}^2 \lesssim  \frac{l^2b_2(r,d,\varepsilon)}{\min(\lambda, 1-\lambda)}(m+n)^{-1} + \frac{l^2d(m+n)}{mn}.
    \end{equation*}
\end{theorem}

The proof follows essentially the same arguments as the one used to prove Theorem \ref{thm:sre_convergence_rate}, but uses the Lipschitz assumption in place of the reverse-triangle inequality. A discussion of this trick is given in Section \ref{sec:srmmd_proof}.  We remark that many of the most popular kernels do satisfy a Lipschitz continuity condition, and practically all do when restricted to bounded domains, so this restriction is not stringent.
 
\subsection{Utility as a Loss Function}

There are many examples in practice where one will try to train a generative model to create high-dimensional data (for example images, video or, audio signals), which requires a measure of how closely an artificially generated dataset matches a real dataset. In high dimensions there is a need for test statistics that converge rapidly since otherwise one won't be able to distinguish model performance from statistical fluctuations. This is one of the main reasons (along with computational ease) that MMD and energy distances have become prominent in generative modelling. These statistics both have $n^{-1/2}$ convergence rates to their population variants, which is also achieved (up to logarithmic factors) by both of our proposed sRE and sRMMD.  We believe this makes sRE and SRMMD strong contenders for high-dimensional generative modelling.

Another important note is that the computation of a high fidelity estimate of the soft rank energy is easily done through the use of Sinkhorn's algorithm \cite{cuturi2013sinkhorn} which is based on fixed point iterations. One can perform a fixed number of iterations and then employ automatic differentiation \cite{paszke2017automatic} to obtain a gradient through this method. This approach is not novel to this work and so-called ``Sinkhorn layers''  have become popular in the neural network literature \cite{adams2011ranking,emami2018learning, feydy2019interpolating}. With these in hand, the computation of the soft rank energy and soft rank MMD becomes straightforward:  simply attach a series of Sinkhorn layers to the end of a generative model to obtain the soft rank map, then use that soft rank map to transform the data, and then compute the soft rank energy or MMD on the transformed data. 

\section{Applications}
\label{sec:applications}
\subsection{sRE and sRMMD as the Loss Function in a Generative Model}\label{sec:mnist}
Generative modeling is used to implicitly approximate a complex, high-dimensional distribution from a finite number of samples. When trained successfully, it allows one to draw new samples from the underlying distribution. A seminal framework  to learn the generative model is the \emph{generative adversarial network (GAN)} \cite{goodfellow2014generative} that optimizes a difficult minimax program. A simpler approach known as a \emph{generative moment matching network (GMMN)} \cite{li2015generative} instead minimizes the differentiable MMD \cite{gretton2012kernel}. In this section,  we train a generative model to minimize the proposed sRE and sRMMD and illustrate their effectiveness compared to GMMN on the benchmark MNIST digits data set \cite{lecun1998gradient}. A brief description of the model architecture and the training procedure is given below.
\paragraph{Architecture:} We use the same architecture used in \cite{li2015generative} which consists of (a) a generative network and (b) an auto-encoder \cite{kingma2013auto}. The generative network consists of 3 intermediate ReLU nonlinear layers and
one logistic sigmoid output layer. The auto-encoder has 4 layers, 2 for the encoder and 2 for the decoder with sigmoid nonlinearities. The auto-encoder is used to learn an almost lossless low-dimensional latent space for the high-dimensional, complex data. The generative network is used to generate samples in this low-dimensional space. The trained decoder then turns these samples into meaningful high-dimensional data.  For further details, we refer the reader to \cite{li2015generative}.

\paragraph{Training:} We train the auto-encoder and generator network separately. First an auto-encoder is learned to produce a low-dimensional representation for the MNIST digits with latent dimension of $8$. Then we train the generative network to learn to generate samples in this low-dimensional representation space via minimizing either $\sre$ or $\srmmd$. Both $\sre$ and $\srmmd$ are computed with entropic regularizer $\varepsilon =1$, where the soft rank maps are obtained via Sinkhorn's algorithm \cite{peyre2019computational} with a maximum of 5000 iterations. To compute $\srmmd$, we employ a Gaussian mixture kernel {\small $k(x, x')= \frac{1}{6}\sum_{q=1}^6\exp(-\frac{\|x-x'\|^2}{2\sigma_q^2})$} with the bandwidth parameter $\sigma= (1, 2, 4, 8, 16, 32)$.

Both the auto-encoder and the generator are trained on a minibatch size of $256$ using the Adam optimizer \cite{kingma2019introduction} with a learning rate of $0.001$ over $100$ epochs. 
\begin{figure}[H] 
  \centering
  \subfloat[][$\mathtt{MMD}$]{\includegraphics[width=.35\linewidth, height = 5.5cm]{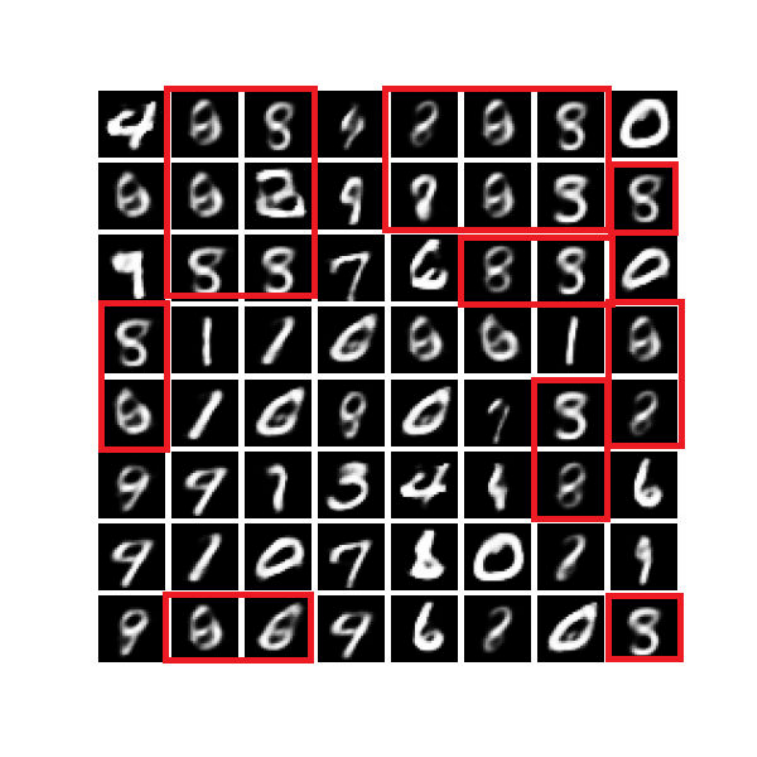}\vspace{-6mm}}\hspace{-8mm}
  \subfloat[][$\mathtt{sRE}$]{\includegraphics[width=.35\linewidth, height = 5.5cm]{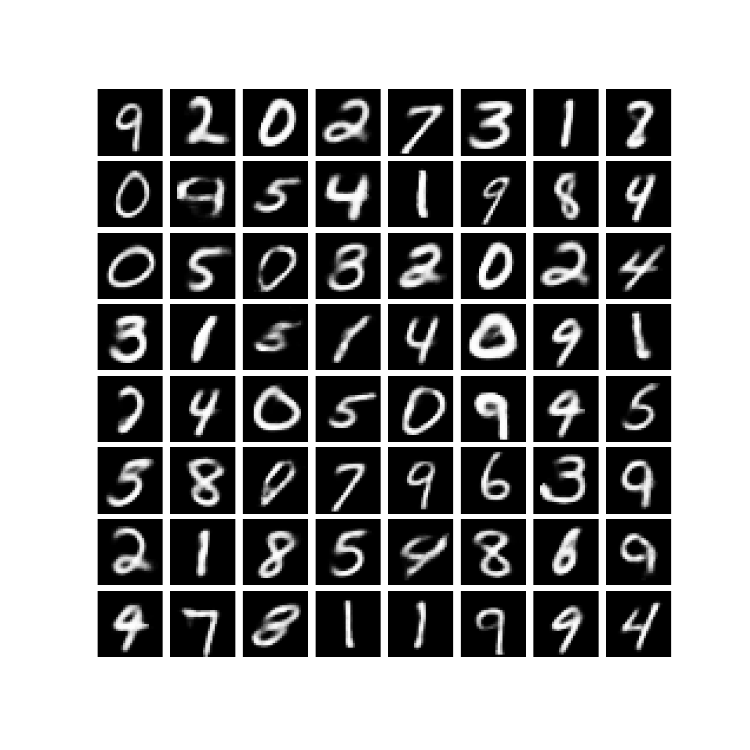}\vspace{-6mm}}\hspace{-8mm}
  \subfloat[][ $\mathtt{sRMMD}$]{\includegraphics[width=.35\linewidth, height = 5.5cm]{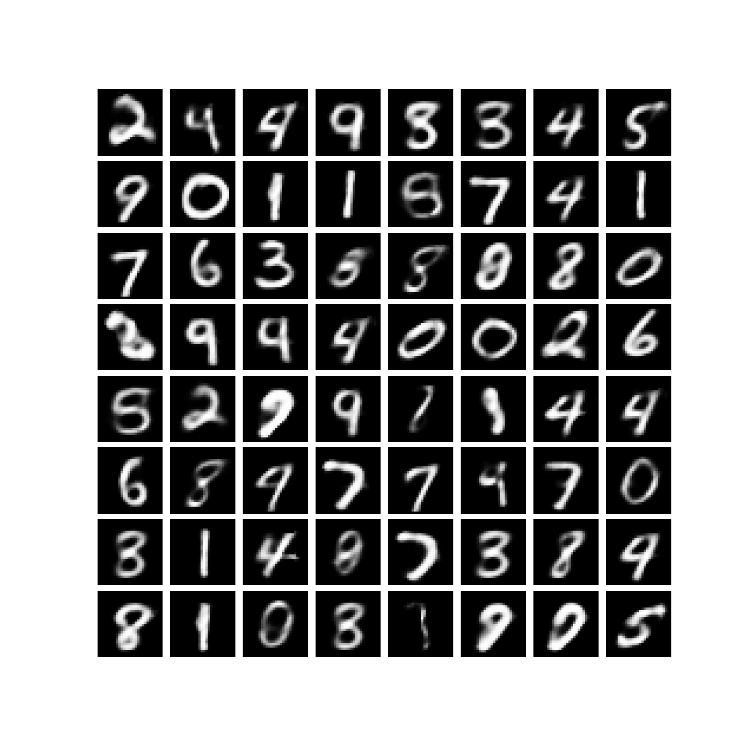}\vspace{-6mm}}
  \caption{Comparison of the MNIST digits generated via minimizing MMD, sRE, and sRMMD. The model architecture and training procedure are kept same for all methods. Red boxes in (a) indicate the abundance of the same digit e.g., 8 when using MMD.}
  \label{fig:mnist_result_01}
  \vspace{-4mm}
\end{figure}

\paragraph{Result:} 
It is apparent that the generator minimizing MMD lacks in diversity (mostly producing 8's). Besides, most of the digits are barely recognizable which indicates that the generator performs poorly if it is trained to minimize MMD. In contrast, the generator minimizing either sRE or sRMMD produces a diverse set of digits and almost all of them are unambiguous. Thus, both sRE and sRMMD appear to be better choices to minimize than MMD. Though in the current setup, the generator minimizing sRMMD works well, we empirically observe that the performance (e.g., diversity, unambiguity) heavily depends on the choice of the bandwidth parameter $\sigma$ and the entropic regularizer $\varepsilon$. In our case, we find it beneficial to use smaller $\sigma$ when using a larger $\varepsilon$. Plots showing the dependency of the sRMMD generator on $\varepsilon$ and $\sigma$ can be found in Section \ref{supp:mnist} in the supplement.
\subsection{Generating Valid Knockoffs using sRMMD}
In applications where the goal is to discover relevant features that can explain certain outcomes (e.g., metabolites or genes related to Crohn's disease \cite{lloyd2019multi, franke2010genome}), it is important that the set of selected features contains as few false discoveries as possible. One way to do that is to control the false discovery rate (FDR) at a prespecified level $q\in(0, 1)$. The classical setup to control FDR depends on assumptions on how the features and the outcomes are related \cite{benjamini1995controlling, gavrilov2009adaptive}. A novel FDR control framework, namely Model-X knockoffs \cite{candes2016panning}, provides an alternative to the traditional methods by assuming no knowledge about the association between the features and the outcomes. Given the set of explanatory random variables $X =(X^1,...,X^d)\in \mathbb R^d$ and the outcome variable $Y\in \mathbb R$, the Model-X knockoff framework works in four steps to select relevant variables while controlling the FDR. 
\begin{itemize}
    \item[(a)] Generate a synthetic set of features called knockoffs $\tilde X =(\tilde X^1, \dots, \tilde X^d)$ which are independent of $Y$ conditionally on $X$, and satisfy what is referred to as the \emph{pairwise exchangeability condition}.
\begin{align}
    & (  X, \tilde{  X})_{swap(B)} \overset{d}{=} (  X, \tilde {  X}), \forall B\subset \{1, \dots, d\}, \label{eq:exchangeability}
\end{align}
where $swap(B)$ exchanges the positions of any variable $X_j$, $j\in B$, with its knockoff $\tilde X_j$.
    \item [(b)] Produce a knockoff statistic $W_j = w_j([ X, \tilde{ X}], y)$ for $j\in\{1, \dots, d\}$ to asses the importance of each feature. Here, $w_j(\cdot)$ is any function with the flip sign property\footnote{a function $f$ is said to have flip-sign property if $f(u, v)= -f(v, u)$.}.
    \item [(c)] Find a data dependent threshold $\tau$ via,
\begin{align}  \label{eq:threshold}
    \tau = \min_{t>0}\Big \{t: \frac{1+ |\{j: W_j \leq -t\}|}{|\{j: W_j \geq t\}|}\leq q\Big \}.
\end{align}
\item[(d)] Select the set of variables: $\hat{\mathcal S} = \{j: W_j\geq \tau\}$.
\end{itemize}

Performance of the Model-X framework depends on the quality of the knockoffs, that is, to what extent they satisfy pairwise exchangeability. One way to achieve this is to approximate only the first two moments (mean and covariance) assuming that the joint distribution of $X$ is a multivariate Gaussian. This is often called a second-order method \cite{candes2016panning}. Other methods proposed in \cite{salimans2016improved,liu2018auto,romano2020deep,sudarshan2020deep} take a generative modeling approach to satisfy \eqref{eq:total_loss} and sample the knockoffs. A brief descriptions of these methods is provided in Section \ref{supp:related_works_knockoffs} in the supplement.

In this paper, we take a generative modeling approach where we propose to use sRMMD as the loss to satisfy the pairwise exchangeability condition \eqref{eq:exchangeability}. 
\subsubsection{An sRMMD-Based Knockoff Generator}\label{sec:architecture} 
We use a generative model similar to the one used in \cite{romano2020deep}. The generative model has a deep neural network $f_\theta$ that takes $X\sim P_X\in \mathbb R^d$ and a noise vector $V\sim \mathcal N(0, I_d)\in \mathbb R^d$ as inputs and returns an approximate copy of knockoff $\tilde X = f_\theta(X, V)\in \mathbb R^d$. Here $\theta$ denotes the set of the parameters which is learned from the data. The network is fed with $\{X_i\}_{i=1}^n\in \mathbb R^d$ independent observations and generates $\tilde X_i= f_{\theta}(X_i, V_i)$ for $1\leq i\leq n$. Let $\mathbf X, \tilde{\mathbf X}\in \mathbb R^{n\times d}$ be the matrices having these observations and their knockoffs as row vectors, respectively. To ensure that the knockoffs are of good quality (i.e., the joint distribution of $(X_i, \tilde X_i)$ satisfies \eqref{eq:exchangeability} and $X_i$ and $\tilde X_i$ are as different as possible, for $1\leq i\leq n$), we minimize the following loss,
\begin{align}\label{eq:total_loss}
    \ell(\mathbf X, \tilde {\mathbf X})\! =\! \underbrace{\srmmd \Big [ (\mathbf X',\! \tilde{\mathbf X}'),\! (\tilde{\mathbf X}'', \mathbf X'')\Big]}_{\text{full-swap}}+ \underbrace{\srmmd \Big [ (\mathbf X',\! \tilde{\mathbf X}'), (\mathbf X'', \tilde{\mathbf X}'')_{\text{swap}(B)}\Big]}_{\text{partial-swap}}\! +\! \gamma \ell_{Decor}(\mathbf X, \tilde{\mathbf X}),
\end{align}
where $\mathbf X', \mathbf X''\in \mathbb R^{n/2\times d}$, and $\mathbf {\tilde X}', \mathbf {\tilde X}''\in\mathbb R^{n/2 \times d} $ are obtained by randomly splitting $\mathbf X$ and $\mathbf {\tilde X}$ in half and $B$ is a chosen random subset of {\small $\{1, \dots, d\}$}, such that $j\in B$ with probability $1/2$. We adapt the idea of splitting and swapping from \cite{romano2020deep}. The first two terms in \eqref{eq:total_loss} help to achieve pairwise exchangeability. The last term in \eqref{eq:total_loss} trades off power versus FDR by decorrelating the
variables with the knockoffs. We adapt this loss term from \cite{romano2020deep} with hyperparameter $\gamma>0$, which is defined as
\begin{align*}
    \ell_{Decor}(\mathbf X, \tilde{\mathbf X}) =  \|\mathtt{diag}( \Sigma_{ X \tilde {X}}) - 1 + s^*_{\mathtt{SDP}}(\Sigma_{ X X})\|_2^2.
\end{align*}
$\Sigma_{XX}$ and $\Sigma_{X\tilde X}$ are the empirical covariance matrix of $X$ and the empirical cross covariance matrix between $X$ and $\tilde X$, respectively, and {\small $s^*_{\mathtt{SDP}}(\Sigma_{ XX}) =  \arg \min_{s\in [0, 1]^d} \sum_{j=1}^p|1- s_j|$} such that $2 \Sigma_{XX} \succeq \mathtt{diag}(s)\succeq 0$. 
The loss \eqref{eq:total_loss} is differentiable and therefore any gradient descent method can be adopted to train the generative model. Generally training is done in minibatches size of $m \ll n$. At each epoch of training, for each batch of size $m$, only one $B$ is picked randomly to compute \eqref{eq:total_loss}, which may prevent one to achieve pairwise exchangeability $\eqref{eq:exchangeability}$ to a great extent. That is why, we recommend generating multiple batches by reshuffling the training set of size $n$ several times at each epoch so that multiple sets of random $B$ are picked. To train efficiently and effectively, we also suggest using $\varepsilon\propto d$ to compute $\srmmd$. This suggestion is backed by the empirical evidence shown in Section \ref{supp:srmmd_w.r.t.epsilon} in the supplement. In addition, we provide empirical justification for using $\srmmd$ over $\sre$ in \eqref{eq:total_loss} in Section \ref{supp:srmmd_over_sre} in the supplement. 

The details of the generative model and the training procedure for any fixed $\varepsilon$ and $\gamma$ are summarized in Sections \ref{supp:knockoff_generation} and \ref{alg:alg1} in the supplement, respectively.
\subsubsection{Knockoff Experiments on Synthetic Benchmarks}\label{subsec:synthetic_setting}
We compare the performance of sRMMD knockoffs with other benchmarks, namely second-order knockoffs \cite{candes2016panning}, knockoffGAN \cite{jordon2018knockoffgan}, deep knockoff \cite{romano2020deep}, and deep direct likelihood knockoff (DDLK) \cite{sudarshan2020deep} on several synthetic (four Gaussian and non-Gaussian distributional settings adapted from \cite{romano2020deep}) and a real-world dataset. To be on an equal footing with deep knockoff, we remove the second-order term from the loss used in \cite{romano2020deep} and denote it as the MMD knockoff. For all comparisons, we use publicly available implementations of the code and used their recommended configurations and hyperparameters (if available).

For each distributional setting, we train the knockoff generator on a set of $n= 2000$ samples each of dimension $d=100$. We compute $\srmmd$ using a Gaussian mixture kernel $k(x, x') = \frac{1}{8}\sum_{k=1}^8 \exp(-\|x - x'\|/(2 \sigma_k^2)])_2^2$ with $\sigma = (1, 2, 4, 8, 16, 32, 64, 128)$, where the soft rank maps are obtained via Sinkhorn's algorithm (with a maximum of 5000 iterations). We update the network \eqref{eq:total_loss} using stochastic gradient descent with momentum \cite{bottou2012stochastic}. The minibatch size, learning rate and the number of epochs are set to 500, 0.01, and 100, respectively. The training procedure is detailed in Algorithm \ref{alg:alg1} in the supplement. 

Below, we briefly describe each distributional setting and other optimal hyperparameters e.g., $\varepsilon$, $\gamma$ for training. \begin{itemize}
    \item[(a)] \textbf{Multivariate Gaussian AR1:} An autoregressive model of order one in which $X\sim \mathcal N(0, \Sigma)$, $\Sigma_{ij}=\rho^{|i-j|}$, $\rho=0.5$. We set the decorrelation penalty $\gamma = 1$, entropic regularizer $\varepsilon = 100$.
    \item[(b)] \textbf{Gaussian Mixture Model (GMM)}: $X\sim \sum_{k=1}^4 \tau_k \mathcal N(0, \Sigma_k)$, where the covariance matrix is $(\Sigma_k)_{ij} = \rho_k^{|i-j|}$ for $k=1,\dots, 4$. $(\rho_1, \rho_2, \rho_3, \rho_4)= (0.6, 0.4, 0.2, 0.1)$,  $(\tau_1, \tau_2, \tau_3, \tau_4)=(0.27, 0.23, 0.23, 0.27)$.  We set $\gamma=1$, and $\varepsilon = 100$.
    \item[(c)] \textbf{Multivariate Student's $t$-Distribution}: A heavy-tailed distribution with zero mean and $\nu=3$ degrees of freedom, such that $X = \sqrt{\frac{(\nu -2)}{\nu}} \frac{ Z}{ \sqrt{\Gamma}}$, where $Z\sim \mathcal N(0, \Sigma)$ with $\Sigma$ as in (a) and $\Gamma$ is independently drawn from a Gamma distribution with shape and rate parameters both equal to $\nu/2$. We set $\gamma=1$ and $\varepsilon=100$.
    \item [(d)] \textbf{Sparse Gaussian}: Given $W\sim \mathcal N(0, 1)$ and a random subset $A \in \{1, \dots, p\}$ of size $|A| = L$, we set $X_j= \sqrt{\frac{\binom{L}{p}}{\binom{L-1}{p-1}}}. \begin{cases}W,\;\; \text{if}\;\; j\in A, \\ 0, \;\; \text{otherwise.}\end{cases}$ We set $L=30$, $\gamma = 0.1$ and $\varepsilon = 100$.
\end{itemize}

After training, we draw $m_t =200$ new i.i.d. samples as the test set and simulate the outcome as $\bm y = \mathbf X_t\beta+ \bm z$, where $\mathbf X_t\in \mathbb R^{m_t\times d},\; d=100$, $\bm y\in \mathbb R^{m_t}$, $\bm z\sim \mathcal N(0, I)$, and $\beta\in \mathbb R^d$ is the coefficient vector.  The vector $\beta$ is all zeros except randomly chosen 20 entries, each having an amplitude equal to $\upsilon/\sqrt{m_t}$, where $\upsilon$ is the amplitude parameter. Then we generate the knockoff matrix $\mathbf {\tilde X}_t$ and perform LASSO regression \cite{friedman2010regularization} on $[\mathbf X_t, \mathbf{\tilde X}_t]$ via solving,
\begin{align*}
    \begin{bmatrix}
    \hat \beta\\ \hat {\tilde \beta}\\
    \end{bmatrix} = \arg \min_{( \beta,  \tilde \beta)}& \frac{1}{m_t} \Big \|\bm y - \begin{bmatrix}
    \mathbf X_{t}\mathbf {\tilde X}_{t}
    \end{bmatrix}
    \begin{bmatrix}
    \beta\\ \tilde \beta\\
    \end{bmatrix}\Big\|_2^2 + \alpha_L  \begin{Vmatrix}
    \beta\\ \tilde {\beta}\\
    \end{Vmatrix}_1, 
\end{align*}
where the $\hat {\beta}\in\mathbb R^d$ and $\hat{\tilde \beta}\in\mathbb R^d$ are the coefficient vectors corresponding to the original variables, and knockoff variables, respectively and $\alpha_L$ is the LASSO penalty. We consider LASSO regression since it  works best when the true model is close to linear. We estimate the coefficient vectors using $\alpha_L = 0.01$ and take the absolute difference {\small $W_j = |\hat \beta_j|- |\hat{\tilde \beta}_j|$} as the knockoff statistic for $1\leq j\leq d$. We repeat the experiments $500$ times for different values of $\upsilon$, and compare the power versus FDR tradeoff with the benchmarks. 
\begin{figure}[htbp]
\centering
  \subfloat[][Multivariate Gaussian AR1]{\includegraphics[width=.48\linewidth]{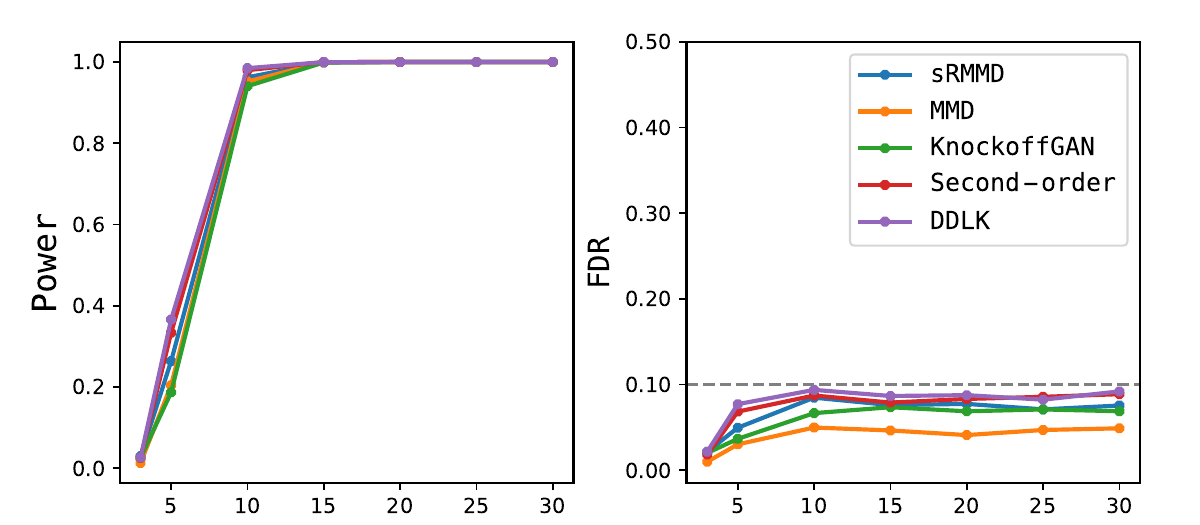}}\quad
  \subfloat[][Gaussian Mixture Model]{\includegraphics[width=.48\linewidth]{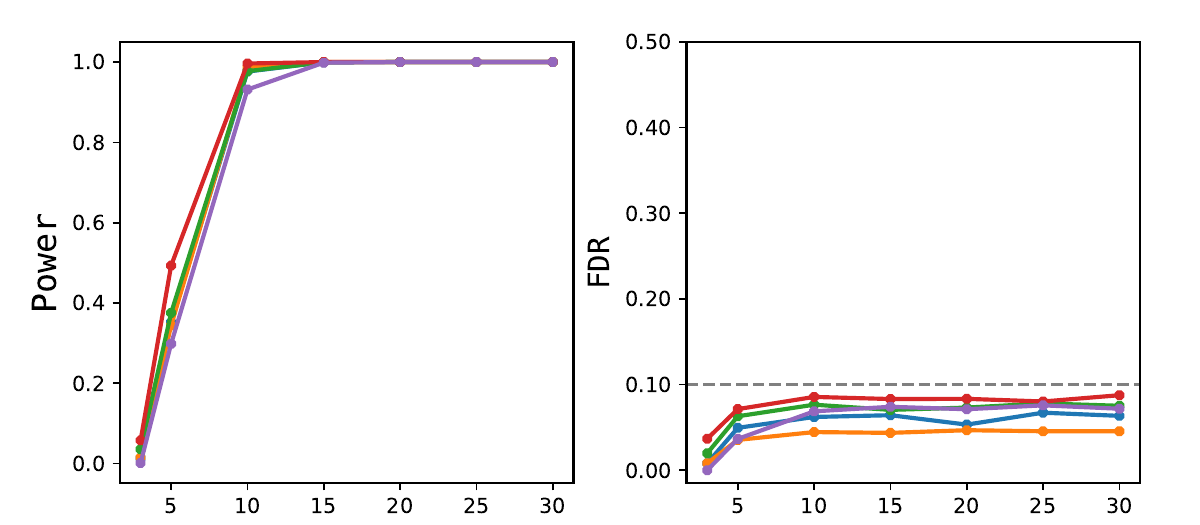}}\\
    \subfloat[][Multivariate Student's t]{\includegraphics[width=.48\linewidth]{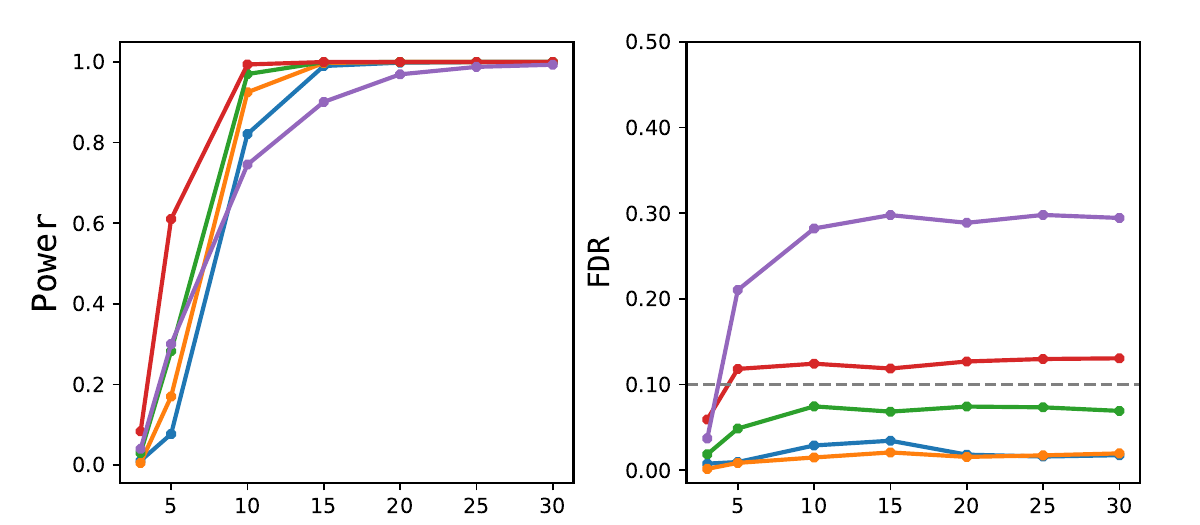}}\quad
  \subfloat[][Sparse  Gaussian]{\includegraphics[width=.48\linewidth]{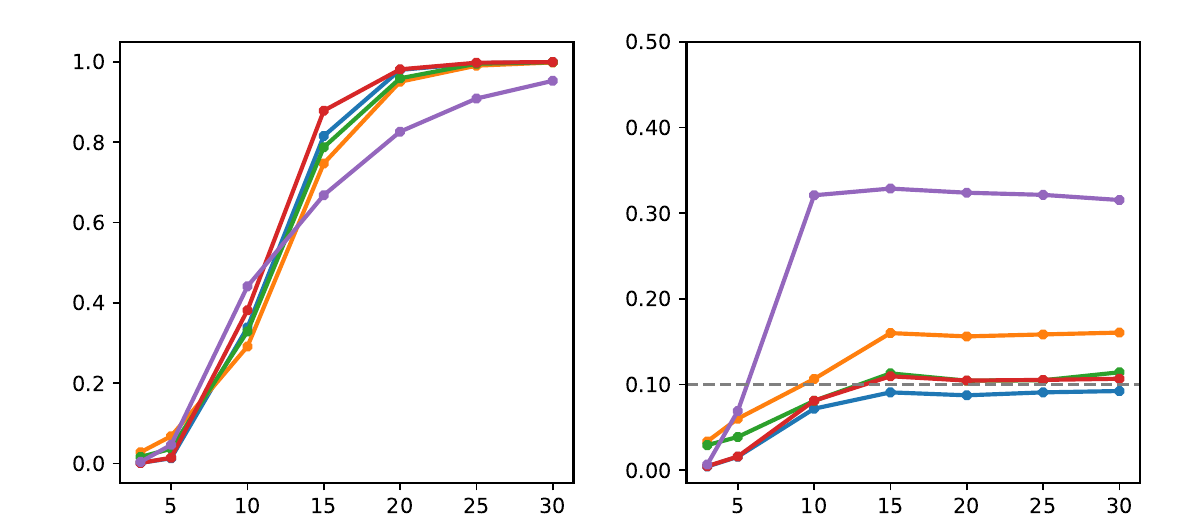}}
    \caption{Average FDR and power computed over 500 independent experiments are shown on the $y$-axis  for each synthetic benchmark. The FDR level is set to 0.1.  The $x$-axis represents the amplitude parameter $\upsilon$.}
    \label{fig:knockoff}
\end{figure}
\vspace{-4mm}
\paragraph{Result:} Figure \ref{fig:knockoff} shows the FDR versus power tradeoff with respect to the amplitude parameter $\upsilon$. In case of Gaussian AR1 setting, all methods showcase similar detection power and FDR control at level $q=0.1$ over the entire amplitude region. In the GMM setting, we observe that each method achieves similar power and controls the FDR at $q= 0.1$ like the multivariate Gaussian setting. This can be explained by the fact that here the GMM and Gaussian settings belong to the same exact family of distributions since the su    m of independent Gaussians is a Gaussian. For the heavy-tailed multivariate Student's $t$-distribution, sRMMD, MMD and knockoffGAN can control the FDR at $q=0.1$, however, the second-order method fails to control the FDR, likely because it assumes that the underlying distribution is multivariate Gaussian. DDLK also fails to control the FDR in this case. For the sparse Gaussian setting, sRMMD knockoffs showcase the best FDR versus power tradeoff among all methods over the entire amplitude region. Second-order, and knockoffGAN methods also achieve similar power and control the FDR at $q=0.1$. In contrast, MMD and DDLK knockoffs fail to control the FDR at $q= 0.1$.  

We do acknowledge the fact that in all settings knockoffGAN also achieves high power and controls the FDR at $q=0.1$. KnockoffGAN is a complex generative adversarial architecture that requires training of four interconnected neural networks. Instead of minimizing a GoF statistic (e.g., MMD or sRMMD), knockoffGAN minimizes the binary cross-entropy loss in order to satisfy the pairwise exchangeability condition. In addition, knockoffGAN uses the mutual information loss to make the variables and their knockoffs as independent as possible which is a much stronger notion than decorrelation. In contrast to knockoffGAN, our method is simple and easy to implement, yet achieves comparable FDR versus power tradeoff. 
\subsubsection{Application to a Real Metabolomics Dataset}
We apply the proposed knockoff filter to a publicly available metabolomics dataset in order to discover important biomarkers with FDR guarantees. We use a study titled \emph{Longitudinal Metabolomics of the Human Microbiome in Inflammatory Bowel Disease} \cite{lloyd2019multi} which is available at the Metabolomics Workbench through the National Metabolomics Data Repository (NMDR) website \url{https://www.metabolomicsworkbench.org/} under the project DOI: 10.21228/M82T15 and sponsored by the Common Fund of the NIH. The study is related inflammatory bowel disease (IBD) and conditions including ulcerative colitis (UC) and Crohn’s disease (CD), and seeks to identify important metabolites (biological products produced as intermediates during metabolism) associated with these diseases. We use the \emph{C18 Reverse-Phase Negative Mode} dataset which was collected under this study. The dataset contains 546 samples, each having an average of 91 metabolites. Each sample belongs to one of the three classes (UC, CD, and non-IBD) and assign the response $y$ to one of $\{0, 1, 2\}$ to reflect this. We preprocess the dataset in three steps: (i) removing the metabolites that have more than $20\%$ missing values which retains only 80 metabolites out of 91, (ii) applying k-nearest neighbor (KNN) missing value imputation technique to fill out the existing missing values, and (ii) standardizing the features by removing the mean and scaling to unit variance. For this dataset, we use the same generative architecture described in Section \ref{sec:architecture}.  We choose entropic regularizer $\varepsilon =50$ and kernel bandwidth $\sigma = (1, 2,4, 8, 16, 32, 64, 128)$ to compute $\srmmd$. We pick $\gamma = 1$. We train the generator on a minibatch of $250$ samples according to Algorithm \ref{alg:alg1}. 

After training, we generate the knockoffs, apply the random forest (RF) classifier \cite{trainor2017evaluation} to produce knockoff statistics. The two RF parameters---the number of features that are randomly selected at each node and the number of trees---are set to 9 (the closest integer to the recommended $\sqrt{80}$) and 500, respectively. We take the difference between the feature importance scores \cite{trainor2017evaluation} corresponding to the original variables and the knockoffs as the knockoff statistics. Since the generated knockoffs are random, we repeat the whole procedure $100$ times and select those metabolites that appear at least 70 times out of 100 instances, setting the FDR level at $q = 0.05$. In absence of the ground truth, to qualitatively analyze the performance we cross-reference the selected metabolites with published literature. We list the selected metabolites along the references in Table \ref{tab:metabolites}.
\begin{table}[H]
    \centering
    \begin{tabular}{c|c|c|c|c|c}
    \hline
    \diagbox{Metabolites}{\makecell{Method\\(N)}} & \makecell{sRMMD \\$(21)$} & \makecell{Second Order\\$(16)$} & \makecell{ MMD\\$(27)$} &  \makecell{KnockoffGAN\\$(20)$}  & Reference\\ \hline 
\makecell{1.2.3.4-tetrahydro-beta-\\carboline-1.3-dicarboxylate}
                       & \ding{52}&\ding{52} &\ding{52} &\ding{52}&\cite{volkova2021predictive}\\
              urobilin & \ding{52} &\ding{52} & \ding{52}& \ding{52}& \cite{qin2012etiology}\\
              adrenate & \ding{52} & \ding{52} & \ding{52}& \ding{52}& \cite{lloyd2019multi}\\
          12.13-diHOME & \ding{52} & \ding{52} & \ding{52}& \ding{52} &  \cite{levan2019elevated}\\
            salicylate & \ding{52} &\ding{52} & \ding{52}& \ding{52}& \cite{caprilli2009long}\\
            saccharin  & \ding{52} &\ding{52} & \ding{52}& \ding{52}& \cite{qin2012etiology}\\
              caproate &\ding{52} & \ding{52}&\ding{52}& \ding{52} & \cite{lee2017oral}\\
            olmesartan & \ding{52} &\ding{52} & \ding{52}& \ding{52}& \cite{saber2019olmesartan}\\
         phenyllactate & \ding{52} &\ding{56} & \ding{52}& \ding{52}& \cite{lavelle2020gut}\\ 
     taurolithocholate & \ding{52} &\ding{56} & \ding{52}& \ding{56}& \cite{bauset2021metabolomics}\\ 
     docosapentaenoate & \ding{52} &\ding{52} & \ding{52}& \ding{52}& \cite{solakivi2011serum}\\ 
      docosahexaenoate & \ding{52} &\ding{56} & \ding{52}& \ding{52}&\cite{solakivi2011serum}\\ 
        dodecanedioate & \ding{52} &\ding{56} & \ding{52}& \ding{56}& \cite{lee2017oral}\\
        hydrocinnamate & \ding{52} &\ding{52} & \ding{52}& \ding{52}& \cite{lee2017oral}\\
       eicosatrienoate & \ding{52} &\ding{56} & \ding{52}& \ding{52}& \cite{kuroki1997serum}\\ 
       9.10-diHOME & \ding{52} & \ding{52} & \ding{52}& \ding{52} &  \cite{lloyd2019multi}\\
          arachidonate & \ding{56} & \ding{56}& \ding{52}& \ding{52}& \cite{lloyd2019multi} \\
           myristate & \ding{56} & \ding{56}& \ding{56}& \ding{56}& \cite{fretland1990colonic} \\
         \hline 
        Total = 18 & 16 & 11 & 17 & 15 \\ \hline
        Detection power $(\%)$ & 76 & 69 & 63 & 75\\\hline 
    \end{tabular}
    \caption{N is the total number of selected metabolites. DDLK  finds almost every metabolites as significant, therefore loses its purpose as a FDR control technique. That is why we refrain from adding it here.}
    \label{tab:metabolites}
    \vspace{-4mm}
\end{table}
Out of $80$, we have found $18$ metabolites to have an impact on IBD in the published literature. Though MMD knockoffs detect most of the significant metabolites, the detection percentage is very low. Second-order method performs better compared to MMD in terms of power though it misses several significant metabolites. On the other hand, sRMMD and KnockoffGAN have almost the same detection power, and outperform second-order and MMD methods. 
\section{Conclusion and Future Directions}
In this paper, we identified some major limitations in use of the recently proposed multivariate rank-based GoF statistics based on the theory of optimal transportation, namely high sample and computational complexity in high dimensions and lack of differentiability, which limits their use in gradient-based machine learning methods. We show that using entropic maps derived from entropic regularization of the optimal transportation problem alleviates these issues and leads to efficient statistics for GoF testing. Furthermore, we show that this regularization allows the use of these GoF statistics for generative modeling in high dimensions.  

One future research direction is to evaluate the effect of different distributions as the target distribution in place of $\text{Unif}([0,1]^d)$ such as spherical uniform distribution \cite{hallin2021distribution}. It may be important to characterize the effect of this choice for the soft rank map and soft rank-based statistics proposed in this paper. 

A related problem concerns the dependence on $\varepsilon$ in the entropic regularization.  In particular, the \textit{convergence rate} of the sRE and sRMMD to the RE and RMMD as $\varepsilon\rightarrow 0^{+}$ is the subject of ongoing work.  While Proposition \ref{prop:sRE_eps_convergence} provides a coarse convergence result, we conjecture that a more precise bound may hold and that the extra smoothness assumptions on the measures may not be required for this result to be true, and that compactness and together with weaker smoothness assumptions may suffice.  Beyond analyzing the convergence rate in $\varepsilon$, it is important to understand the advantages of taking $\varepsilon\gg 0$, beyond the improvements in statistical and computational complexity. For example, one may wish to understand the impact $\varepsilon$ has on the robustness, or stability, of sRE in response to small perturbations in the distributions. Finally, extending the result of Theorem \ref{thm:subg_conv_rate} to relax the compactness assumption on the target measure as well is an important direction for future research.

\section*{Acknowledgements}
This research was sponsored by the U.S. Army DEVCOM Soldier Center, and was accomplished under Cooperative Agreement Number W911QY-19-2-0003. The views and conclusions contained in this document are those of the authors and should not be interpreted as representing the official policies, either expressed or implied, of the U.S. Army DEVCOM Soldier Center, or the U.S. Government. The U. S. Government is authorized to reproduce and distribute reprints for Government purposes notwithstanding any copyright notation hereon.

SM was totally supported by W911QY-19-2-0003. MW is supported by NSF CCF-1553075. JMM acknowledges support from the NSF through grants DMS-1912737, DMS-1924513. SA acknowledges support by NSF CCF-1553075, NSF DRL 1931978, NSF EEC 1937057, and AFOSR FA9550-18-1-0465. All authors acknowledge  support through the Tufts TRIPODS Institute, supported by the NSF under grant CCF-1934553.

\newpage
\bibliographystyle{unsrt}
\bibliography{ref_arxiv.bib}

\newpage

\centerline{\huge{\textbf{Supplement}}}

\section{Lack of Gradient for the Rank Energy} \label{sec:lack_of_grad}

The issue is to estimate the gradient of $\ell(\theta) = \REn(\{X_i\}_{i=1}^m, \{T_\theta(Y_j)\}_{j=1}^n)^2$ with respect to $\theta$. In our notation this expression should actually be written $\REn(P_X, (T_\theta)_\#P_Y)^2$. Recalling the definition of $\REn(P_X, (T_\theta)_\#P_Y)^2$, we have
\begin{align*}
    &\REn(\{X_i\}_{i=1}^m, \{T_\theta(Y_j)\}_{j=1}^n)^2 \\
    =& \REn(X_m, T_\theta(Y_n))^2 \\
    =& \frac{2}{mn}\sum_{i=1}^{m}\sum_{j=1}^n \|\Rn(X_i)- \Rn(T_\theta(Y_j))\|- \frac{1}{m^2}\sum_{i,j=1}^{m}\|\Rn(X_i)- \Rn(X_j)\|\\ -&\frac{1}{n^2}\sum_{i,j}^{n}\|\Rn(T_\theta(Y_i))- \Rn(T_\theta(Y_j))\| \\
    =& \frac{2}{mn}\sum_{i=1}^{m}\sum_{j=1}^n \| U_{\sigma_\theta(X_i)} - U_{\sigma_\theta(T_\theta(Y_j))}\| - \frac{1}{m^2}\sum_{i,j=1}^{m}\|U_{\sigma_\theta(X_i)} - U_{\sigma_\theta(X_j)}\|- \frac{1}{n^2}\sum_{i,j}^{n}\|U_{\sigma_\theta(T_\theta(Y_i))} - U_{\sigma_\theta(T_\theta(Y_j))}\|, 
\end{align*}
where $\sigma_\theta$ is the optimal permutation for transporting $\{X_1,...,X_m\} \cup \{T_\theta(Y_1),...,T_\theta(Y_n)\}$ to $\{U_1,...,U_{m+n}\}$ where $U_i \sim \text{Unif}([0,1]^d)$. From the last expression it is clear that the expression only changes value when the permutation $\sigma_\theta$ changes. This poses problems for the derivative of the loss functions with respect to $\theta$. If $T_\theta(y)$ varies smoothly with $y$ then any small enough change in $\theta$ can either leave the value of $ \mathtt{RE}^{m,n}(X_m, T_\theta(Y_n))^2$ unchanged since the permutation is unchanged, or it causes a jump in the objective when the permutation changes. In the first case the derivative is zero, and in the second it is not well-defined. 

\section{Proof of Theorem \ref{thm:subg_conv_rate}} \label{sec:cr_proof}

The proofs of this result is quite involved. We first review some background and notation and then build up to the result using a series of lemmas and propositions. To help with the exposition, the proofs of these steps are given at the end of the section. We also will adopt the convention that all constants $C,C_i,C_{j,r,d}$ \textit{do not change values from line to line}. At the end of the section we include a table of the constants, including their relations to each other or the source from which they are taken. 

We first introduce the notation for the entropy-regularized optimal transport distance between $P,Q$:
\begin{align}
        S_\varepsilon(P,Q) = \inf_{\pi \in \Pi(P,Q)} &\int \int \frac{1}{2}\norm{x-y}^2 d\pi(x,y) + \varepsilon D_{KL}(\pi || P \otimes Q)  \label{eq:ent_primal} \\
        = \sup_{f \in L^1(P), g \in L^1(Q)} &\int f dP + \int g dQ \nonumber \\
        \hspace{1cm}- \varepsilon&\int \int \exp \left (\frac{1}{\varepsilon}\left [f(x) + g(y) - \frac{1}{2}\norm{x-y}^2 \right ]\right ) dP(x) dQ(y) + \varepsilon. \label{eq:ent_dual} 
\end{align}
The optimal coupling of $(P,Q)$ in (\ref{eq:ent_primal}) will be denoted $\pi_\varepsilon$, and for $(P,Q^n)$ it will be denoted $\pi_\varepsilon^n$. These are both guaranteed to exist if $P,Q$ have finite second moment which is always the case for both bounded and subgaussian distributions.
The optimal dual potentials of $(P,Q)$ in (\ref{eq:ent_dual}) will be denoted $(\feps,\geps)$. For $(P,Q^n)$ the optimal dual potentials will be denoted $(\feps^n,\geps^n)$. As a consequence of (\ref{eq:opt_relation}), we can always choose $\feps,\geps$ to satisfy 
\begin{align}
    \int \exp \left (\frac{1}{\varepsilon}\left [\feps(x) + \geps(y) - \frac{1}{2}\norm{x-y}^2 \right ]\right ) dP(x) &= 1 \text{ for all } y \in \mathbb{R}^d \label{eq:int_1_all_y} \\
    \int \exp \left (\frac{1}{\varepsilon}\left [\feps(x) + \geps(y) - \frac{1}{2}\norm{x-y}^2 \right ]\right ) dQ(y) &= 1 \text{ for all } x \in \mathbb{R}^d \label{eq:int_1_all_x} 
\end{align}
and we will make use of this property many times. This is because $\exp \left (\frac{1}{\varepsilon}\left [\feps(x) + \geps(y) - \frac{1}{2}\norm{x-y}^2 \right ]\right ) dP(x)$ is the conditional density of $\pi_\varepsilon$ given $y$ while $\exp \left (\frac{1}{\varepsilon}\left [\feps(x) + \geps(y) - \frac{1}{2}\norm{x-y}^2 \right ]\right ) dQ(y)$ is the conditional density of $\pi_\varepsilon$ given $x$. For further discussion see \cite{pooladian2021entropic}. 

We also remark that we can also add or subtract a constant $c$ from $\feps$ and $\geps$, that is choose alternate settings $f = \feps + c, g = \geps - c$. As a result of this we can enforce that
\begin{equation*}
    \mathbb{E}_P[\feps(X)] = \mathbb{E}_Q[\geps(Y)] = \frac{1}{2}S_\varepsilon(P,Q)
\end{equation*}
which will be important for the proof of Theorem \ref{thm:subg_conv_rate}.

For brevity we introduce the notations
$$
    c(x,y) \triangleq \frac{1}{2}\norm{x-y}^2 \hspace{0.5cm} \text{and} \hspace{0.5cm}
    \gamma(x,y) \triangleq \exp\left ( \frac{1}{\varepsilon} \left [ \feps(x) + \geps(y) - c(x,y) \right ] \right ).
$$

To start we borrow a result which shows that one can take the supremum in (\ref{eq:ent_dual}) over an even larger space of functions.
\begin{proposition}[\cite{pooladian2021entropic} Proposition 1] \label{prop:eta_dual}
Letting $\pi_\varepsilon$ denote the optimal coupling in (\ref{eq:ent_primal}) between $P$ and $Q$. Then 
\begin{equation*} 
    S_\varepsilon(P,Q) = \sup_{\eta \in L^1(\pi_\varepsilon)} \int \eta(x,y) d\pi_\varepsilon(x,y) - \varepsilon \left ( \int \int \exp\left( \frac{1}{\varepsilon} \left[\eta(x,y) - \frac{1}{2}\norm{x-y}^2 \right ] \right )dQ(y)dP(x) - 1 \right ).
\end{equation*}
\end{proposition}
This is a very useful formula when combined with a clever restriction of the class of test functions. The following result does this by considering functions $\eta$ of the form
$$
\eta(x,y) = \varepsilon\chi(x,y) + \feps(x) + \geps(y),
$$
where $\chi \in L^1(\pi_\varepsilon^n)$. 
\begin{proposition}  \label{prop:dual_prop}
Let $\pi_\varepsilon^n$ be the optimal coupling in (\ref{eq:ent_primal}) between $P$ and $Q^n$ and let $(\feps, \geps)$ be the optimal dual potentials for $(P,Q)$. Then
\begin{align*}
    &\sup_{\chi \in L^1(\pi_\varepsilon^n)} \int \chi(x,y) d\pi_\varepsilon^n(x,y) - \left (\int\int \exp(\chi(x,y))\gamma(x,y) dQ^n(y)dP(x) - 1 \right ) \nonumber \\
    \leq &\frac{1}{\varepsilon}\left(S_\varepsilon(P,Q^n) - S_\varepsilon(P,Q) + \int \geps(y) d(Q-Q^n)(y)\right).
\end{align*}
\end{proposition}
For convenience we define $G \triangleq S_\varepsilon(P,Q^n) - S_\varepsilon(P,Q) + \int \geps(y) d(Q-Q^n)(y).$  Importantly, $G$ is a \textit{random} scalar variable, determined by the random batch of samples observed.

Proposition \ref{prop:dual_prop} is useful for two reasons. The first reason is that both terms on the right hand side have been studied before and can be shown to have good convergence properties as $n$ grows. Indeed, the first term has been analyzed in \cite{pooladian2021entropic} and the second term is easily controlled using standard ideas in Monte Carlo integration since $\geps$ is known to have good regularity properties \cite{genevay2019sample}. Therefore, the right hand side is generally easy to control. The  second reason is that there is still sufficient flexibility for choosing a test functions on the left hand side. In particular we will further restrict the supremum to $\chi$ of the form 
$$\chi(x,y) = h(x)^T(y-\Teps^n(x)) - a||h(x)||^2$$
for $a > 0$ and $h:\mathbb{R}^d \rightarrow \mathbb{R}^d$.  
The motivation for this choice is that we don't know the exact form of $\Teps^n(x) - \Teps(x)$, since $\Teps^n$ is itself random, and instead we aim for uniform control over a function class which surely contains it, regardless of the random samples drawn. It is important to note that since $Q \in\mathcal{P}(B_2^d(0,r))$ it always holds that $\Teps(x) \in B_2^d(0,r)$ and $\Teps^n(x) \in B_2^d(0,r)$. This will ensure that $h$ is a bounded function. The extra variable $a$ allows us to establish a few useful inequalities later.  
\begin{proposition}\label{prop:dual_prop_2}
    Let $\pi_\varepsilon^n$ be the optimal coupling in (\ref{eq:ent_primal}) between $P$ and $Q^n$ and let $(\feps, \geps)$ be the optimal dual potentials for $(P,Q)$. Then for any $a > 0$,
    \begin{align*}
        &\sup_h \int (h(x)^T(y - \Teps^n(x)) - a||h(x)||^2) d\pi_\varepsilon^n(x,y) \nonumber \\
        &- \left (\int\int \exp\left ( h(x)^T(y - \Teps^n(x)) - a||h(x)||^2 \right )\gamma(x,y) dQ^n(y)dP(x)d - 1 \right ) \nonumber \\
        &\leq \frac{1}{\varepsilon}G,
    \end{align*}
    where the supremum is over all $h:\mathbb{R}^d\rightarrow \mathbb{R}^d$ such that $h(x)^T(y - \Teps^n(x)) - a||h(x)||^2 \in L^1(\pi_\varepsilon^n)$.
\end{proposition} 
Essentially the proof is just restricting the supremum in Proposition \ref{prop:dual_prop} to the class of test functions of the form $h(x)^T(y - \Teps^n(x)) - a||h(x)||^2 \in L^1(\pi_\varepsilon^n)$ for some $h$. In principle one could go directly from Proposition \ref{prop:eta_dual} to Proposition \ref{prop:dual_prop_2} by considering functions of the form $\eta = \varepsilon(h(x)^T(y - \Teps^n(x)) - a||h(x)||^2) + \feps(x) + \geps(y)$, however the proof using this approach is incredibly cumbersome.

Importantly, we have managed to insert $\Teps^n$ into the left hand side above. The next result marks a large amount of progress and boils down to a choice of $h$ and a large number of simplifications. 
\begin{lemma} \label{lem:h0_bound}
    Define $h_0:\mathbb{R}^d\rightarrow \mathbb{R}^d$ by $\displaystyle
        h_0(x) = \frac{1}{2a}\left (\Teps^n(x) - \Teps(x) \right ).$
    Then in the setting of Proposition \ref{prop:dual_prop_2} we have
    \begin{equation*}
        \frac{1}{4a}||\Teps^n - \Teps||_{L^2(P)}^2 \leq \frac{1}{\varepsilon}G + \left (\int\int \exp\left (h_0(x)^T(y - \Teps(x)) - a||h_0(x)||^2 \right )\gamma(x,y) dQ^n(y)dP(x) - 1 \right ). 
    \end{equation*}
\end{lemma}
The next step is to swap the trailing ``$-1$'' in the bound above for a term which will make the convergence more explicit. This is done in the following lemma.
\begin{lemma} \label{lem:true_Q_bound}
    Suppose $Q \in \mathcal{P}(B_2^d(0,r))$. Then for $a \geq C_0r^2$ and $h_0$ defined as above it holds that
    \begin{equation*}
        \int\int \exp\left (h_0(x)^T(y - \Teps(x)) - a||h_0(x)||^2 \right )\gamma(x,y) dQ(y) dP(x) \leq 1
    \end{equation*}
    where $C_0$ is an absolute constant.
\end{lemma}
Combining Corollary \ref{lem:true_Q_bound} and Lemma \ref{lem:h0_bound} we obtain the following.
\begin{proposition} \label{proposition:pre_ep} Suppose $Q \in \mathcal{P}(B_2^d(0,r))$. Then for $a \geq C_0r^2$ and $h_0$ defined as above we have
\begin{equation*}
        \frac{1}{4a}||\Teps^n - \Teps||_{L^2(P)}^2 \leq \frac{1}{\varepsilon}G + \int\int \exp\left (h_0(x)^T(y - \Teps(x)) - a||h_0(x)||^2 \right )\gamma(x,y) [dQ^n - dQ](y)dP(x).
\end{equation*}
\end{proposition}

If for a moment one ignores the randomness in $h_0$, the right hand side would be expected to converge to 0 at rate $n^{-1/2}$, since it would be the error in Monte Carlo integration of a single variable function. However, since the function $h_0$ is random one needs to work a bit harder and introduce tools from empirical process theory. 

\begin{lemma} \label{lem:int_bound_subg}
    Assume $P$ is $\sigma^2$-subgaussian and $Q \in \mathcal{P}(B_2^d(0,r))$. Let $C_1 = \max(C_0,2)$. Then with $h_0$ defined as in Lemma \ref{lem:h0_bound} for $a \geq C_1r^2$ we have
    \begin{align*}
        &\mathbb{E} \int \int \exp\left (h_0(x)^T(y - \Teps(x)) - a||h_0(x)||^2 \right )\gamma(x,y) [dQ^n - dQ](y)dP(x) \\
        &\leq \frac{C_2\sqrt{d}}{\sqrt{n}} 
        \exp \left (
            \frac{1}{\varepsilon} \left (  
                \sigma^2 + \sqrt{2 d} \sigma r + r^2
            \right ) 
        \right ) 
        \left ( 
            \exp \left (
                \frac{8rd\sigma^2}{\varepsilon^2}
            \right ) 
            + 2 
        \right )
    \end{align*}
    where $C_2$ is an absolute constant.
\end{lemma}

Having controlled the right most term in Proposition \ref{proposition:pre_ep}, we now turn our attention to controlling $G$. This has already been done in the literature and we state these results for completeness. The first term of $G$ is controlled by the following result.
\begin{proposition}[\cite{mena2019statistical}] \label{prop:mena2019} Let $P,Q$ be $\sigma^2$-subgaussian. Let $\varepsilon > 0$. Then
\begin{equation*}
    \mathbb{E}S_\varepsilon(P,Q^n) - S_\varepsilon(P,Q) \leq K_{d,0} \cdot \varepsilon \left ( 1 + \frac{\sigma^{\lceil 5d / 2 \rceil + 6}}{\varepsilon^{\lceil 5d/4 \rceil + 3}} \right )\frac{1}{\sqrt{n}}.
\end{equation*}
\end{proposition}
Note that if $\sigma \geq \frac{r}{\sqrt{2d \log 2}}$ and $Q \in \mathcal{P}(B_2^d(0,r))$ then $Q$ always satisfies the condition of the Proposition \ref{prop:mena2019}, as can be shown by a direct calculation. 

For the other term we have simply that
\begin{equation*}
    \mathbb{E}\int \geps(y) d(Q-Q^n)(y) = 0.
\end{equation*}
We can now state the ``zero-to-one'' sample theorem for the convergence of the entropic map, which follows by combining Proposition \ref{proposition:pre_ep}, Lemma \ref{lem:int_bound_subg}, and Proposition \ref{prop:mena2019}.

\begin{theorem} \label{thm:0_to_1_subg}
      Assume $P$ is $\sigma^2$-subgaussian and $Q \in \mathcal{P}(B_2^d(0,r))$. Then
     \begin{equation*}
         \mathbb{E}||\Teps^n - \Teps||_{L^2(P)}^2 \leq 
         \exp \left (
            \frac{1}{\varepsilon} 
            \left (  \sigma^2 + \sqrt{2 d} \sigma r + r^2 \right ) \right ) \left ( \exp \left ( \frac{8rd\sigma^2}{\varepsilon^2}\right ) + 2 \right )\frac{C_3r^2\sqrt{d}}{\sqrt{n}} +  \left ( 1 + \frac{\sigma^{\lceil 5d / 2 \rceil + 6}}{\varepsilon^{\lceil 5d/4 \rceil + 3}} \right )\frac{K_{d,1}r^2}{\sqrt{n}}.
     \end{equation*}
\end{theorem}

\begin{proof}
    Taking expectations and applying the bounds in Proposition \ref{proposition:pre_ep} followed by Lemma \ref{lem:int_bound_subg} Proposition \ref{prop:mena2019} we obtain
    \begin{align*}
        &\mathbb{E} \frac{1}{4a}||\Teps^n - \Teps||_{L^2(P)}^2 \\
        &\leq\mathbb{E} \left [ \frac{1}{\varepsilon}G + \int\int \exp\left (h_0(x)^T(y - \Teps(x)) - a||h_0(x)||^2 \right )\gamma(x,y) [dQ^n - dQ](y)dP(x) \right ] \\
        &\leq \frac{1}{\varepsilon} \left (  K_{d,0} \cdot \varepsilon \left ( 1 + \frac{\sigma^{\lceil 5d / 2 \rceil + 6}}{\varepsilon^{\lceil 5d/4 \rceil + 3}} \right )\frac{1}{\sqrt{n}} \right ) +  \frac{C_2\sqrt{d}}{\sqrt{n}} 
        \exp \left (
            \frac{1}{\varepsilon} \left (  
                \sigma^2 + \sqrt{2 d} \sigma r + r^2
            \right ) 
        \right ) 
        \left ( 
            \exp \left (
                \frac{8rd\sigma^2}{\varepsilon^2}
            \right ) 
            + 2 
        \right ) \\
        &=
          K_{d,0}  \left ( 1 + \frac{\sigma^{\lceil 5d / 2 \rceil + 6}}{\varepsilon^{\lceil 5d/4 \rceil + 3}} \right )\frac{1}{\sqrt{n}} +  \frac{C_2\sqrt{d}}{\sqrt{n}} 
        \exp \left (
            \frac{1}{\varepsilon} \left (  
                \sigma^2 + \sqrt{2 d} \sigma r + r^2
            \right ) 
        \right ) 
        \left ( 
            \exp \left (
                \frac{8rd\sigma^2}{\varepsilon^2}
            \right ) 
            + 2 
        \right ) 
    \end{align*}
    Choosing $a = C_1r^2$ and multiplying by both sides proves the result with $K_{d,1} = 4C_1K_{d,0}$ and $C_3 = 4C_1C_2$.
\end{proof}

A much easier proof is also possible for the ``one-to-two'' sample theorem for the convergence of the entropic map.
\begin{theorem} \label{thm:1_to_2_subg}Let $P$ and $Q$ be $\sigma^2$-subgaussian and 
    let $\Teps^{n,n}$ be the entropic map from $P^n$ to $Q^n$, and let $\Teps^n$ be the entropic map from $P$ to $Q^n$. Then $\Teps^{n,n}$ satisfies
    \begin{equation*}
        \mathbb{E}\norm{\Teps^n - \Teps^{n,n}}_{L^2(P)}^2 \leq r^2  K_{d,2} \cdot \left ( 1 + \frac{\sigma^{\lceil 5d/2\rceil + 6}}{\varepsilon^{\lceil 5d / 4 \rceil + 3}}\right ) \frac{1}{\sqrt{n}}.
    \end{equation*}
\end{theorem}
As noted above, this result automatically covers bounded measures since all bounded measures are $\sigma^2$-subgaussian with $\sigma^2$ controlled by the radius of the ball containing the support of $Q$. 

Combining Theorems \ref{thm:0_to_1_subg} and \ref{thm:1_to_2_subg} we have our result.
\begin{proof} (Theorem \ref{thm:subg_conv_rate})
    \noindent \begin{align}
        \mathbb{E} ||\Teps^{n,n} - \Teps||_{L^2(P)}^2 &= \mathbb{E} ||(\Teps^{n,n} - \Teps^n) + (\Teps^n - \Teps)||_{L^2(P)}^2 \nonumber \\
        &\leq \mathbb{E} \left ( ||(\Teps^{n,n} - \Teps^n)||_{L^2(P)} + ||(\Teps^n - \Teps)||_{L^2(P)} \right )^2 \nonumber \\
        &\leq 2\mathbb{E}||(\Teps^{n,n} - \Teps^n)||_{L^2(P)}^2 + 2\mathbb{E}||(\Teps^n - \Teps)||_{L^2(P)}^2 \nonumber \\
        &\leq \frac{2C_3r^2\sqrt{d}}{\sqrt{n}}
         \exp \left (
            \frac{1}{\varepsilon} 
            \left (  \sigma^2 + \sqrt{2 d} \sigma r + r^2 \right ) \right ) \left ( \exp \left ( \frac{8rd\sigma^2}{\varepsilon^2}\right ) + 2 \right )  \nonumber \\
        & \hspace{2cm} + 2(K_{d,1}+K_{d,2})r^2 \left ( 1 + \frac{\sigma^{\lceil 5d / 2 \rceil + 6}}{\varepsilon^{\lceil 5d/4 \rceil + 3}} \right )\frac{1}{\sqrt{n}}. \label{eq:b1_expression}
    \end{align}
\end{proof}

\subsection{Proof of Proposition \ref{prop:dual_prop}}
\begin{proof}
    Let $\chi \in L^1(\pi_\varepsilon^n)$ and define $\eta$ by 
    \begin{equation*}
        \eta(x,y) = \varepsilon\chi(x,y) + \feps(x) + \geps(y).
    \end{equation*}
    Note that $\eta \in L^1(\pi_\varepsilon^n)$ since $\chi, (\feps + \geps) \in L^1(\pi_\varepsilon^n)$. By Proposition \ref{prop:eta_dual} and the fact that $P,Q^n$ are the marginals of $\pi_\varepsilon^n$ it follows
    \begin{align*}
        S_\varepsilon(P,Q^n) &\geq \int \eta(x,y) d\pi_\varepsilon^n(x,y) - \varepsilon \left ( \int \int \exp\left ( \frac{1}{\varepsilon}\left [\eta(x,y) - \frac{1}{2}\norm{x-y}^2 \right ] \right ) dQ^n(y)dP(x) - 1 \right ) \\
        &= \int \left [ \varepsilon\chi(x,y) + \feps(x) + \geps(y) \right ] d\pi_\varepsilon^n(x,y) \nonumber \\
        &\hspace{1cm}- \varepsilon \left ( \int \int \exp\left ( \frac{1}{\varepsilon}\left [\left[\varepsilon\chi(x,y) + \feps(x) + \geps(y)\right ] - \frac{1}{2}\norm{x-y}^2 \right ] \right ) dQ^n(y)dP(x) - 1 \right ) \\
        &= \varepsilon \int \chi(x,y) d\pi_\varepsilon^n(x,y) + \int \feps(x) dP(x) + \int \geps(y) dQ^n(y) \nonumber \\
        &\hspace{1cm} - \varepsilon \left ( \int \int \exp\left (\chi(x,y) + \frac{1}{\varepsilon}\left[\feps(x) + \geps(y) - \frac{1}{2}\norm{x-y}^2\right ] \right ) dQ^n(y)dP(x) - 1 \right ) \\
        &= \varepsilon \int \chi(x,y) d\pi_\varepsilon^n(x,y) + \int \feps(x) dP(x) + \int \geps(y) dQ^n(y) \nonumber \\
        &\hspace{1cm} - \varepsilon \left ( \int \int \exp(\chi(x,y))\gamma(x,y) dQ^n(y)dP(x) - 1 \right ).
    \end{align*}
    Rearranging we have 
    \begin{align}
        &\int \chi(x,y) d\pi_\varepsilon^n(x,y) - \left ( \int  \int \exp(\chi(x,y))\gamma(x,y) dQ^n(y)dP(x) - 1 \right ) \nonumber  \\
        \leq &\frac{1}{\varepsilon} S_\varepsilon(P,Q^n) - \frac{1}{\varepsilon} \left ( \feps(x) dP(x) + \int \geps(y) dQ^n(y) \right ). \label{eq:chi_bound_1}
    \end{align}
    Dropping for a moment the $\frac{1}{\varepsilon}$ we have 
    \begin{align}
        &S_\varepsilon(P,Q^n) - \left ( \int \feps(x) dP(x) + \int \geps(y) dQ^n(y) \right ) \nonumber \\
        =& (S_\varepsilon(P,Q^n) - S_\varepsilon(P,Q)) + \left ( S_\varepsilon(P,Q) - \int \feps(x) dP(x) - \int \geps(y) dQ(y) \right ) + \int \geps(y) d(Q - Q^n)(y) \nonumber \\
        =& S_\varepsilon(P,Q^n) - S_\varepsilon(P,Q) + \int \geps(y) d(Q - Q^n)(y). \label{eq:Qn_int_trick}
    \end{align}
    In the last line we have used that by (\ref{eq:int_1_all_y})
    \begin{equation*}
        - \varepsilon\int \int \exp \left (\frac{1}{\varepsilon}\left [\feps(x) + \geps(y) - \frac{1}{2}\norm{x-y}^2 \right ]\right ) dQ(y) dP(x)  + \varepsilon = -\varepsilon \int 1 dQ(y) + \varepsilon = 0
    \end{equation*}
    and therefore 
    \begin{align*}
        S_\varepsilon(P,Q) &= \int \feps dP + \int \geps dQ - \varepsilon\int \int \exp \left (\frac{1}{\varepsilon}\left [\feps(x) + \geps(y) - \frac{1}{2}\norm{x-y}^2 \right ]\right ) dQ(y)dP(x)  + \varepsilon \\
        &= \int \feps dP + \int \geps dQ 
    \end{align*}
    which implies 
    \begin{equation*}
        S_\varepsilon(P,Q) - \int \feps dP - \int \geps dQ  = 0.
    \end{equation*}
    Plugging (\ref{eq:Qn_int_trick}) into (\ref{eq:chi_bound_1}) we obtain
    \begin{align*}
        &\int \chi(x,y) d\pi_\varepsilon^n(x,y) - \left ( \int \int \exp(\chi(x,y))\gamma(x,y) dQ^n(y)dP(x) - 1 \right ) \\
        &\leq \frac{1}{\varepsilon} \left [  S_\varepsilon(P,Q^n) - S_\varepsilon(P,Q) + \int \geps(y) d(Q - Q^n)(y) \right ]. 
    \end{align*}
    Since $\chi$ was chosen arbitrarily in $L^1(\pi_\varepsilon^n),$ this bound holds uniformly over the class. Therefore we have
    \begin{align*}
        &\sup_{\chi \in L^1(\pi_\varepsilon^n)} \int \chi(x,y) d\pi_\varepsilon^n(x,y) - \left (\int\int \exp(\chi(x,y))\gamma(x,y) dQ^n(y)dP(x) - 1 \right ) \\
    &\leq \frac{1}{\varepsilon}(S_\varepsilon(P,Q^n) - S_\varepsilon(P,Q)) + \frac{1}{\varepsilon}\int \geps(y) d(Q-Q^n)(y)
    \end{align*}
    as desired.
\end{proof}

\subsection{Proof of Proposition \ref{prop:dual_prop_2}} 
\begin{proof}
    Define the set 
    \begin{equation*}
        \mathcal{H} = \left \{ H(x,y) = h(x)^T(y - \Teps^n(x)) - a||h(x)||^2 \hspace{0.05cm} \bigg | \hspace{0.05cm} h:\mathbb{R}^d \rightarrow \mathbb{R}^d , \int \left | h(x)^T(y - \Teps^n(x)) - a||h(x)||^2\right | d\pi_\varepsilon^n < \infty \right \}.
    \end{equation*}
    Then $\mathcal{H} \subset L^1(\pi_\varepsilon^n)$ and by Proposition \ref{prop:dual_prop} we have
    \begin{align*} 
         &\sup_{h} \int h(x)^T(y - \Teps^n(x)) - a||h(x)||^2 d\pi_\varepsilon^n(x,y) \nonumber \\
         & - \left (\int\int \exp\left ( h(x)^T(y - \Teps^n(x)) - a||h(x)||^2 \right )\gamma(x,y) dQ^n(y)dP(x) - 1 \right ) \\
        &= \sup_{H \in \mathcal{H}} \int H(x,y) d\pi_\varepsilon^n(x,y) - \left (\int\int \exp(H(x,y) )\gamma(x,y) dQ^n(y)dP(x) - 1 \right ) \\
        &\leq \sup_{\chi \in L^1(\pi_\varepsilon^n)} \int \chi(x,y) d\pi_\varepsilon^n(x,y) - \left (\int\int \exp(\chi(x,y))\gamma(x,y) dQ^n(y)dP(x) - 1 \right ) \\
        &\leq \frac{1}{\varepsilon}(S_\varepsilon(P,Q^n) - S_\varepsilon(P,Q)) + \frac{1}{\varepsilon}\int \geps(y) d(Q-Q^n)(y) = \frac{1}{\varepsilon}G.
    \end{align*}
\end{proof}

\subsection{Proof of Lemma \ref{lem:h0_bound}}
\begin{proof}
Using that the first marginal of $\pi_\varepsilon^n$ is $P$ we have
\begin{equation}
    \int (h_0(x)^T(y - \Teps(x)) - a||h_0(x)||^2) d\pi_\varepsilon^n(x,y) = \int h_0(x)^T(y - \Teps(x)) d\pi_\varepsilon^n(x,y) - \int a ||h_0(x)||^2 dP(x). \label{eq:zeroth_int}
\end{equation}
Now let us consider the two integrals separately. For the first we have the following chain.
\begin{align}
    &\int h_0(x)^T(y - \Teps(x)) d\pi_\varepsilon^n(x,y)\nonumber \\
    &= \int \left (\frac{1}{2a}(\Teps^n(x) - \Teps(x))\right )^T(y - \Teps(x)) d\pi_\varepsilon^n(x,y) & \nonumber \\
    &=\frac{1}{2a} \int  (\Teps^n(x) - \Teps(x))^Ty  d\pi_\varepsilon^n(x,y) - \frac{1}{2a}\int (\Teps^n(x) - \Teps(x))^T\Teps(x)d\pi_\varepsilon^n(x,y) \nonumber \\
    &= \frac{1}{2a} \int  (\Teps^n(x) - \Teps(x))^Ty  d\pi_\varepsilon^n(x,y) - \frac{1}{2a}\int (\Teps^n(x) - \Teps(x))^T\Teps(x)dP(x)  \nonumber \\
    &= \frac{1}{2a} \int  (\Teps^n(x) - \Teps(x))^T\Teps^n(x)dP(x) - \frac{1}{2a}\int (\Teps^n(x) - \Teps(x))^T\Teps(x)dP(x) \label{eq:hard_step} \\
    &= \frac{1}{2a} \int  (\Teps^n(x) - \Teps(x))^T(\Teps^n(x) - \Teps(x))  dP(x)  \nonumber \\
    &= \frac{1}{2a} \int  ||\Teps^n(x) - \Teps(x)||^2 dP(x) = \frac{1}{2a} ||\Teps^n - \Teps||_{L^2(P)}^2.  \label{eq:first_int}
\end{align}
The third equality uses that $P$ is the marginal of $\pi_\varepsilon^n$. To see (\ref{eq:hard_step}), note
\begin{align*}
    \int  (\Teps^n(x) - \Teps(x))^Ty  d\pi_\varepsilon^n(x,y) &= \int  \left [ \int (\Teps^n(x) - \Teps(x))^Ty d\pi_\varepsilon^n(y|x) \right ]  dP(x) \\
    &= \int  (\Teps^n(x) - \Teps(x))^T\left [ \int y d\pi_\varepsilon^n(y|x) \right ]  dP(x) & (\text{Linearity}) \\
    &= \int  (\Teps^n(x) - \Teps(x))^T \Teps^n(x)  dP(x). & (\text{Eq} \ref{eq:entropic_map})
\end{align*}
For the second integral we have
\begin{align}
    \int a ||h_0(x)||^2 dP(x) &= \int a ||\frac{1}{2a}[\Teps^n(x) - \Teps(x)]||^2 dP(x) \nonumber \\
    &= \frac{1}{4a} ||\Teps^n - \Teps||_{L^2(P)}^2 \label{eq:second_int}
\end{align}
Plugging (\ref{eq:first_int}),(\ref{eq:second_int}) into (\ref{eq:zeroth_int}) we have
\begin{equation*}
    \int h_0(x)^T(y - \Teps(x)) - a||h(x)||^2 d\pi_\varepsilon^n(x,y) = \frac{1}{2a}||\Teps^n - \Teps||_{L^2(P)}^2 - \frac{1}{4a}||\Teps^n - \Teps||_{L^2(P)}^2 = \frac{1}{4a}||\Teps^n - \Teps||_{L^2(P)}^2. 
\end{equation*}
Now by Proposition \ref{prop:dual_prop_2} we have
\begin{align*}
    \frac{1}{\varepsilon} G 
    &\geq \sup_{h}  \int h(x)^T(y - \Teps(x)) - a||h(x)||^2 d\pi_\varepsilon^n(x,y) \nonumber 
    \\ 
    & \hspace{1cm}-  \left (\int\int \exp\left (h(x)^T(y - \Teps(x)) - a||h(x)||^2 \right )\gamma(x,y) dQ^n(y)dP(x) - 1 \right ) \\
    &\geq \int h_0(x)^T(y - \Teps(x)) - a||h_0(x)||^2 d\pi_\varepsilon^n(x,y) \nonumber 
    \\ 
    & \hspace{1cm}-  \left (\int\int \exp\left (h_0(x)^T(y - \Teps(x)) - a||h_0(x)||^2 \right )\gamma(x,y) dQ^n(y)dP(x) - 1 \right ) \\
    &= \frac{1}{4a}||\Teps^n - \Teps||_{L^2(P)}^2 - \left (\int\int \exp\left (h_0(x)^T(y - \Teps(x)) - a||h_0(x)||^2 \right )\gamma(x,y) dQ^n(y)dP(x) - 1 \right ).
\end{align*}
If we re-arrange the first and last inequality we have
\begin{equation*}
    \frac{1}{4a}||\Teps^n - \Teps||_{L^2(P)}^2 \leq \frac{1}{\varepsilon}G + \left (\int\int \exp\left (h_0(x)^T(y - \Teps(x)) -  a||h_0(x)||^2 \right )\gamma(x,y) dQ^n(y)dP(x) - 1 \right )
\end{equation*}
which proves the result.
\end{proof}

\subsection{Proof of Lemma \ref{lem:true_Q_bound}} 

In order to prove this result we first collect two further facts. The first is proved by a direct calculation.

\begin{lemma} \label{lem:h_int_0}
    In the setting of Proposition \ref{prop:dual_prop_2}, for any $h$ and any $x$ we have
    \begin{equation*}
        \int h(x)^T(y - \Teps(x))\gamma(x,y)dQ(y) = 0.
    \end{equation*}
\end{lemma} 

\begin{proof}
By linearity we have
\begin{equation*}
     \int h(x)^T(y - \Teps(x))\gamma(x,y)dQ(y) = h(x)^T \left [\int (y - \Teps(x))\gamma(x,y) dQ(y)\right ]
\end{equation*}
and the integral on the inside can be expressed as 
\begin{align*}
    &\int (y - \Teps(x))\gamma(x,y) dQ(y) \\
    &= \int y\gamma(x,y) - \gamma(x,y)\left [\frac{
            \displaystyle\int y_0 \exp\left (\frac{1}{\varepsilon}\left [\geps(y_0) - c(x,y_0) \right ] \right ) dQ(y_0)
        }{
            \displaystyle\int \exp\left (\frac{1}{\varepsilon}\left [\geps(y_1) - c(x,y_1) \right ] \right ) dQ(y_1)
        } \right ] dQ(y) &(\text{Def of } \Teps) \\
    &= \int y\gamma(x,y) - \gamma(x,y)\left [\frac{
            \displaystyle\int y_0 \exp\left (\frac{1}{\varepsilon}\left [\geps(y_0) + \feps(x) - c(x,y_0) \right ] \right ) dQ(y_0)
        }{
            \displaystyle\int \exp\left (\frac{1}{\varepsilon}\left [\geps(y_1) + \feps(x) - c(x,y_1) \right ] \right ) dQ(y_1)
        } \right ] dQ(y) \\
    &= \int y\gamma(x,y) - \gamma(x,y)\left [
            \int y_0 \exp\left (\frac{1}{\varepsilon}\left [\geps(y_0) + \feps(x) - c(x,y_0) \right ] \right ) dQ(y_0)
        \right ] dQ(y) & (\text{Eq. } \ref{eq:int_1_all_x}) \\
    &= \int y\gamma(x,y)dQ(y) - \int\gamma(x,y)\left [ \int y_0 \gamma(x,y_0) dQ(y_0)\right ]dQ(y) \\
    &= \int y\gamma(x,y)dQ(y) - \left [ \int y_0 \gamma(x,y_0) dQ(y_0)\right ]\int \gamma(x,y)dQ(y) \\
    &= \int y\gamma(x,y)dQ(y) -  \int y_0 \gamma (x,y_0) dQ(y_0) = 0. & (\text{Eq \ref{eq:int_1_all_x}})
\end{align*}
\end{proof}

In particular this result holds when $h = h_0$, and going a step further, it holds \textit{independent} of both the choice of $x$ and the form of the random function $h_0$. This is critical for establishing uniform control. The mean-zero condition also enables us to use a key inequality which only holds for mean-zero random variables.

\begin{lemma} \label{lem:a_bound}
    There exists an absolute constant $C_0$ such that for all $a \geq C_0r^2$ and $h_0$ defined as in Lemma \ref{lem:h0_bound}, it holds for all $x$ that
    \begin{equation*}
        \int \exp(h_0(x)^T(y - \Teps(x)) \gamma(x,y) dQ(y) \leq \exp(a||h_0(x)||^2).
    \end{equation*}
\end{lemma}

\begin{proof}
    We need a few facts from \cite{vershynin2018high}. First, if $X$ is a bounded random variable such that $||X||_\infty < B$ then 
    \begin{equation*}
        ||X||_{\psi_2} \leq C_5 B
    \end{equation*}
    where $||X||_{\psi}$ denotes the subgaussian norm of $X$ and $C_5$ is an absolute constant (see Example 2.5.8.iii in \cite{vershynin2018high}). Second, if $X$ is mean-zero, then for all $\lambda \in \mathbb{R}$,
    \begin{equation}
        \mathbb{E}\exp(\lambda X) \leq \exp(C_4\lambda^2||X||_{\psi_2}^2) \label{eq:hoeffding_type}
    \end{equation}
    where $C_4$ is another absolute constant
    (See Proposition 2.5.2 or (2.16) in \cite{vershynin2018high}). 
    
    Now for a fixed $x$, let $Y^x$ be the random variable whose law is the conditional distribution of $\pi_\varepsilon$ with $X = x$. Further define
    \begin{equation*}
        Z^x \triangleq (\Teps^n(x) - \Teps(x))^T(Y^x - \Teps(x)) = 2ah_0(x)^T(Y^x - \Teps(x)).
    \end{equation*}
    First note that since $\Teps(x),\Teps^n(x),Y^x \in B_2^d(0,r)$ we have
    \begin{align}
        |Z^x| &= |(\Teps^n(x) - \Teps(x))^T(Y^x - \Teps(x))| \nonumber \\
        &\leq ||\Teps^n(x) - \Teps(x)||||Y^x - \Teps(x)|| \nonumber \\
        &\leq (2r)||\Teps^n(x) - \Teps(x)||  \nonumber \\
        \implies ||Z^x||_{\psi_2} &\leq 2rC_5||\Teps^n(x) - \Teps(x)||. \label{eq:sg_norm_bound}
    \end{align}
    Next, we have by Lemma \ref{lem:h_int_0} 
    \begin{align*}
        \mathbb{E}[Z^x] &= \int (\Teps^n(x) - \Teps(x))^T(y - \Teps(x)) d\pi_\varepsilon^x(y) \\
        &= (\Teps^n(x) - \Teps(x))^T\left (\int yd\pi_\varepsilon^x(y) - \Teps(x)\right ) \\
        &= (\Teps^n(x) - \Teps(x))^T\left (\Teps(x) - \Teps(x)\right ) = 0.
    \end{align*}
    This shows that $Z^x$ satisfies the conditions to use (\ref{eq:hoeffding_type}). Doing so we obtain
    \begin{align*}
        &\int \exp  (h_0(x))^T(y - \Teps(x)) \gamma(x,y) dQ(y) \\
        = &\int \exp \left (\frac{1}{2a} (\Teps^n(x) - \Teps(x))^T(y - \Teps(x))  \right ) \gamma(x,y) dQ(y) \nonumber \\
        = &\mathbb{E}\left[\exp \left (\frac{1}{2a} Z^x \right ) \right ] \nonumber \\
        \leq &\exp \left ( C_4 \frac{1}{4a^2} 4C_5^2r^2||\Teps^n(x) - \Teps(x)||^2 \right ) & (\text{Eq.} (\ref{eq:hoeffding_type}) \text{ and } (\ref{eq:sg_norm_bound})) \\
        = &\exp \left ( C_6 \frac{r^2}{a^2} ||\Teps^n(x) - \Teps(x)||^2 \right ) 
    \end{align*}
    where $C_6 = C_5^2C_4$.
    
    From here it is enough to show that
    \begin{equation*}
        \exp \left ( C_6 \frac{r^2}{a^2} ||\Teps^n(x) - \Teps(x)||^2 \right ) \leq  \exp \left (\frac{1}{4a}||\Teps^n(x) - \Teps(x)||^2 \right ) = \exp(a||h_0(x)||^2).
    \end{equation*}
    By monotonicity of $\exp$ it is enough to show that 
    \begin{equation*}
        C_6 \frac{r^2}{a^2} ||\Teps^n(x) - \Teps(x)||^2 \leq \frac{1}{4a}||\Teps^n(x) - \Teps(x)||^2
    \end{equation*}
    and this holds true if $a \geq 4C_6r^2$. Letting $C_0 = 4C_6$ proves the result.
\end{proof}

With the preceding lemma in hand, the proof of Lemma \ref{lem:true_Q_bound} becomes straightforward.
\begin{proof} (Lemma \ref{lem:true_Q_bound})
    For $a \geq C_0r^2$ we have
    \begin{align*}
        &\int\int \exp\left (h_0(x)^T(y - \Teps(x)) - a||h_0(x)||^2 \right )\gamma(x,y) dQ(y) dP(x) \\
        =& \int \exp(-a||h_0(x)||^2) \left [ \int \exp\left (h_0(x)^T(y - \Teps(x)) \right )\gamma(x,y) dQ(y) \right ] dP(x) \\
        \leq& \int \exp(-a||h_0(x)||^2) \left [ \exp(a||h_0(x)||^2) \right ] dP(x) & (\text{Lemma \ref{lem:a_bound}}) \\
        =& 1.
    \end{align*}
\end{proof}

\subsection{Proof of Lemma \ref{lem:int_bound_subg}}

To start we have the following result, which is essentially contained in Proposition A.1 of \cite{mena2019statistical}. We include its proof here for completeness and to specify the bounds to our setting.
\begin{proposition} 
\label{prop:mena_weed19} 
Let $P$ be a $\sigma^2$-subgaussian. Let $Q \in \mathcal{P}(B_2^d(0,r))$. Then, there exist smooth optimal potentials $(\feps, \geps)$ for $S_\varepsilon(P,Q)$ such that, 
\begin{align*}
    \geps(y) - \frac{1}{2}\| x - y \|^2 \leq d \sigma^2 + (\|x\| + \sqrt{2} d \sigma)\|y\| - \|x\|^2,
\end{align*}
\begin{align*}
    f_{\varepsilon}(x) \leq \frac{1}{2} ( r + \|x\|)^2.
\end{align*}
\end{proposition}
\begin{proof}
We note that one can chose $\feps, \geps$ such that 
\begin{align*}
    \feps(x) = - \varepsilon \log \left( \mathbb{E} \exp \left(  \frac{1}{\varepsilon}(\geps(Y) - 1/2 \| x - Y\|^2 )  \right) \right) \\
    \geps(y) = - \varepsilon \log \left( \mathbb{E} \exp \left(  \frac{1}{\varepsilon}(\feps(X) - 1/2 \| X - y\|^2 )  \right) \right) 
\end{align*}
and $\mathbb{E}[\feps(X)] \geq 0, \mathbb{E}[\geps(X)] \geq 0$. Given these choices, we note that by the convexity of $-\varepsilon \log$ and Jensen's inequality that
\begin{align*} 
    \geps(y)  &=  - \varepsilon \log \left( \mathbb{E} \exp \left(  \frac{1}{\varepsilon}(\feps(X) - 1/2 \| X - y\|^2 )  \right) \right)  \\
    &\leq \mathbb{E} -\varepsilon \log \exp \left ( \frac{1}{2} \left(  \frac{1}{\varepsilon}(\feps(X) - 1/2 \| X - y\|^2 )  \right) \right) \\
    &= \mathbb{E} \frac{1}{2}\|X - y\|^2 - \feps(X) \\
    &\leq \frac{1}{2}\mathbb{E} \|X - y\|^2.
\end{align*}
This implies that 
\begin{align*}
     \geps(y) - \frac{1}{2} \|x - y\|^2 & \leq \frac{1}{2} \mathbb{E} \|X - y\|^2 - \frac{1}{2} \|x - y\|^2\\
     & = \frac{1}{2} (\mathbb{E} \|X\|^2 +  \|y\|^2) - \mathbb{E}[X]^\top y - \frac{1}{2} (\|x\|^2 + \|y\|^2) + x^\top y\\
     & \leq d \sigma^2 + \sqrt{2 d} \sigma \|y\|  - \frac{\|x\|^2}{2} + \|x\| \|y\|.
\end{align*}
Similarly, applying Jensen's inequality as above but to $\feps$ we obtain, 
\begin{align*}
    \feps(x)  \leq \frac{1}{2} \mathbb{E} \|Y - x\|^2 \leq \frac{1}{2} (r + \|x\|)^2.
\end{align*}
\end{proof}

For our analysis this proposition implies the following bound on $\gamma(x,y)$ for $P$ a.e. $x$ and for $Q$ a.e. $y$. 
\begin{lemma}
\label{lem:subg_ub_gamma}
Under the assumptions of Proposition \ref{prop:mena_weed19},
\begin{align*}
     \gamma(x,y) & \leq \exp \left (\frac{1}{\varepsilon} \left [ d \sigma^2 + \sqrt{2 d} \sigma r + r^2 +  2r\|x\| \right ] \right ).
\end{align*}
\end{lemma}
\begin{proof}
    Using Proposition \ref{prop:mena_weed19} we have
    \begin{align*}
    \varepsilon \log \gamma(x,y) & = \feps(x) + \geps(y) - \frac{1}{2} \| x - y\|^2 \\ 
    & \leq d \sigma^2 + \sqrt{2 d} \sigma \|y\|  - \frac{\|x\|^2}{2} + \|x\| \|y\| + \frac{1}{2} (r + \|x\|)^2 \\
    & = d \sigma^2 + \sqrt{2 d} \sigma r + \|x\|r + \frac{1}{2}(r^2 + 2 \|x\| r) \\
    & = d \sigma^2 + \sqrt{2 d} \sigma r + r^2 + 2r\|x\|.
\end{align*}
Dividing by $\varepsilon$ and exponentiating proves the result.
\end{proof}

Before continuing to the main proof we require one more basic tool. The following result is a standard bound in empirical process theory which will help us to control the randomness of $h_0$. To state it we must introduce the notation $\mathcal{N}(T,d_T,\delta)$ which is the covering number of the metric space $(T,d_T)$ by balls of radius at most $\delta$ whose centers lie in $T$.

\begin{theorem}[Dudley's Inequality, \cite{vershynin2018high} Theorem 8.1.3] \label{thm:hdp_chain}
    Let $(X_t)_{t\in T}$ be a mean-zero random process on a metric space $(T,d_T)$ with subgaussian increments satisfying $||X_t - X_s||_{\psi_2} \leq Kd_T(t,s)$ for all $t,s \in T$. Then
    \begin{equation*}
        \mathbb{E}\sup_{t \in T} X_t \leq C_7K \int_0^\infty \sqrt{\log \mathcal{N}(T,d_T,\delta)} d\delta.
    \end{equation*}
\end{theorem}

We can now proceed to the proof of Lemma \ref{lem:int_bound_subg}. 
\begin{proof}
    
    First we re-write our integral in a form that is more convenient for us.
    \begin{align*}
        &\int\int \exp\left (h_0(x)^T(y - \Teps(x)) - a||h_0(x)||^2 \right )\gamma(x,y) [dQ^n - dQ](y)dP(x) \nonumber \\
        =& \int\int \exp\left ( \frac{1}{2a}\left ( \Teps^n(x) - \Teps(x) \right )^T(y - \Teps(x)) - \frac{1}{4a}||\Teps^n(x) - \Teps(x)||^2 \right )\gamma(x,y)[dQ^n - dQ](y)dP(x).
    \end{align*}
    Now note that $T_\varepsilon^n(x)$ is a vector contained in $B_2^d(0,r)$ so we always have the following bound (since we can choose $v = T_\varepsilon^n(x)$)
    \begin{align*}
        &\int\int \exp\left ( \frac{1}{2a}\left ( \Teps^n(x) - \Teps(x) \right )^T(y - \Teps(x)) - \frac{1}{4a}||\Teps^n(x) - \Teps(x)||^2 \right )\gamma(x,y)[dQ^n - dQ](y)dP(x) \nonumber \\
        \leq &\int \left [ \sup_{v \in B_2^d(0,r)} \int \exp\left ( \frac{1}{2a}\left ( v - \Teps(x) \right )^T(y - \Teps(x)) - \frac{1}{4a}||v - \Teps(x)||^2 \right )\gamma(x,y)[dQ^n - dQ](y) \right ] dP(x)
    \end{align*}
    Taking the expectation of both sides of this inequality, and subsequently changing the order of integration we have
    \begin{align*}
        &\mathbb{E}_{Y^n} \int\int \exp\left (h_0(x)^T(y - \Teps(x)) - a||h_0(x)||^2 \right )\gamma(x,y) [dQ^n - dQ](y)dP(x) \nonumber \\ 
        \leq& \mathbb{E}_{Y^n} \int  \left [ \sup_{v \in B_2^d(0,r)} \int \exp\left ( \frac{1}{2a}\left ( v - \Teps(x) \right )^T(y - \Teps(x)) - \frac{1}{4a}||v - \Teps(x)||^2 \right )\gamma(x,y)[dQ^n - dQ](y) \right ] dP(x) \\
        =& \int \mathbb{E}_{Y^n} \left [ \sup_{v \in B_2^d(0,r)} \int \exp\left ( \frac{1}{2a}\left ( v - \Teps(x) \right )^T(y - \Teps(x)) - \frac{1}{4a}||v - \Teps(x)||^2 \right )\gamma(x,y)[dQ^n - dQ](y) \right ] dP(x)
    \end{align*}
    Therefore it is enough to uniformly bound over $x$ the inner expectation, which is what we do now. From here, $x$ is treated as fixed.
    
    Recall that $Q^n = \frac{1}{n} \sum_{i=1}^n \delta_{Y_i}$ where $Y_1,...,Y_n$ are i.i.d. according to $Q$. We now define the same empirical process $(Z_v)_{v \in B_2^d(0,r)}$ where $Z_v$ is defined by
    \begin{align*}
        Z_v &\triangleq \int \exp\left ( \frac{1}{2a}\left ( v - \Teps(x) \right )^T(y - \Teps(x)) - \frac{1}{4a}||v - \Teps(x)||^2 \right )\gamma(x,y)[dQ^n - dQ](y) \\
        &= \frac{1}{n}\sum_{i=1}^n  \exp\left ( \frac{1}{2a}\left ( v - \Teps(x) \right )^T(Y_i - \Teps(x)) - \frac{1}{4a}||v - \Teps(x)||^2 \right )\gamma(x,Y_i) \\
        & \hspace{1cm} - \mathbb{E}_Y \exp\left ( \frac{1}{2a}\left ( v - \Teps(x) \right )^T(Y - \Teps(x)) - \frac{1}{4a}||v - \Teps(x)||^2 \right )\gamma(x,Y) \\
        &= \frac{1}{n}\sum_{i=1}^n A^i_v - \mathbb{E}A_v
    \end{align*}
    where $A_v^i \triangleq \exp\left ( \frac{1}{2a}\left ( v - \Teps(x) \right )^T(Y_i - \Teps(x)) - \frac{1}{4a}||v - \Teps(x)||^2 \right )\gamma(x,Y_i)$.
    
   Our first task is to control the increments of the process $Z_v$. Namely, we seek a bound of the form
    \begin{equation*}
        ||Z_u - Z_v||_{\psi_2} \leq K||u - v||.
    \end{equation*}
    By applying Lemma 2.6.8 and then Proposition 2.6.1 from \cite{vershynin2018high} we have
    \begin{align}
        ||Z_u - Z_v||_{\psi_2} &= ||(\frac{1}{n}\sum_{i=1}^n A^i_u - \mathbb{E}A_u) - (\frac{1}{n}\sum_{i=1}^n A^i_v - \mathbb{E}A_v)||_{\psi_2} \nonumber \\
        &= ||\frac{1}{n}\sum_{i=1}^n (A^i_u - A^i_v) - \mathbb{E}(A_u - A_v)||_{\psi_2} \nonumber \\
        &\leq C_8 ||\frac{1}{n}\sum_{i=1}^n (A^i_u - A^i_v)||_{\psi_2} \nonumber \\
        &\leq C_8C_9 \frac{1}{n} \left ( \sum_{i=1}^n  ||A^i_u - A^i_v||_{\psi_2}^2 \right )^{1/2} \label{eq:AuAv_squared}
    \end{align}
    where $C_8$ and $C_9$ are absolute constants. Let $C_{10} = C_8C_9$.
    
    Now we need to control $||A^i_u - A^i_v||_{\psi_2}$. Define
    \begin{equation*}
        \Gamma(x,r, \varepsilon) \triangleq  \frac{1}{\varepsilon} \left (\sigma^2 + \sqrt{2 d} \sigma r + r^2 + 2r\|x\| \right )
    \end{equation*}
    so that by Lemma \ref{lem:subg_ub_gamma} we have with probability 1
    \begin{equation*}
        \gamma(x,Y) \leq \exp\left (\Gamma(x,r, \varepsilon) \right ).
    \end{equation*}
    For the moment suppressing the dependence on $i$ we have 
    \begin{align*}
        |A_u - A_v| &= \bigg | 
            \exp \left (
                \frac{1}{2a}\left ( u - \Teps(x) \right )^T(Y - \Teps(x)) - \frac{1}{4a}||u - \Teps(x)||^2 
            \right ) \gamma(x,Y) \nonumber \\
            &\hspace{1cm} - \exp \left (
                \frac{1}{2a}\left ( v - \Teps(x) \right )^T(Y - \Teps(x)) - \frac{1}{4a}||v - \Teps(x)||^2 
            \right ) \gamma(x,Y)
        \bigg | \nonumber \\
        &= \gamma(x,Y) \bigg | 
            \exp \left (
                \frac{1}{2a}\left ( u - \Teps(x) \right )^T(Y - \Teps(x)) - \frac{1}{4a}||u - \Teps(x)||^2 
            \right ) \nonumber \\
        &\hspace{1cm} - \exp \left (
                \frac{1}{2a}\left ( v - \Teps(x) \right )^T(Y - \Teps(x)) - \frac{1}{4a}||v - \Teps(x)||^2 
            \right )
        \bigg | \nonumber\\
        &\leq \exp\left(\Gamma(x,r, \varepsilon) \right)  \bigg | 
            \exp \left (
                \frac{1}{2a}\left ( u - \Teps(x) \right )^T(Y - \Teps(x)) - \frac{1}{4a}||u - \Teps(x)||^2 
            \right ) \nonumber \\
        &\hspace{1cm} - \exp \left (
                \frac{1}{2a}\left ( v - \Teps(x) \right )^T(Y - \Teps(x)) - \frac{1}{4a}||v - \Teps(x)||^2 
            \bigg )
        \right |.
    \end{align*}
    Now by the assumption that $a \geq 2r^2$ and the fact that $u,v,Y,\Teps(x) \in B_2^d(0,r)$ we have 
    \begin{align*}
        \frac{1}{2a}\left ( u - \Teps(x) \right )^T(Y - \Teps(x)) - \frac{1}{4a}||u - \Teps(x)||^2 &\leq \frac{1}{2a}\left ( u - \Teps(x) \right )^T(Y - \Teps(x)) \\
        &\leq \frac{1}{2a} ||u-\Teps(x)||||Y - \Teps(x)|| \\
        &\leq \frac{4r^2}{2a} \leq \frac{4r^2}{4r^2} = 1 
    \end{align*} 
    and by an analogous computation replacing $u$ be $v$,
    \begin{align*}
        \frac{1}{2a}\left ( v - \Teps(x) \right )^T(Y - \Teps(x)) - \frac{1}{4a}||v - \Teps(x)||^2 &\leq 1.
    \end{align*}
    Using this, and the inequality valid for all $b,c \leq 1$ that $|e^b - e^c| \leq e|b-c|$, we have
    \begin{align*}
        &\bigg | \exp \left (
                \frac{1}{2a}\left ( u - \Teps(x) \right )^T(Y - \Teps(x)) - \frac{1}{4a}||u - \Teps(x)||^2 
            \right ) \\
            & - \exp \left (
                \frac{1}{2a}\left ( v - \Teps(x) \right )^T(Y - \Teps(x)) - \frac{1}{4a}||v - \Teps(x)||^2 \right ) \bigg | \\
        \leq& e \left |\frac{1}{2a}\left ( u - \Teps(x) \right )^T(Y - \Teps(x)) - \frac{1}{4a}||u - \Teps(x)||^2 - \frac{1}{2a}\left ( v - \Teps(x) \right )^T(Y - \Teps(x)) + \frac{1}{4a}||v - \Teps(x)||^2 \right | \\
        =& \frac{e}{2a}\left | (u-v)^T(Y - \Teps(x)) + \frac{1}{2}||v-\Teps(x)||^2 - \frac{1}{2}||u-\Teps(x)||^2 \right | \\
        =& \frac{e}{2a} \left |(u-v)^TY  + \frac{1}{2}||v||^2 - \frac{1}{2}||u||^2 \right | \\
        \leq& \frac{e}{2a}||u-v||||Y|| + \frac{e}{4a}\left | (||u|| - ||v||) (||u|| + ||v||)\right | \\
        \leq& \frac{re}{2a}||u-v|| + \frac{re}{2a} ||u-v|| \\
        =&\frac{re}{a}||u-v||.
    \end{align*}
    Using this we can conclude
    \begin{equation*}
        |A_u - A_v| \leq  \exp\left( \Gamma(x,r, \varepsilon)\right)  \frac{re}{a}||u-v||,
    \end{equation*}
    which implies (as in the proof of Lemma \ref{lem:a_bound})
    
    \begin{equation} \label{eq:AuAv_sg}
        ||A_u - A_v||_{\psi_2} \leq C_5 \exp\left( \Gamma(x,r, \varepsilon)\right)  \frac{re}{a}||u-v||.
    \end{equation}

    Now plugging (\ref{eq:AuAv_sg}) into (\ref{eq:AuAv_squared}) we obtain
    \begin{align*}
        ||Z_u - Z_v||_{\psi_2} &\leq C_{10} \frac{1}{n} \left( \sum_{i=1}^n \left(C_5 \exp\left(\Gamma(x,r, \varepsilon)\right)\frac{re}{a}||u-v|| \right )^2 \right )^{1/2} \\
        &= \frac{1}{\sqrt{n}} C_{11} \exp\left( \Gamma(x,r, \varepsilon)\right)\frac{re}{a}||u-v||
    \end{align*}

    where $C_{11} = C_5C_{10}$.
    Now using Theorem \ref{thm:hdp_chain} above with $K(x, r, \varepsilon) = \frac{1}{\sqrt{n}} C_{11} \exp\left( \Gamma(x,r, \varepsilon))\right) \frac{re}{a}$ we have
    \begin{align}
        &\mathbb{E}_{Y^n} \left [ \sup_{v \in B_2^d(0,r)} \int \exp\left ( \frac{1}{2a}\left ( v - \Teps(x) \right )^T(y - \Teps(x)) - \frac{1}{4a}||v - \Teps(x)||^2 \right )\gamma(x,y)[dQ^n - dQ](y) \right ] \nonumber \\
        = & \mathbb{E} \sup_{v \in B_2^d(0,r)} Z_v \nonumber \\ 
        \leq & C_6 K(x, r, \varepsilon) \int_0^{\infty} \sqrt{\log \mathcal{N}(B_2^d(0,r), ||\cdot||, \delta)} d\delta  \nonumber \\
        \leq&  C_6 K(x, r, \varepsilon) \int_0^{\infty} \sqrt{d\log (3r/\delta)} d\delta \nonumber \\
        \leq&  C_6 K(x, r, \varepsilon) \int_0^r \sqrt{d\log (3r/\delta)} d\delta \nonumber \\
        =&  C_6 K(x, r, \varepsilon) \sqrt{d} \int_0^r \sqrt{\log (r/\delta) + \log 3}  d\delta \nonumber \\
        &\leq C_6 K(x, r, \varepsilon) \sqrt{d}\left (r\sqrt{3} + \int_0^r \sqrt{\log (r/\delta) + \log 3}  d\delta \right ) \nonumber \\
        =& C_{12}K(x, r, \varepsilon)r\sqrt{d} \label{eq:to_integrate_over}.
    \end{align}
    Above we have used that $\mathcal{N}(B_2^2(0,r), \norm{\cdot}, \delta) = 1$ if $\delta \geq 1$ since in this case we have $B_2^2(0,r) \subset B_2^2(0,\delta)$, and taking the log of this makes the integrand 0 for all $\delta \geq r$. We have also used a standard bound on the covering number of balls in $\mathbb{R}^d$,
    $$
    \mathcal{N}(B_2^d(0,r), ||\cdot||, \delta) \leq \left ( \frac{3r}{\delta} \right )^d
    $$
    (see \cite{vershynin2018high} for a proof) and also
    \begin{equation*}
        \int_{0}^r \sqrt{\log (r/\delta)} d\delta = r\int_0^1 \sqrt{\log(1/\delta)} = \frac{r\sqrt{\pi}}{2}.
    \end{equation*}
    The next step is to unfix $x$ and integrate the bound in (\ref{eq:to_integrate_over}) against the distribution of $x$. This is done as follows
    \begin{align*}
    &\mathbb{E} \int\int \exp\left (h_0(x)^T(y - \Teps(x)) - a||h_0(x)||^2 \right )\gamma(x,y) [dQ^n - dQ](y)dP(x) \\
    &\leq \int C_{12}K(x, r, \varepsilon) r\sqrt{d} \,\,dP(x). \\
    &= \int C_{12}\frac{1}{\sqrt{n}} C_{11} \exp\left( \Gamma(x,r, \varepsilon))\right) \frac{re}{a} r\sqrt{d} dP(x) \\
    &= C_{13}\frac{r^2e\sqrt{d}}{a\sqrt{n}} \int \exp\left( \frac{1}{\varepsilon} \left (\sigma^2 + \sqrt{2 d} \sigma r + r^2 + 2r\|x\| \right ) \right ) dP(x) \\
    &= C_{13}\frac{r^2e\sqrt{d}}{a\sqrt{n}} \exp \left (\frac{1}{\varepsilon} \left (  \sigma^2 + \sqrt{2 d} \sigma r + r^2 \right ) \right ) \int \exp \left ( \frac{2r}{\varepsilon}\norm{x}\right ) dP(x).
    \end{align*}
    To conclude the proof we handle the integral by using the subgaussian condition on $\norm{X}$:
    \begin{align*}
        \mathbb{E} \exp\left ( \frac{2\norm{X}r}{\varepsilon} \right ) &= \mathbb{E} \exp\left ( \frac{2\norm{X}r}{\varepsilon} \right )\bm{1}[\norm{X} < (4rd\sigma^2/\varepsilon)] + \mathbb{E} \exp\left ( \frac{2\norm{X}r}{\varepsilon} \right )\bm{1}[\norm{X} \geq (4rd\sigma^2/\varepsilon)] \\
        &\leq \exp \left ( \frac{8rd\sigma^2}{\varepsilon^2}\right ) + \mathbb{E} \exp \left ( \frac{\norm{X}^2}{2d\sigma^2} \right ) \\
        &\leq \exp \left ( \frac{8rd\sigma^2}{\varepsilon^2}\right ) + 2.
    \end{align*}
    Plugging this in above and using that $a \geq C_1r^2$ completes the proof.
    \end{proof}

\subsection{Proof of Theorem \ref{thm:1_to_2_subg}} 
The main body of the proof is spent in establishing a result similar to Proposition \ref{prop:dual_prop_2}. It starts with the following result:
\begin{proposition}
     Let $P,Q$ be $\sigma^2$-subgaussian and let $P^n, Q^n$ be the random empirical measures. Then
     \begin{equation*}
         \mathbb{E}\left [ \sup_{\chi \in L^(\pi^n_\varepsilon)} \int \chi d\pi_\varepsilon^n - \int \int \left [ e^{\chi(x,y)} - 1\right ] \gamma_\varepsilon^{n,n} dP(x)dQ^n(y) \right ] \leq \frac{1}{\varepsilon}\mathbb{E} \left [ \int \feps^{n,n} dP^n - dP\right ]
     \end{equation*}
     where $\pi_\varepsilon^n$ is the random optimal coupling of $(P,Q^n),$ the function $\gamma^{n,n}_\varepsilon(x,y) = \exp(\frac{1}{\varepsilon}[\feps^{n,n}(x) + \geps^{n,n}(y) - 1/2\norm{x-y}^2])$, and $(\feps^{n,n}, \geps^{n,n})$ are the optimal potentials for $P^n,Q^n$.
\end{proposition} 
\begin{proof}
    Let $(\feps^{n,n},\geps^{n,n})$ be the optimal potentials for $(P^n,Q^n)$ and define $\eta(x,y) = \varepsilon\chi(x,y) + \feps^{n,n}(x) + \geps^{n,n}(y)$ for any $\chi \in L^1(\pi_\varepsilon^n)$. Applying Proposition \ref{prop:eta_dual} and calculating one has 
    \begin{align*}
        S_\varepsilon(P,Q_n) &\geq \int \eta d\pi_\varepsilon^n - \varepsilon \int\int \exp\left (\frac{1}{\varepsilon}\left[\eta(x,y) -\frac{1}{2}\norm{x-y}^2 \right ]  \right ) dP(x)dQ^n(y) + \varepsilon \\
        &= \varepsilon \int \chi(x,y) d\pi_\varepsilon^n + \int \feps^{n,n} dP(x) + \int \geps^{n,n}(y)dQ^n(y) - \varepsilon \int\int \left [e^{\chi(x,y)} - 1\right ]\gamma_\varepsilon^{n,n}(x,y) dP(x)dQ_n(y). 
    \end{align*}
    Rearranging, one obtains
    \begin{equation*}
        \int \chi(x,y)d\pi_\varepsilon^n - \int \int \left [e^{\chi(x,y)} - 1\right ]\gamma_\varepsilon^{n,n}(x,y) dP(x)dQ_n(y) \leq \frac{1}{\varepsilon} \left [ S_\varepsilon(P,Q^n) - \int \feps^{n,n} dP - \int \geps^{n,n} dQ^n \right ].
    \end{equation*}
    Since this holds uniformly over $\chi \in L^1(\pi_\varepsilon^n)$ we can take the supremum on the left as well
    \begin{align*}
       \sup_{\chi \in L^1(\pi_\varepsilon^n)} & \int \chi(x,y)d\pi_\varepsilon^n - \int \int \left [e^{\chi(x,y)} - 1\right ]\gamma_\varepsilon^{n,n}(x,y) dP(x)dQ_n(y) \nonumber \\
       &\leq \frac{1}{\varepsilon} \left [ S_\varepsilon(P,Q^n) - \int \feps^{n,n} dP - \int \geps^{n,n} dQ^n \right ]
    \end{align*}
    We now turn our attention to the right side. Let $(\feps^n,\geps^n)$ be optimal for $(P,Q^n)$. By optimality we have
    \begin{align*}
        &\int \feps^{n,n} dP^n + \int \geps^{n,n}dQ^n = S_\varepsilon(P^n,Q^n) \\
        &\geq \int \feps^n dP^n + \int \geps^n dQ^n - \varepsilon \int\int \exp \left ( \frac{1}{\varepsilon} \left [ \feps^n(x) + \geps^n(y) - \frac{1}{2}\norm{x-y}^2\right ] \right )dP^n(x)dQ^n(y) + \varepsilon \\
        &= \int \feps^n dP^n + \int \geps^n dQ^n - \varepsilon \int \left [ \int \exp \left ( \frac{1}{\varepsilon} \left [ \feps^n(x) + \geps^n(y) - \frac{1}{2}\norm{x-y}^2\right ] \right )dQ^n(y) \right ] dP^n(x) + \varepsilon \\
        &= \int \feps^n dP^n + \int \geps^n dQ^n - \varepsilon \int 1 dP^n(x) + \varepsilon \\
        &= \int \feps^n dP^n + \int \geps^n dQ^n.
    \end{align*}
    Comparing the first and last we have shown
    \begin{equation*}
        \int \feps^{n,n} dP^n + \int \geps^{n,n}dQ^n \geq \int \feps^n dP^n + \int \geps^n dQ^n.
    \end{equation*}
    From here we can develop a bound as follows
    \begin{align*}
        &S_\varepsilon(P,Q_n) - \int \feps^{n,n} dP - \int \geps^{n,n} dQ^n \\
        &= \int \feps^{n} dP + \int \geps^{n} dQ^n - \int \feps^{n,n} dP - \int \geps^{n,n} dQ^n \\
        &= \int \feps^n - \feps^{n,n} dP + \int \geps^n dQ^n + \int \feps^n dP^n - \int \feps^n dP^n - \int \geps^{n,n} dQ^n \\
        &\leq \int \feps^n - \feps^{n,n} dP + \int \geps^{n,n} dQ^n + \int \feps^{n,n} dP^n - \int \feps^n dP^n - \int \geps^{n,n} dQ^n \\
        &= \int \feps^{n,n} [dP^n - dP] + \int \feps^{n} [dP - dP^n].  
    \end{align*}
    Note that $\feps^n$ is independent of $P^n$, and so conditioned on $Q^n$ we have
    \begin{equation*}
        \mathbb{E}\left[\int \feps^{n} [dP - dP^n] \right ] = 0.
    \end{equation*}
    Backtracking and taking expectations we have shown 
    \begin{align*}
         &\mathbb{E}\left [ \sup_{\chi \in L^(\pi^n_\varepsilon)} \int \chi d\pi_\varepsilon^n - \int \int \left [ e^{\chi} - 1\right ] \gamma_\varepsilon^{n,n}(x,y) dP(x)d^nQ(y) \right ] \\
         &\leq \frac{1}{\varepsilon} \mathbb{E}  \left [ S_\varepsilon(P,Q^n) - \int \feps^{n,n} dP - \int \geps^{n,n} dQ^n \right ] \\
         &\leq \frac{1}{\varepsilon}\mathbb{E}\int \feps^{n,n} [dP^n - dP] + \int \feps^{n} [dP - dP^n]  \\
         &= \frac{1}{\varepsilon}\mathbb{E}\int \feps^{n,n} [dP^n - dP]
    \end{align*}
    which proves the result.
\end{proof}

The next step is to control $\mathbb{E}\int \feps^{n,n}[dP^n - dP]$. Thankfully, this has already been controlled in the literature. While not explicitly stated in this form, the calculations in \cite{mena2019statistical} show
\begin{theorem} If $P,Q$ are $\sigma^2$-subgaussian, then 
    \begin{equation*}
        \mathbb{E}\int \feps^{n,n} [dP^n - dP] \leq K_{d,0} \cdot \varepsilon \left ( 1 + \frac{\sigma^{\lceil 5d/2\rceil + 6}}{\varepsilon^{\lceil 5d / 4 \rceil + 3}}\right ) \frac{1}{\sqrt{n}}.
    \end{equation*}
\end{theorem}

We are now ready to proceed to the proof of
Theorem \ref{thm:1_to_2_subg}.
\begin{proof}
    Choose $\chi(x,y) = h(x)^T(y - \Teps^{n,n}(x)) - a\norm{h(x)}^2$ for $a$ and $h$ to be specified. First note that for fixed $x$ we have
    \begin{align*}
        \int h(x)^T(y-\Teps^{n,n}(x))\gamma_\varepsilon^{n,n}(x,y) dQ^n(y) &= h(x)^T \left [ \int y\gamma_\varepsilon^{n,n}(x,y) dQ^n(y) - \Teps^{n,n}(x) \right ] \\
        &= h(x)^T \left [ \Teps^{n,n}(x) - \Teps^{n,n}(x) \right ] = 0 
    \end{align*}
    which means that by Hoeffding's inequality we have for all fixed $x$
    \begin{align*}
        \int  \exp\left ( \chi(x,y) \right ) \gamma_\varepsilon^{n,n}(x,y) dQ(y) &= \exp(-a\norm{h(x)}^2) \int \exp(h(x)^T(y-\Teps^{n,n}(x))) \gamma_\varepsilon^{n,n}(x,y) dQ^n(y) \\
        &\leq \exp(-a\norm{h(x)}^2) \int \exp(C_4(h(x)^T(y-\Teps^{n,n}(x)))^2) \gamma_\varepsilon^{n,n}(x,y) dQ^n(y) \\
        &\leq \exp(-a\norm{h(x)}^2) \int \exp(C_44r^2\norm{h(x)}_2^2 ) \gamma_\varepsilon^{n,n}(x,y) dQ^n(y) \\
        &= \exp((C_44r^2 - a)\norm{h(x)}^2)
    \end{align*}
    In particular for $a \geq C_44r^2$ this gives
    \begin{equation*}
        \int  \exp\left ( \chi(x,y) \right ) \gamma_\varepsilon^{n,n}(x,y) dQ^n(y) \leq 1
    \end{equation*}
    which implies 
    \begin{equation*}
        \int  \left [ \exp\left ( \chi(x,y) \right ) - 1 \right ] \gamma_\varepsilon^{n,n}(x,y) dQ^n(y)dP(x) \leq 0.
    \end{equation*}
    Now set $h_0(x) = \frac{1}{2a}\norm{\Teps^n(x) - \Teps^{n,n}(x)}$. A direct calculation gives
    \begin{align*}
        \int h_0(x)^T(y - \Teps^{n,n}) - a\norm{h_0(x)}d\pi_\varepsilon^n  &= \int  \frac{1}{4a}\norm{\Teps^n(x) - \Teps^{n,n}(x)}^2 dP(x) \\
        &= \frac{1}{4a}\norm{\Teps^n - \Teps^{n,n}}_{L^2(P)}^2
    \end{align*}
    Combining this with the fact that $\int  \left [ \exp\left ( \chi(x,y) \right ) - 1 \right ] \gamma_\varepsilon^{n,n}(x,y) dQ(y)dP(x) \leq 0$ we have after taking expectations and applying the results above that
    \begin{align*}
        \mathbb{E}\frac{1}{4a}\norm{\Teps^n - \Teps^{n,n}}_{L^2(P)}^2 &\leq \mathbb{E}\int \chi d\pi_\varepsilon^n - \int  \left [ \exp\left ( \chi(x,y) \right ) - 1 \right ] \gamma_\varepsilon^{n,n}(x,y) dQ^n(y)dP(x) \\
        &\leq \mathbb{E} \sup_\chi \int \chi d\pi_\varepsilon^n - \int  \left [ \exp\left ( \chi(x,y) \right ) - 1 \right ] \gamma_\varepsilon^{n,n}(x,y) dQ^n(y)dP(x) \\
        &\leq \frac{1}{\varepsilon}\mathbb{E}\int \feps^{n,n} [dP^n - dP]  \\
        &\leq K_{d,0} \cdot \left ( 1 + \frac{\sigma^{\lceil 5d/2\rceil + 6}}{\varepsilon^{\lceil 5d / 4 \rceil + 3}}\right ) \frac{1}{\sqrt{n}}
    \end{align*}
    valid for all $a \geq 4C_4r^2$. Multiplying both sides by this quantity gives
    \begin{equation*}
        \mathbb{E}\norm{\Teps^n - \Teps^{n,n}}_{L^2(P)}^2 \leq 16C_4r^2  K_{d,2} \cdot \left ( 1 + \frac{\sigma^{\lceil 5d/2\rceil + 6}}{\varepsilon^{\lceil 5d / 4 \rceil + 3}}\right ) \frac{1}{\sqrt{n}}
    \end{equation*}
    where $K_{d,2} = 16C_4r^2  K_{d,0}$
\end{proof}

\begin{table}[ht]
\centering
\begin{tabular}{|c|c|c|}
\hline
Constant    & First Introduction         & Value or Source                                       \\ \hline \hline
$C_d$         & Statement of Definition 2     & $ (2\Gamma(d/2)  )^{-1} \sqrt{\pi}(d-1)\Gamma((d-1)/2)$                                                \\ \hline
$C_0$       & Statement of Lemma 2       & $4C_5^2C_4$                                           \\ \hline
$C_1$       & Statement of Lemma 3       & $\max(C_0,2)$                                         \\ \hline
$C_2$    & Statement of Lemma 3           & $eC_{13}/C_1$                   \\ \hline
$C_3$       & Statement of Theorem \ref{thm:0_to_1_subg}           & $4C_1C_2$     \\ \hline
$C_4$       & Proof of Lemma 2           & (2.5.2) in\cite{vershynin2018high}     \\ \hline
$C_5$       & Proof of Lemma 2           & (2.3.8.iii) in \cite{vershynin2018high}, can be taken as $(\log 2)^{-1/2}$ \\ \hline
$C_6$       & Proof of Lemma 2           & $C_5^2C_4$                                            \\ \hline
$C_7$       & Proof of Lemma 3           & (8.1.3) in \cite{vershynin2018high}                                       \\ \hline
$C_8$       & Proof of Lemma 3           & (2.6.8) in \cite{vershynin2018high}                                     \\ \hline
$C_9$       & Proof of Lemma 3           & (2.6.1) in \cite{vershynin2018high}                                     \\ \hline
$C_{10}$    & Proof of Lemma 3           & $C_8C_9$                                              \\ \hline
$C_{11}$    & Proof of Lemma 3           & $C_5C_{10}$                                           \\ \hline
$C_{12}$    & Proof of Lemma 3           & $C_7(\sqrt{\log 3} + \sqrt{\pi}/2)$                   \\ \hline
$C_{13}$    & Proof of Lemma 3           & $C_{11}C_{12}$                   \\ \hline
$K_{d,0}$    & Statement of Proposition \ref{prop:mena2019}           & See \cite{mena2019statistical}  \\ \hline
$K_{d,1}$    & Statement of Theorem \ref{thm:0_to_1_subg}           & $4C_1K_{d,0}$  \\ \hline
$K_{d,2}$    & Statement of Theorem \ref{thm:1_to_2_subg}           & $16C_4K_{d,0}$  \\ \hline
\end{tabular}

\caption{Table of constants used in the proofs of the theoretical results.}
\end{table}

\section{Proof of Proposition \ref{prop:properties}}\label{supp:properties}
\begin{proof}

\begin{itemize}
\setlength \itemsep{3mm}
    \item[(a)] Given $ X, X'\overset{i.i.d.}{\sim} \mu_{X}\in \mathcal P(\Omega)$, and $ Y, Y'\overset{i.i.d.}{\sim} \mu_{Y}\in \mathcal P(\Omega)$, using Lemmas 2.2, 2.3 from \cite{baringhaus2004new}, we have,
    \begin{align*}
        2\mathbb E\|X -  Y\|-\mathbb E\|    X -     X'\|-\mathbb E\| Y- Y'\|\notag = C_d\int_{\mathcal S^{d-1}}\int_{\mathbb R} \left ( \mathbb{P}(   a^\top   X \leq t) -  P(  a^\top    Y \leq t)\right )^2 d\kappa(a) dt.
    \end{align*}
    Following the above expression, for the rank transformed random variables $\sR(X)$, $\sR(X')$, $\sR(Y)$, and $\sR(  Y')$ corresponding to $  X, X', Y$, and $Y'$, respectively, we can express the soft rank energy as,  
     \begin{align*}
        &2\mathbb E\| \sR(X) -  \sR(Y)\|- \mathbb E\|  \sR(X) -   \sR(X')\|-\mathbb E\|  \sR(Y) -   \sR(Y')\ |\notag \\ 
        =& C_d \int_{\mathcal S^{d-1}}\int_{\mathbb R} \left ( \mathbb P(a^\top \sR(X) \leq t) -\mathbb P(a^\top  \sR(Y) \leq t)\right )^2 \text d\kappa(a)dt.
    \end{align*}
    
    \item [(b)] Following the definition of $\sre_{\lambda,\varepsilon}$,
    \begin{align*}
        \sRE(P_X,P_Y)^2 &=C_d  \int_{\mathbb R} \int_{\mathcal S^{d-1}} \Big( \mathbb{P}(  a^\top  \sR(X)\leq t)
        - \mathbb{P}(a^\top \sR(Y)\leq t)\Big)^2 d\kappa( a) dt \notag\\
        & =C_d  \int_{\mathbb R} \int_{\mathcal S^{d-1}} \Big( \mathbb{P}(a^\top \sR(  Y)\leq t)
        - \mathbb{P}( a^\top \sR(X)\leq t)\Big)^2 d\kappa(a) dt \notag \\
        & = \sre_{\lambda,\varepsilon}(P_Y,P_X)^2. 
    \end{align*}
    \item [(c)] Assuming $X \overset{d}{=}Y$, we have $\mathbb{P}(a^\top X\leq t)=\mathbb{P}(a^\top Y\leq t)$ for all $a\in \mathcal S^{d-1}$ and $t\in \mathbb R$ \cite{baringhaus2004new}. Since $\sR(X)$, $\sR(Y)$ are also mapped random vectors corresponding to $X, Y$, respectively, we see that $\pr\big(a^\top\sR(X)\leq t\big) =\pr\big(a^\top\sR(Y)\leq t\big)$ and the result follows. 
\end{itemize}
\end{proof}

\section{Proof of Proposition \ref{prop:sRE_eps_convergence}}

\begin{proof}
    It is equivalent to show that
    \begin{equation*}
         \lim_{\varepsilon \rightarrow 0^+} |\sRE(P_X,P_Y)^2 - \RE(P_X,P_Y)^2| = 0.
    \end{equation*}
    Using the original definitions, we have
    \begin{align*}
        &|\sRE(P_X,P_Y)^2 - \RE(P_X,P_Y)^2| \\
        =& \bigg| 2\mathbb{E} \left [ ||\sR(X) - \sR(Y) || -  ||\R(X) - \R(Y) ||\right ] - \mathbb{E} \left [ ||\sR(X) - \sR(X') || -  ||\R(X) - \R(X') ||\right ]  \\
        &- \mathbb{E} \left [ ||\sR(Y) - \sR(Y') || -  ||\R(Y) - \R(Y') ||\right ]  \bigg |\\ 
        \leq & 2 \mathbb{E} \bigg | ||\sR(X) - \sR(Y) || -  ||\R(X) - \R(Y) || \bigg | + \mathbb{E} \bigg | ||\sR(X) - \sR(X') || -  ||\R(X) - \R(X') || \bigg | \\
        & + \mathbb{E} \bigg | ||\sR(Y) - \sR(Y') || -  ||\R(Y) - \R(Y') || \bigg | \\
        \leq & 2 \mathbb{E} ||[\sR(X) - \sR(Y)] -  [\R(X) - \R(Y)] || + \mathbb{E}  ||[\sR(X) - \sR(X') ] - [\R(X) - \R(X')] || \\
        & + \mathbb{E} ||[\sR(Y) - \sR(Y')] - [\R(Y) - \R(Y')] || \\
        \leq & 4 \mathbb{E} ||\R(X) - \sR(X)|| + 4 \mathbb{E} ||\R(Y) - \sR(Y)||\\
        \leq & 4 \sqrt{\mathbb{E} ||\R(X) - \sR(X)||^2} + 4  \sqrt{\mathbb{E} ||\R(Y) - \sR(Y)||^2} \\
        \leq & 8 \sqrt{\mathbb{E} ||\R(X) - \sR(X)||^2 + \mathbb{E} ||\R(Y) - \sR(Y)||^2} \\
        \leq & \frac{8}{\min(\sqrt{\lambda}, \sqrt{1-\lambda})} \sqrt{\mathbb{E} [\lambda||\R(X) - \sR(X)||^2  + (1-\lambda)||\R(Y) - \sR(Y)||^2]} \\
        = & \frac{8}{\min(\sqrt{\lambda}, \sqrt{1-\lambda})} ||\R - \sR||_{L^2(P_\lambda)} \\
        \lesssim & \frac{8}{\min(\sqrt{\lambda}, \sqrt{1-\lambda})} \sqrt{\varepsilon^2I_0(P_\lambda, \text{Unif}([0,1]^d)) + \varepsilon^{\min(\alpha, 3)/2}}
    \end{align*}
    where for the final inequality we have used Proposition 1 in \cite{pooladian2022debiaser} and where the implicit constant is independent of $\varepsilon$. Here $I_0$ is the integrated Fisher information along the geodesic from $P_\lambda$ and $\text{Unif}([0,1]^d)$ which under the assumptions made is guaranteed to be finite, \cite{chizat2020faster, pooladian2021entropic}.
    Taking the limit in the last expression we have
    \begin{align*}
        0 &\leq \lim_{\varepsilon \rightarrow 0^+} |\sRE(P_X,P_Y)^2 - \RE(P_X,P_Y)^2| \\
        &\lesssim \lim_{\varepsilon \rightarrow 0^+} \frac{8}{\min(\sqrt{\lambda}, \sqrt{1-\lambda})} \sqrt{\varepsilon^2I_0(P_\lambda, \text{Unif}([0,1]^d)) + \varepsilon^{\min(\alpha, 3)/2}} = 0
    \end{align*}
    which shows indeed $\lim_{\varepsilon \rightarrow 0^+} |\sRE(P_X,P_Y)^2 - \RE(P_X,P_Y)^2| = 0.$
\end{proof}

\section{Proof of Theorem \ref{thm:sre_convergence_rate}} \label{sec:sre_cr_proof}

\begin{proof}
    Using the notation that $\sRn$ is the random independently estimated map and $X^m = (X_1,...,X_m), Y^n = (Y_1,...,Y_n)$ are the samples used to evaluate the statistic (which are independent of $\sRn$) we immediately have the following 
\[||\sREn(P_X, P_Y)^2 - \sRE(P_X,P_Y)^2||_{L^2}^2 
        = \underset{\sRn}{\mathbb{E}} \left [ \underset{X^n,Y^m}{\mathbb{E}} \left [ \left ( \sREn(P_X, P_Y)^2 - \sRE(P_X,P_Y)^2 \right )^2 \bigg | \sRn \right ] \right ].\]
    We first consider the inner expectation. For brevity we suppress the conditioning on $\sRn$ and also introduce the following six collections of random variables
    \begin{align*}
        R^X \triangleq \frac{1}{m^2}\sum_{i,j=1}^m||\sR(X_i) - \sR(X_j)||,  \hspace{0.5cm} & \hspace{0.5cm} \hat{R}^X \triangleq \frac{1}{m^2}\sum_{i,j=1}^m||\sRn(X_i) - \sRn(X_j)||, \\ 
        R^Y \triangleq \frac{1}{n^2}\sum_{i,j=1}^n||\sR(Y_i) - \sR(Y_j)||,  \hspace{0.5cm} & \hspace{0.5cm} \hat{R}^Y \triangleq \frac{1}{n^2}\sum_{i,j=1}^n||\sRn(Y_i) - \sRn(Y_j)||, \\ 
        R^{XY} \triangleq \frac{2}{nm}\sum_{i=1}^m\sum_{j=1}^n||\sR(X_i) - \sR(Y_j)||,  \hspace{0.5cm} & \hspace{0.5cm} \hat{R}^{XY} \triangleq \frac{2}{nm}\sum_{i=1}^m\sum_{j=1}^n||\sRn(X_i) - \sRn(Y_j)||.
    \end{align*}
    These are further compressed into 
    \begin{align*}
        \hat{R} &\triangleq \hat{R}^{XY} - \hat{R}^{X} - \hat{R}^{Y}, \\
        R &\triangleq R^{XY} - R^X - R^Y.
    \end{align*}
    We also introduce the notation
    \begin{align*}
        E^X &\triangleq \underset{X,X'}{\mathbb{E}}||\sR(X) - \sR(X')||, \\
        E^Y &\triangleq \underset{Y,Y'}{\mathbb{E}}||\sR(Y) - \sR(Y')||, \\
        E^{XY} &\triangleq 2\underset{X,Y}{\mathbb{E}}||\sR(X) - \sR(Y)||, \\
        E &\triangleq E^{XY} - E^X - E^Y.
    \end{align*}
    With this notation we have
    \begin{align*}
        \mathbb{E} \left [ \left ( \sREn(P_X, P_Y)^2 - \sRE(P_X,P_Y)^2 \right )^2 \right ]  &= \mathbb{E} ( \hat{R} - E )^2 \\
        &= \mathbb{E} ( [\hat{R} - R] - [R - E])^2 \\
        &\leq 2\mathbb{E} (\hat{R} - R)^2 + 2\mathbb{E}(R - E)^2
    \end{align*}
    We now control the two expectations separately:
    \begin{align*}
        \mathbb{E}(\hat{R} - R)^2 &= \mathbb{E} \left ( [\hat{R}^{XY} - R^{XY}] + [R^X - \hat{R}^X] + [R^Y - \hat{R}^Y] \right )^2 \\
        &\leq 3\mathbb{E}(\hat{R}^{XY} - R^{XY})^2 + 3\mathbb{E}(\hat{R}^{X} - R^{X})^2 + 3\mathbb{E}(\hat{R}^{Y} - R^{Y})^2.
    \end{align*}
    Next we control these three expectations separately:
    \begin{align*}
        \mathbb{E}(\hat{R}^{XY} - R^{XY})^2 &= \mathbb{E} \left (\frac{2}{nm}\sum_{i=1}^m\sum_{j=1}^n||\sRn(X_i) - \sRn(Y_j)|| - ||\sR(X_i) - \sR(Y_j)||  \right )^2 \\
        &\leq 4 \mathbb{E} \frac{1}{nm}\sum_{i=1}^m\sum_{j=1}^n\left ( ||\sRn(X_i) - \sRn(Y_j)|| - ||\sR(X_i) - \sR(Y_j)|| \right )^2 \\
        &\leq 4 \mathbb{E} \frac{1}{nm}\sum_{i=1}^m\sum_{j=1}^n ||[\sRn(X_i) - \sRn(Y_j)] - [\sR(X_i) - \sR(Y_j)]||^2 \\
        &\leq 4 \mathbb{E} \frac{1}{nm}\sum_{i=1}^m\sum_{j=1}^n \left (||[\sRn(X_i) - \sR(X_i)|| + ||\sRn(Y_j) - \sR(Y_j)]||\right )^2 \\
        &\leq 8 \mathbb{E} \frac{1}{nm}\sum_{i=1}^m\sum_{j=1}^n ||[\sRn(X_i) - \sR(X_i)||^2 + ||\sRn(Y_j) - \sR(Y_j)]||^2 \\
        &= 8 \mathbb{E} \frac{1}{m} \sum_{i=1}^m ||[\sRn(X_i) - \sR(X_i)||^2 + 8 \mathbb{E} \frac{1}{n} \sum_{i=1}^n ||[\sRn(Y_i) - \sR(Y_i)||^2 \\
        &= 8\mathbb{E} ||\sRn(X) - \sR(X)||^2 + ||\sRn(Y) - \sR(Y)||^2 \\
        &\leq \frac{8}{\min(\lambda, 1-\lambda)} \mathbb{E} \lambda||\sRn(X) - \sR(X)||^2 + (1-\lambda)||\sRn(Y) - \sR(Y)||^2 \\
        &= \frac{8}{\min(\lambda, 1-\lambda)} ||\sRn - \sR||_{L^2(P_\lambda)}^2.
    \end{align*}
    In the third and fourth lines we have used the reverse triangle inequality followed by the triangle inequality to re-group the terms. 
    In the second to last line we have made use of the inequality valid for all $a,b \geq 0, c \in (0,1)$
    $$
    a + b \leq \frac{c}{\min(c, 1-c)}a + \frac{1-c}{\min(c,1-c)}b = \frac{1}{\min(c, 1-c)}\left [ca + (1-c)b \right ].
    $$
    The last line follows from the fact that $P_\lambda = \lambda P_X + (1-\lambda)P_Y$. 
    Next we have
    \begin{align*}
        \mathbb{E}(\hat{R}^X - R^X)^2 &= \mathbb{E} \left (\frac{1}{n^2} \sum_{i,j=1}^n ||\sRn(X_i) - \sRn(X_j)|| - ||\sR(X_i) - \sR(X_j)||  \right )^2 \\
        &\leq \mathbb{E}\frac{1}{n^2} \sum_{i,j=1}^n \left ( ||\sRn(X_i) - \sRn(X_j)|| - ||\sR(X_i) - \sR(X_j)|| \right )^2 \\
        &\leq \mathbb{E}\frac{1}{n^2} \sum_{i,j=1}^n  ||[\sRn(X_i) - \sRn(X_j)] - [\sR(X_i) - \sR(X_j)]||^2 \\ 
        &\leq \mathbb{E}\frac{1}{n^2} \sum_{i,j=1}^n  \left ( ||\sRn(X_i) - \sR(X_i)|| +  ||\sRn(X_j) - \sR(X_j)|| \right ) ^2 \\ 
        &\leq 2\mathbb{E} \frac{1}{n^2} \sum_{i,j=1}^n  ||\sRn(X_i) - \sR(X_i)||^2 +  ||\sRn(X_j) - \sR(X_j)||^2 \\
        &= 4\mathbb{E} \frac{1}{n}\sum_{i=1}^n  ||\sRn(X_i) - \sR(X_i)||^2 \\
        &= 4 \mathbb{E} ||\sRn(X) - \sR(X)||^2.
    \end{align*} 
    Again in the third and fourth lines we have used the reverse-triangle inequality followed by the triangle inequality.
    An exactly analogous calculation for $Y$ shows
    \begin{equation*}
        \mathbb{E}(\hat{R}^Y - R^Y)^2 \leq 4\mathbb{E} ||\sRn(Y) - \sR(Y)||^2.
    \end{equation*}
    Combining these two we have similarly to above that
    \begin{align*}
        \mathbb{E}(\hat{R}^X - R^X)^2 + \mathbb{E}(\hat{R}^Y - R^Y)^2 &\leq 4\mathbb{E} \left [ ||\sRn(X) - \sR(X)||^2 + ||\sRn(Y) - \sR(Y)||^2 \right ] \\
        &\leq \frac{4}{\min(\lambda, 1-\lambda)} ||\sRn - \sR||_{L^2(P_\lambda)}^2.
    \end{align*}
    Combining bounds we have
    \begin{align*}
        \mathbb{E}(\hat{R} - R)^2 &\leq \frac{24}{\min(\lambda, 1-\lambda)}||\sRn - \sR||_{L^2(P_\lambda)}^2 + \frac{12}{\min(\lambda, 1-\lambda)}||\sRn - \sR||_{L^2(P_\lambda)}^2 \\
        &= \frac{36}{\min(\lambda, 1-\lambda)}||\sRn - \sR||_{L^2(P_\lambda)}^2.
    \end{align*}
    Now we turn our attention to $\mathbb{E}(R - E)^2$. First we show that $R- E$ is mean-zero:
    \begin{align*}
        \mathbb{E}R &= \mathbb{E} \frac{2}{nm} \sum_{i=1}^m\sum_{j=1}^n ||\sR(X_i) - \sR(Y_j)|| - \frac{1}{m^2} \sum_{i,j=1}^m ||\sR(X_i) - \sR(X_j)|| - \frac{1}{n^2} \sum_{i,j=1}^n ||\sR(Y_i) - \sR(Y_j)|| \\
        &= 2\mathbb{E} ||\sR(X) - \sR(Y)|| - \mathbb{E}||\sR(X) - \sR(X')|| - \mathbb{E}||\sR(Y) - \sR(Y')|| \\
        &= E^{XY} - E^X - E^Y = E.
    \end{align*}
    Subtracting $E$ from the first and last shows $\mathbb{E}[R - E] = 0$. Using this we have
    \begin{equation*}
        \mathbb{E}[(R-E)^2] = \mathbb{E}[(R-E)^2] - \mathbb{E}[R-E]^2 = \text{Var}(R-E).
    \end{equation*}
    To control the variance we apply the Efron-Stein inequality (\cite{boucheron2013concentration} Theorem 3.1) to the function
    $$
    f(X_1,...,X_m,Y_1,...,Y_n) = \frac{2}{nm} \sum_{i=1}^m\sum_{j=1}^n ||\sR(X_i) - \sR(Y_j)|| - \frac{1}{m^2} \sum_{i,j=1}^m ||\sR(X_i) - \sR(X_j)|| - \frac{1}{n^2} \sum_{i,j=1}^n ||\sR(Y_i) - \sR(Y_j)||.
    $$
    First note that we have the bounds
    \begin{align*}
        &|f(X_1,...,X_{i-1},X_i,X_{i+1},...,X_m,Y_1,...,Y_n) - f(X_1,...,X_{i-1},X_i',X_{i+1},...,X_m,Y_1,...,Y_n)| \\
        &= \left |\frac{2}{nm}\sum_{j=1}^n \left (||\sR(X_i) - \sR(Y_j)|| - ||\sR(X_i') - \sR(Y_j)|| \right ) - \frac{1}{m^2} \sum_{j \neq i}^m \left ( ||\sR(X_i) - \sR(X_j)|| - ||\sR(X_i') - \sR(X_j)|| \right )  \right | \\
        &\leq \frac{2}{nm} \sum_{j=1}^n \left | ||\sR(X_i) - \sR(Y_j)|| - ||\sR(X_i') - \sR(Y_j)|| \right | + \frac{1}{m^2} \sum_{j \neq i}^m \left | ||\sR(X_i) - \sR(X_j)|| - ||\sR(X_i') - \sR(X_j)|| \right | \\
        &\leq \frac{2}{nm} \sum_{j=1}^n ||\sR(X_i) - \sR(X_i')|| + \frac{1}{m^2} \sum_{j\neq i}^m  ||\sR(X_i) - \sR(X_i')|| \\
        &\leq \frac{3}{m} ||\sR(X_i) - \sR(X_i')|| \\
        &\leq \frac{3\sqrt{d}}{m}
    \end{align*}
    where we have in the last line used the $\sR$ maps into $[0,1]^d$ and the diameter is $\sqrt{d}$. A completely analogous computation shows
    $$|f(X_1,...,X_m,Y_1,...,Y_{i-1},Y_i,Y_{i+1},...,Y_n) - f(X_1,...,X_m,Y_1,...,Y_{i-1},Y_i',Y_{i+1},...,Y_n)| \leq \frac{3\sqrt{d}}{m}.$$
    
    Using this bound in the Efron-Stein inequality we have
    \begin{align*}
        \mathbb{E}[(R-E)^2] &= \text{Var}(R-E) \\
        &\leq \frac{1}{2}\sum_{i=1}^m \mathbb{E} \left ( f(X_1,...,X_{i-1},X_i,X_{i+1},...,X_m,Y_1,...,Y_n) - f(X_1,...,X_{i-1},X_i',X_{i+1},...,X_m,Y_1,...,Y_n) \right )^2 \\
        &+ \frac{1}{2} \sum_{i=1}^n \left ( f(X_1,...,X_m,Y_1,...,Y_{i-1},Y_i,Y_{i+1},...,Y_n) - f(X_1,...,X_m,Y_1,...,Y_{i-1},Y_i',Y_{i+1},...,Y_n) \right )^2 \\
        &\leq \frac{1}{2}\sum_{i=1}^m \mathbb{E} \left ( \frac{3\sqrt{d}}{m} \right )^2 + \frac{1}{2}\sum_{i=1}^n \mathbb{E} \left ( \frac{3\sqrt{d}}{n} \right )^2 \\
        &= \frac{9d}{2m} + \frac{9d}{2n} \\
        &= \frac{9d(m+n)}{2mn}.
    \end{align*}
    Collecting terms we have, conditionally on the estimate of $\sRn$ that
    \begin{align*}
        \mathbb{E} \left [ \left ( \sREn(P_X, P_Y)^2 - \sRE(P_X,P_Y)^2 \right )^2 \right ]
        &\leq 2\mathbb{E} (\hat{R} - R)^2 + 2\mathbb{E}(R - E)^2 \\
        &\leq \frac{72}{\min(\lambda, 1-\lambda)} ||\sRn - \sR||_{L^2(P_\lambda)}^2
        + \frac{9d(m+n)}{mn}.
    \end{align*}
    Unconditioning and applying either Theorem \ref{thm:subg_conv_rate} or Theorem \ref{thm:bounded_conv_rate} completes the proof.
\end{proof}

\section{Proof of Theorem \ref{thm:srmmd_convergence_rate}} \label{sec:srmmd_proof}

The proof is essentially the same as the proof of Theorem \ref{thm:sre_convergence_rate}. The key difference is that we require a kernel analog of the reverse-triangle, followed by triangle inequality trick:
\begin{align*}
    |||\sRn(X_i) - \sRn(Y_j)|| - ||\sR(X_i) - \sR(Y_j)||| &\leq ||[\sRn(X_i) - \sRn(Y_j)] - [\sR(X_i) - \sR(Y_j)]|| \\
    &\leq ||\sRn(X_i) - \sR(X_i)|| + ||\sRn(Y_j) - \sR(Y_j)|| 
\end{align*}
This is achieved through
\begin{align*}
    &|k(\sRn(X_i),\sRn(Y_j)) - k(\sR(X_i),\sR(Y_j))| \\
    =&|[k(\sRn(X_i),\sRn(Y_j)) - k(\sR(X_i),\sRn(Y_j))] + [k(\sR(X_i),\sRn(Y_j)) -  k(\sR(X_i),\sR(Y_j))]| \\
    \leq& |k(\sRn(X_i),\sRn(Y_j)) - k(\sR(X_i),\sRn(Y_j))| + |k(\sR(X_i),\sRn(Y_j)) -  k(\sR(X_i),\sR(Y_j))| \\ 
    \leq& l||\sRn(X_i) - \sR(X_i)|| + l ||\sRn(Y_j) - \sR(Y_j)||.
\end{align*}
With this key inequality established, one can show the result by following the proof of Theorem \ref{thm:sre_convergence_rate}, substituting this inequality in the two places where the reverse-triangle followed triangle inequality is used. In both places these inequalities are done inside of a squaring so ultimately we gain a factor of $l^2$.

\section{Related Works on Knockoff Generation}\label{supp:related_works_knockoffs}
The seminal work \cite{candes2016panning} assumes that the joint feature distribution follows a multivariate Gaussian distribution and satisfies the pairwise exchangeability condition \eqref{eq:total_loss} via approximating only the first two moments (mean and covariance). In cases where the distributions are not multivariate Gaussian, the second order method in \cite{candes2016panning} cannot guarantee any control of the FDR. In contrast, methods like knockoffGAN \cite{salimans2016improved}, deep knockoff \cite{romano2020deep}, auto-encoding knockoff \cite{liu2018auto} focus on learning generative models to sample knockoffs. KnockoffGAN is a complex architecture which consists of four different neural networks and optimizes a difficult minimax problem. A comparatively simpler approach adopted by deep knockoff employs MMD \cite{gretton2012kernel} as the discriminating statistic for testing pairwise exchangeability in \eqref{eq:total_loss}. As we see in Section \ref{supp:srmmd_w.r.t.epsilon} the generator learned using the MMD performs poorly in high dimensions, and fails to approximate the input distribution properly. Auto-encoding knockoff uses a variational autoencoder \cite{kingma2019introduction} that learns a low-dimensional latent space for the high-dimensional data. The performance of the auto-encoding knockoff depends on the latent space dimension. A higher dimensional latent space
can be used to make the model better but may lead to diminished power if the covariates violate the low-dimensional approximation. Another method, called deep direct likelihood knockoff (DDLK) \cite{sudarshan2020deep} minimizes the KL (Kullback-Liebler) divergence to test for pairwise exchangeability. 
\subsection{Schematic of the Knockoff Generator}\label{supp:knockoff_generation}
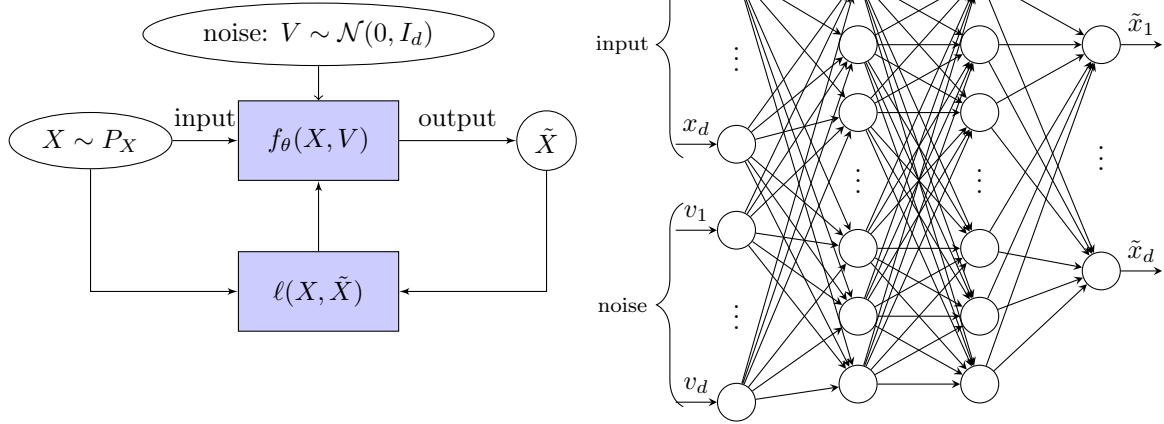
\begin{figure}[H]
\begin{minipage}{.5\textwidth}
\begin{tikzpicture}[auto, node distance=2cm,>=latex']
    \node [input, name=input] {};
    \node [sum, right of = input,
            node distance=3cm] (system) {$ X\sim P_X$};
    \node [block, right of = system, pin={[pinstyle]above:noise: $ V\sim \mathcal N( 0, I_d)$},
            node distance=3cm] (system2) {$f_{\theta}( X,  V)$};
    \node [sum, right of = system2,
            node distance=3cm] (system3) {$ \tilde X$};
    \node [block, below of=system2] (measurements) {$\ell( X, \tilde X)$};

    \draw [draw,->] (system) -- node {input} (system2);
    \draw [->] (system2) -- node {output} (system3);
    \draw [->] (measurements) -- node {} (system2);
    \draw [->] (system) |- node {} (measurements);
    \draw [->] (system3) |- node {} (measurements);
\end{tikzpicture}
\end{minipage}
\hspace{-8mm}
\begin{minipage}{.8\textwidth}

  \begin{tikzpicture}[x=0.8cm, y=0.6cm, >=stealth]

\foreach \m/\l [count=\y] in {1,missing,2, 3,missing, 4}
  \node [every neuron/.try, neuron \m/.try] (input-\m) at (0,2.5-\y*1.9) {};

\foreach \m [count=\y] in {1,2,3, missing, 4, 5,6}
  \node [every neuron/.try, neuron \m/.try ] (hidden1-\m) at (2,2-\y*1.5) {};
  
\foreach \m [count=\y] in {1,2,3, missing, 4, 5,6}
  \node [every neuron/.try, neuron \m/.try ] (hidden2-\m) at (4,2-\y*1.5) {};

\foreach \m [count=\y] in {1,missing, 2}
  \node [every neuron/.try, neuron \m/.try ] (output-\m) at (6,1.5-\y*2.5) {};

\foreach \l [count=\i] in {x_1,x_d,v_1, v_d}
  \draw [<-] (input-\i) -- ++(-1,0)
    node [above, midway] {$\l$};
\foreach \l [count=\i] in {1,d}
  \draw [->] (output-\i) -- ++(1,0)
    node [above, midway] {$\tilde{x}_\l$};

\foreach \i in {1,...,4}
  \foreach \j in {1,2,3,4, 5, 6}
    \draw [->] (input-\i) -- (hidden1-\j);

\foreach \i in {1,2,3,4, 5, 6}
  \foreach \j in {1,2,3,4, 5, 6}
    \draw [->] (hidden1-\i) -- (hidden2-\j);
    
\foreach \i in {1,2,3,4, 5, 6}
  \foreach \j in {1,...,2}
    \draw [->] (hidden2-\i) -- (output-\j);
    
\draw [decorate,decoration={brace,amplitude=10pt},xshift=-4pt,yshift=0pt]
(-0.7,-3.5) -- (-0.7,1.5) node [black,midway,xshift=-0.8cm] 
{\footnotesize input};

\draw [decorate,decoration={brace,amplitude=10pt},xshift=-4pt,yshift=0pt]
(-0.7,-9) -- (-0.7,-4.5) node [black,midway,xshift=-0.8cm] 
{\footnotesize noise};
\foreach \l [count=\x from 0] in {Input, H_1, H_2, Output}
  \node [align=center, above] at (\x*2,1) {$\l$};
\end{tikzpicture}
\end{minipage}
\caption{\emph{Left:} schematic of the deep generative model for knockoff generation. \emph{Right:} schematic of $f_\theta$ is shown for 2 hidden layers (for all experiments we use $6$ hidden layers).} 
\label{fig:architecture}
\end{figure}

 Figure \ref{fig:architecture} shows the schematic of the deep generative model used for the knockoff generator (Section \ref{sec:architecture}). The model has a fully connected neural network $f_\theta$, where $\theta$ represents the parameters of the network (weights $w$ and biases $b$). $f_\theta$ has $6$ has hidden layers each of them having $6\cdot d$ units. The first layer of the neural network takes a vector of original variables $X\in \mathbb R^d$ and a $d$-dimensional noise vector $V\sim \mathcal N(0, I)$. Each unit in the hidden layers is produced by first taking linear combinations of the input followed by applying to each a nonlinear activation function. Parametric rectified linear unit is used as the activation function \cite{xu2015empirical}. The output layer returns a $d$-dimensional knockoff vector as depicted in Figure \ref{fig:architecture}.
\subsection{Algorithm to Train Knockoff Generator}\label{alg:alg1}
\begin{algorithm}
    \SetKwFunction{isOddNumber}{isOddNumber}
    \SetKwInOut{KwIn}{Input}
    \SetKwInOut{KwOut}{Output}
  \SetKwFunction{FMain}{Main}
    \KwIn{training data: $\mathbf X\in \mathbb R^{n\times d}$, learning rate: $\eta$, entropic regularizer: $\varepsilon$, decorrelation parameter: $\gamma$, $\theta_0$: initialization of network parameters, no of epochs: T, batch size: $m$, no of batches: $n_b$}
    \KwOut{knockoff generator: $f_{\theta_T}$}
    \For{$t \leftarrow 0$ \KwTo T}{
        \For{$j \leftarrow 0$ \KwTo $n_b$}{
        $X_i, \;\;\text{for all}\;\; 1 \leq i\leq m$\tcp*[f]{samples for a minibatch}\\
        $V_i\sim \mathcal N(0, I)\;\;\text{for all}\;\; 1 \leq i\leq m$ \tcp*[f]{noise sampling.}\\
            $\tilde X_i \leftarrow f_{\theta_t}(X_i, V_i), \;\;\text{for all}\;\; 1 \leq i\leq m$    \tcp*[f]{knockoff generation.}\\
            $B\subset \{1, \dots, d\}$ \tcp*[f]{picking a random subset.}\\
            $J_{\theta_t} (\mathbf X_{m}, \mathbf{\tilde X}_m)\leftarrow  \ell(\mathbf X_m, \mathbf {\tilde X}_m)$ \tcp*[f]{loss calculation using \eqref{eq:total_loss}}\\
            $\theta_{t+1} = \theta_t - \eta \nabla_{\theta_t}( J_{\theta_t})$\tcp*[f]{parameter update}
        }
    }
    \caption{Training of the Knockoff Generator}
\end{algorithm}
\subsection{Compute Resources}
We use NVIDIA TESLA-K80 24 GB GPU for each simulation. On average, training time of the proposed knockoff generator is around 2 hours for each distributional setting and for the MNIST image generator, it is about 45 minutes.
\subsection{Baseline Models for Knockoff Generation}
For KnockoffGAN, we use the code from \url{https://bitbucket.org/mvdschaar/mlforhealthlabpub/src/master/}. For DDLK, code is taken from a publicly available repository \url{https://github.com/rajesh-lab/ddlk}. For other two benchmarks, second-order and MMD knockoffs, we use the code from this repository \url{https://github.com/msesia/deepknockoffs}. 

\section{Additional Experiments}
\subsection{Effect of \texorpdfstring{$\varepsilon$}{epsilon}, \texorpdfstring{$\sigma$}{sigma} on sRMMD-Based MNIST Image Generator (Section \ref{sec:mnist})}\label{supp:mnist}

\begin{figure}[H] 
  \centering
\subfloat[][]{\includegraphics[width=.28\linewidth, height = 5.5cm]{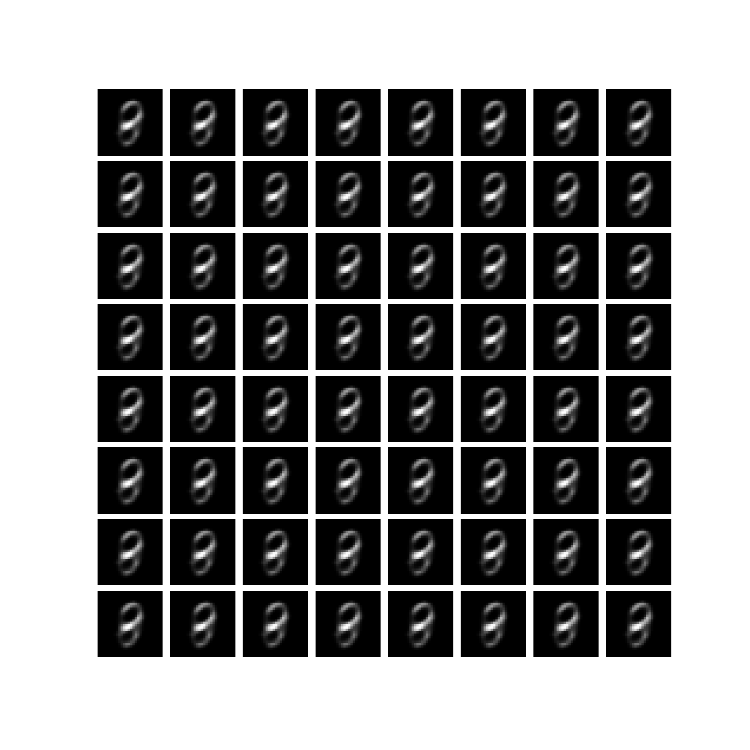}\vspace{-6mm}}\hspace{-8mm}
\subfloat[][ ]{\includegraphics[width=.28\linewidth, height = 5.5cm]{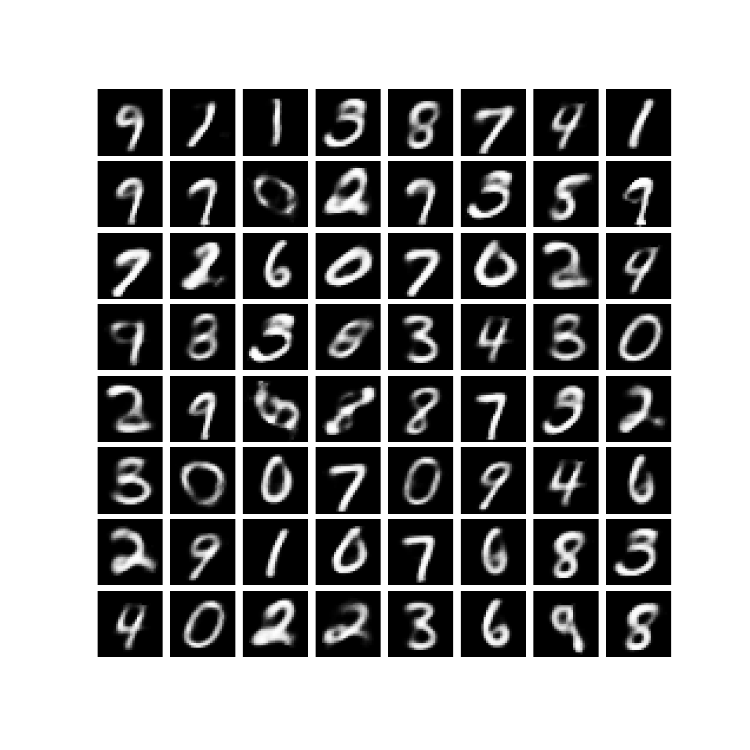}\vspace{-6mm}}\hspace{-8mm}
\subfloat[][]{\includegraphics[width=.28\linewidth, height = 5.5cm]{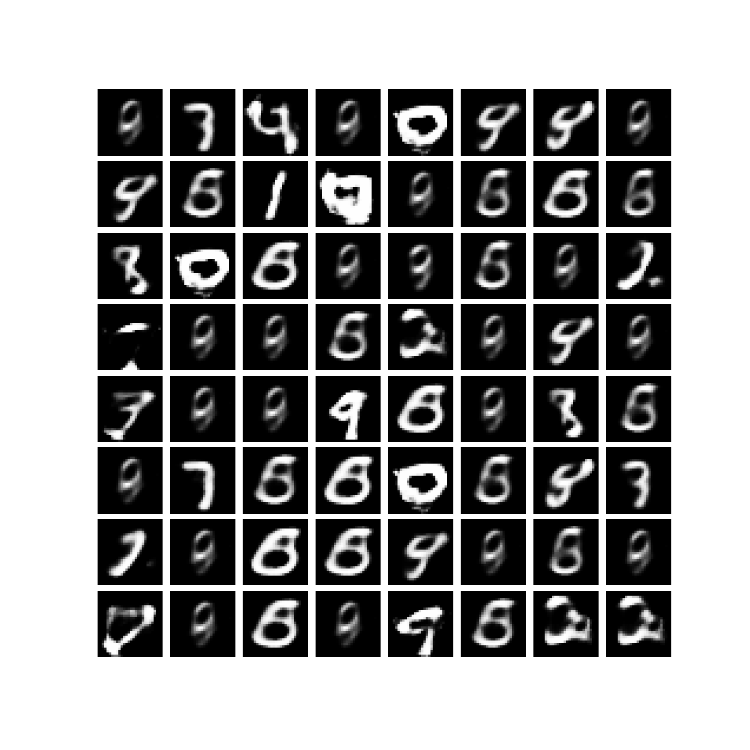}\vspace{-6mm}}\hspace{-8mm}
\subfloat[][]{\includegraphics[width=.28\linewidth, height = 5.5cm]{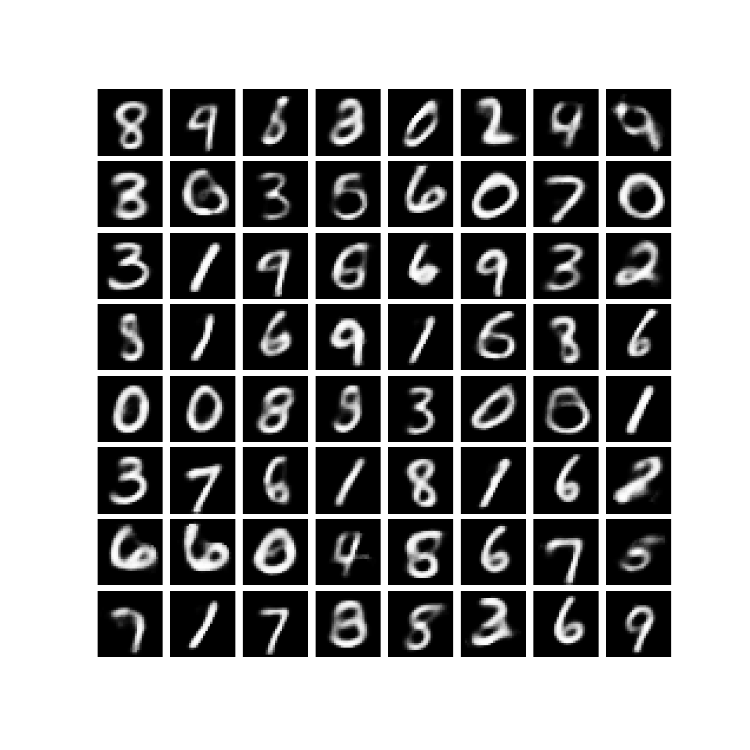}\vspace{-6mm}}
  \caption{sRMMD-based MNIST image generator (latent space dimension 8) using (a) $\varepsilon= 10,\sigma =(1,2,4,8,16,32)$. (b) $\varepsilon= 10, \sigma =(0.01, 0.02, 0.04, 0.06, 0.08)$ (c) $\varepsilon= 20, \sigma =(0.01, 0.02, 0.04, 0.06, 0.08)$, and (d) $\varepsilon= 20, \sigma =(0.001, 0.002, 0.004, 0.006, 0.008)$.}
  \label{fig:mnist_result_supp}
\end{figure}
\begin{figure}[H] 
  \centering
\subfloat{\includegraphics[width=.45\linewidth,height = 5.8cm]{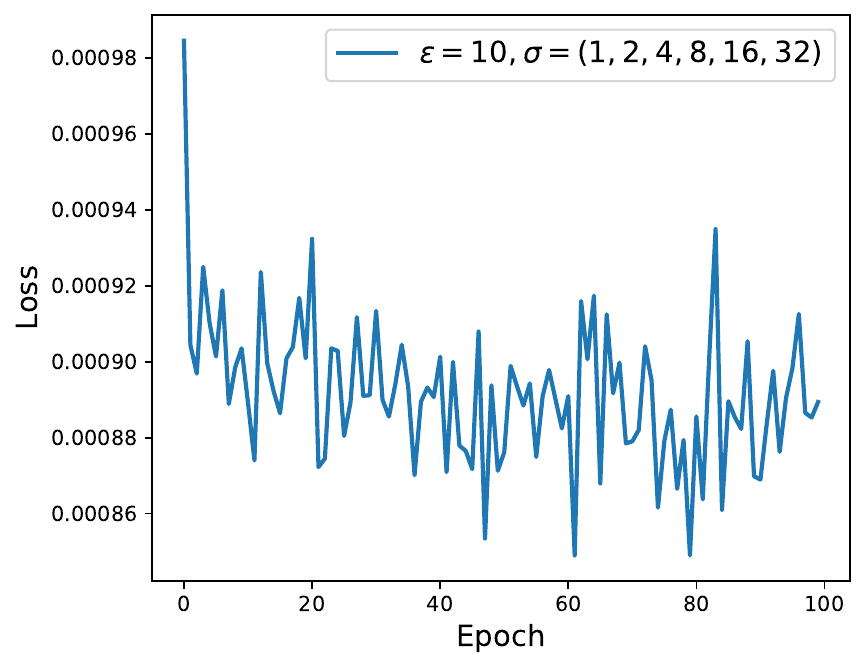}\vspace{-8mm}}\hspace{8mm}
\subfloat{\includegraphics[width=.45\linewidth, height = 5.8cm]{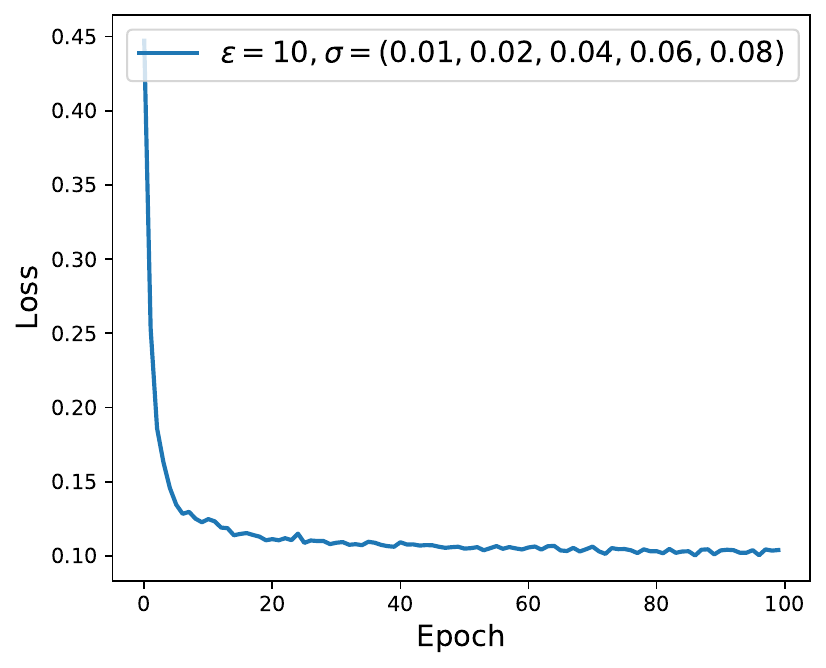}\vspace{-8mm}}\hspace{8mm}
  \caption{In these plots, the $y$-axis is the generator's loss (sRMMD) and the $x$-axis is the training epoch for $\varepsilon =10$ with two different $\sigma$. Left: with large $\sigma$, the loss fluctuates and model does not converge. Right: for small $\sigma$ loss gradually decreases with the epochs and model converges faster.}
  \label{fig:loss_mnist_result_supp}
  \vspace{-4mm}
\end{figure}

MNIST-image generator minimizing the sRMMD loss produces a lot of ambiguous digits of similar shape when $\varepsilon = 10$ and $\sigma =(1, 2, 4, 8, 16, 32)$ are used. To understand why this may be the case, we plot the generator's loss over the training epochs (Figure \ref{fig:loss_mnist_result_supp}). We observe that the loss is nearly zero at the beginning of the training, does not decrease smoothly and becomes very unstable after few epochs. This instability consequently leads to a poor trained generator. In contrast, for $\varepsilon=10$, using $\sigma = (0.01, 0.02, 0.04, 0.06, 0.08)$ maintains a smooth, and decreasing loss over the epochs and the generator converges rapidly which brings about an improved image generator (Figure \ref{fig:mnist_result_supp}(b)).
We also observe that if we increase $\varepsilon$ further e.g., $\varepsilon=20$, using $\sigma =(0.01, 0.02, 0.04, 0.06, 0.08)$ still leads to a poor generator (Figure \ref{fig:mnist_result_supp}c). This situation can be avoided by reducing $\sigma$ to $(0.001, 0.002, 0.004, 0.006, 0.008)$. Based on these empirical evidences, it can be said that using smaller bandwidth in case of larger $\varepsilon$ leads to a better generator. Note that the choices of $\varepsilon$ and $\sigma$ we mentioned are specific to our case where we use latent space with dimension $8$. It may require to find the optimal combination of $\varepsilon$ and $\sigma$ to get a good generator in case a different latent space dimension is used.

\subsection{Effect of \texorpdfstring{$\varepsilon$}{epsilon} on sRMMD-Based Knockoff Generator (Section \ref{sec:architecture})}
\label{supp:srmmd_w.r.t.epsilon}
To examine the impact of the entropic regularizer $\varepsilon$ on the knockoff generator minimizing sRMMD, we consider a Gaussian mixture model with $4$ different modes, $P_X= \sum_{k=1}^4 \tau_k \mathcal N(\xi_k, \Sigma_k)\in \mathbb R^d, d=100$. The covariance matrices are $\Sigma_k = \rho_k^{|i-j|}$ with the correlation coefficients $(\rho_1, \rho_2, \rho_3, \rho_4)= (0.6, 0.4, 0.2, 0.1)$. Mean tuple $(\xi_1, \xi_2, \xi_3, \xi_4)$ and the proportion tuple  $(\tau_1, \tau_2, \tau_3, \tau_4)$ are set to $(0, 20, 40, 60)$, and $(0.27, 0.23, 0.23, 0.27)$, respectively. The generator is trained with different values of $\varepsilon$. The training is done on $n=2000$ samples according to Algorithm \ref{alg:alg1} and we keep other training parameters same for each $\varepsilon$.
\begin{figure}[H]
    \centering
    \includegraphics[width=\linewidth]{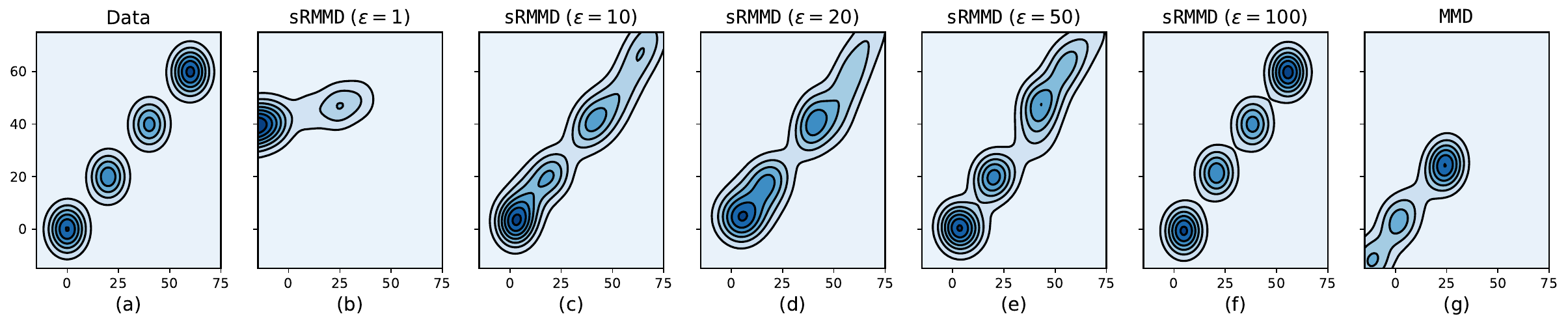}\\
    \caption{Visualizing two randomly selected dimensions of the original data and the generated knockoffs. The rightmost plot shows the reconstructed distribution when using MMD in \eqref{eq:total_loss} instead of sRMMD.}
    \label{fig:epsilon_check}
    \vspace{-4mm}
\end{figure}
From Figure \ref{fig:epsilon_check}, it is evident that when $\varepsilon \ll d$, the knockoff generator fails to reconstruct the data distribution. As we increase $\varepsilon$, the generator improves and perfectly reconstructs the original distribution when $\varepsilon = 100$. On the other hand, MMD-based generator fails to capture all the modes. This indicates that for an appropriate choice of entropic regularizer $\varepsilon$, sRMMD is a better loss function to minimize compared to MMD. 
\subsection{Why sRMMD not sRE?}\label{supp:srmmd_over_sre}
\begin{figure}[H]
    \centering
    \includegraphics[width =\linewidth]{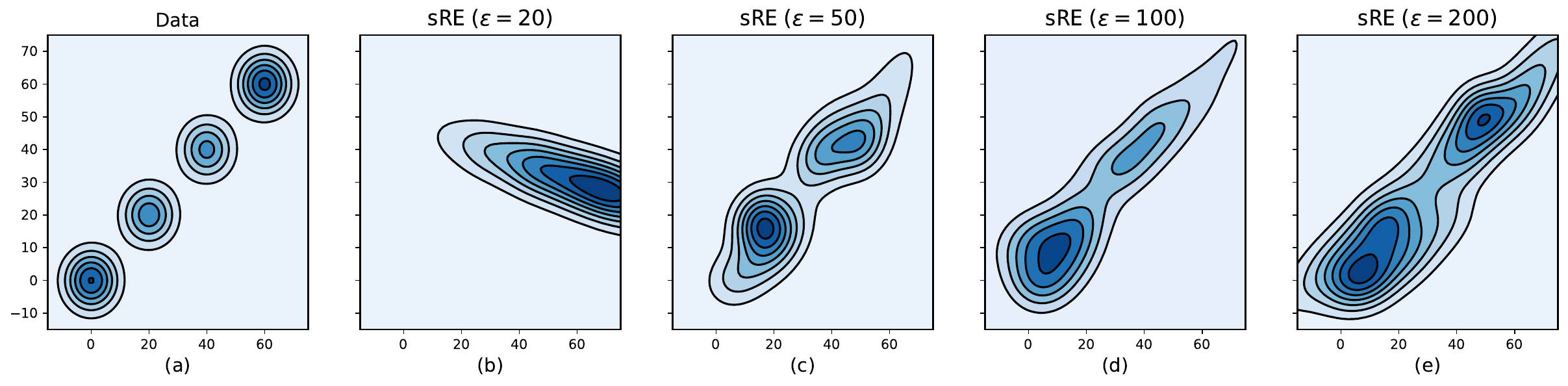}
    \caption{Visualizing two randomly selected dimensions of the original data used in Section \ref{supp:srmmd_w.r.t.epsilon} and the generated knockoffs. sRE-based knockoff generator did not converge when using $\varepsilon =1 $ and $10$ (we encountered `nan' after few epochs). That is why we refrain from adding it here.}
    \label{fig:epsilon_check_sRE}
\end{figure}

Figure \ref{fig:epsilon_check_sRE} shows that unlike sRMMD, the generator minimizing sRE in \eqref{eq:total_loss} cannot capture all four modes perfectly with $\varepsilon =100$ when $d = 100$. Though similar to sRMMD, the reconstruction ability gets better with $\varepsilon$, sRE still fails to capture every mode even when $\varepsilon$ is doubled. Moreover, a direct comparison between sRE and sRMMD when $\varepsilon = 20, 50, 200$ indicates that sRMMD outperforms sRE in reconstructing the original distribution. 
\subsection{Impact of Choice of Decorrelation Parameter \texorpdfstring{$\gamma$}{gamma}}
\begin{figure}[H]
\centering
  \subfloat[][Multivariate Gaussian AR1]{\includegraphics[width=.48\linewidth]{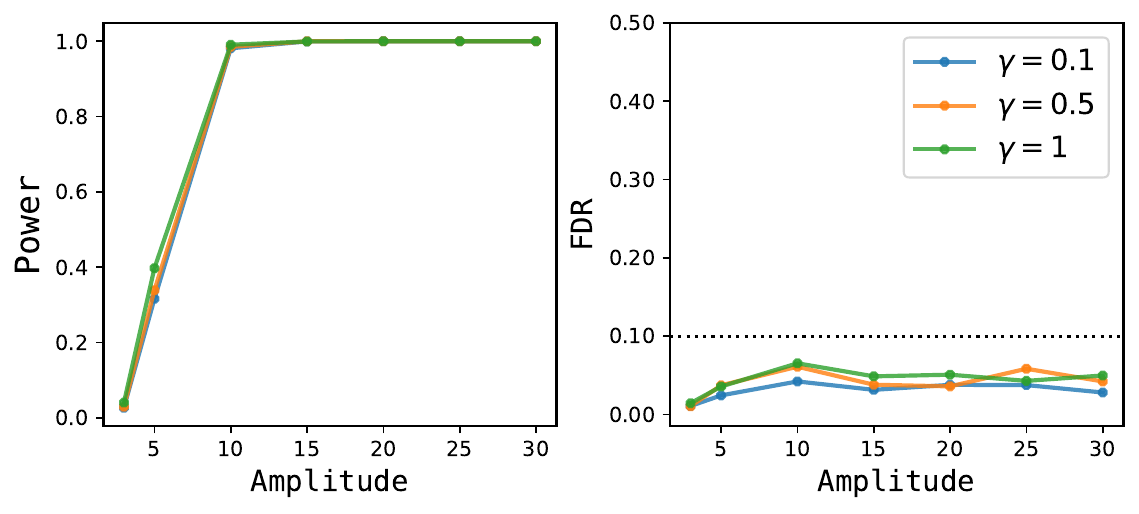}}\quad
  \subfloat[][Gaussian Mixture Model]{\includegraphics[width=.48\linewidth]{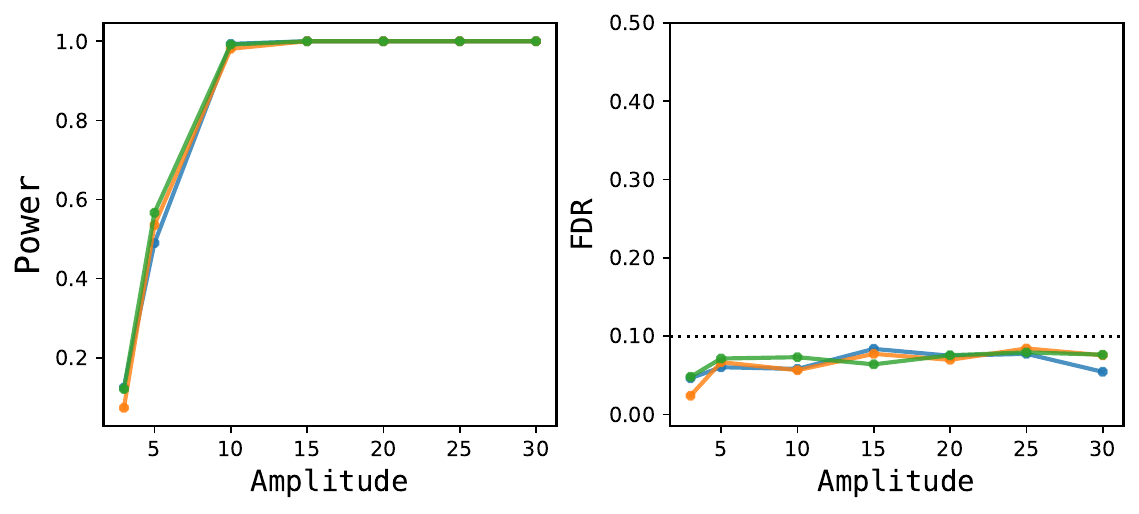}}\\
    \subfloat[][Multivariate Student's t]{\includegraphics[width=.48\linewidth]{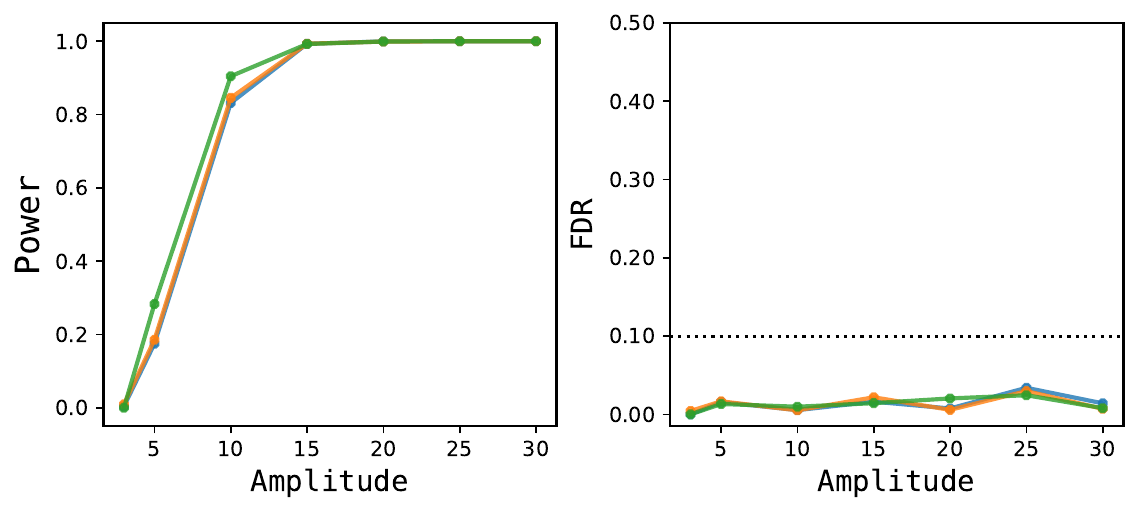}}\quad
  \subfloat[][Sparse Gaussian]{\includegraphics[width=.48\linewidth]{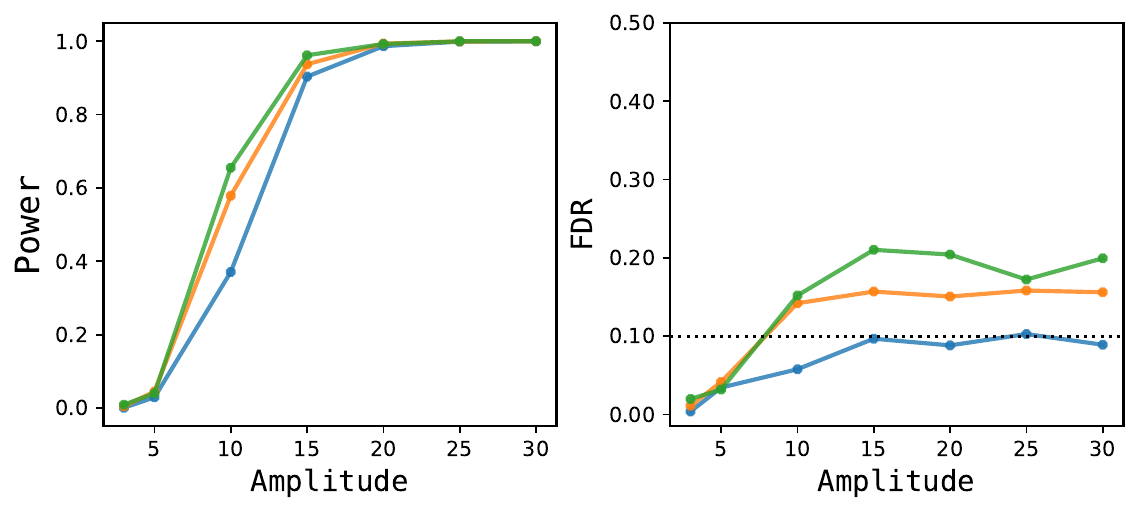}}
  \caption{Average FDR and power computed over 200 independent experiments are shown on the $y$-axis  for each synthetic benchmark. The FDR level is set to 0.1.  The $x$-axis represents the amplitude parameter $\upsilon$.}
    \label{fig:fdr_decor}
    \vspace{-4mm}
\end{figure}

The selection of the optimal decorrelation parameter $\gamma$ is difficult as we cannot perform cross-validation due to lack of access to the ground truth. Therefore we pick the optimal $\gamma$ for each distributional setting by investigating the sensitivity of the results to different values of $\gamma$. Figure \ref{fig:fdr_decor} shows the FDR versus power tradeoff w.r.t. the amplitude parameter for several values of $\gamma$ for each distributional setting considered in Section \ref{subsec:synthetic_setting}.
For multivariate Gaussian, GMM and multivariate Student's t distributions, $\srmmd$-based knockoff generator is not sensitive to the value of $\gamma$. In these cases, each $\gamma$, $\srmmd$ controls the FDR at $q=0.1$ and achieves nearly identical power over the entire amplitude region. In case of sparse Gaussian setting, $\gamma= 0.1$ controls the FDR at $q=0.1$. As $\gamma$ increases e.g., $0.5, 1$, the power also increases but fails to keep the FDR below $0.1$. 

\subsection{Quality of the sRMMD Knockoffs  w.r.t. \texorpdfstring{$\varepsilon$}{epsilon}}
\begin{figure}[H]
\centering
  \subfloat[][Multivariate Gaussian AR1]{\includegraphics[width=.48\linewidth]{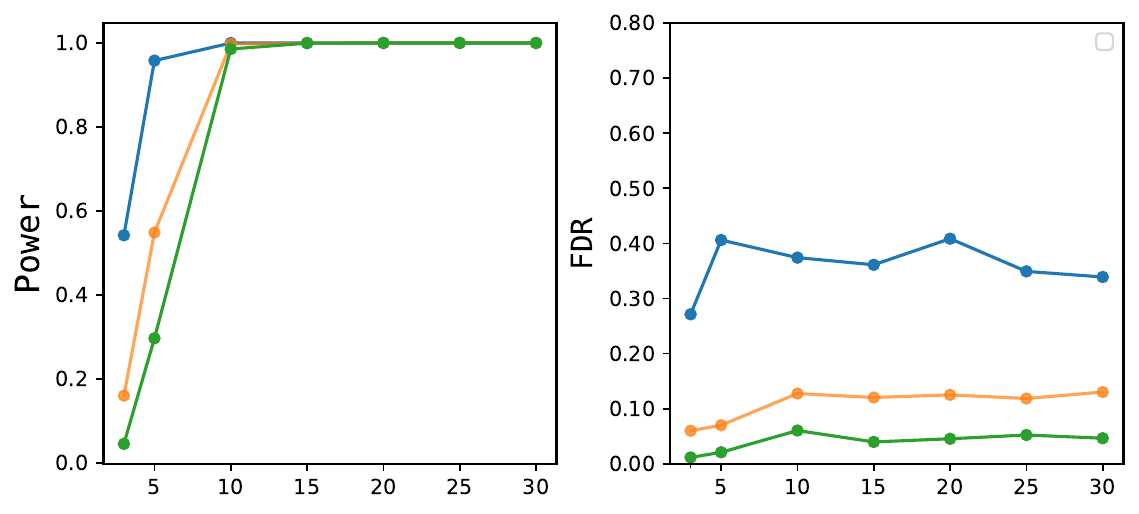}}\quad
  \subfloat[][Gaussian Mixture Model]{\includegraphics[width=.48\linewidth]{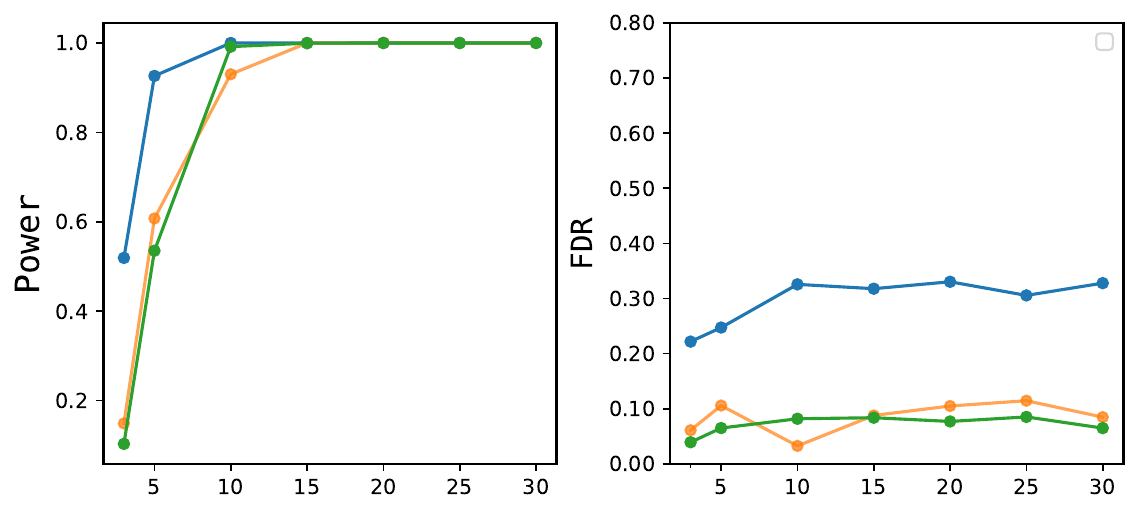}}\\
    \subfloat[][Multivariate Student's t]{\includegraphics[width=.48\linewidth]{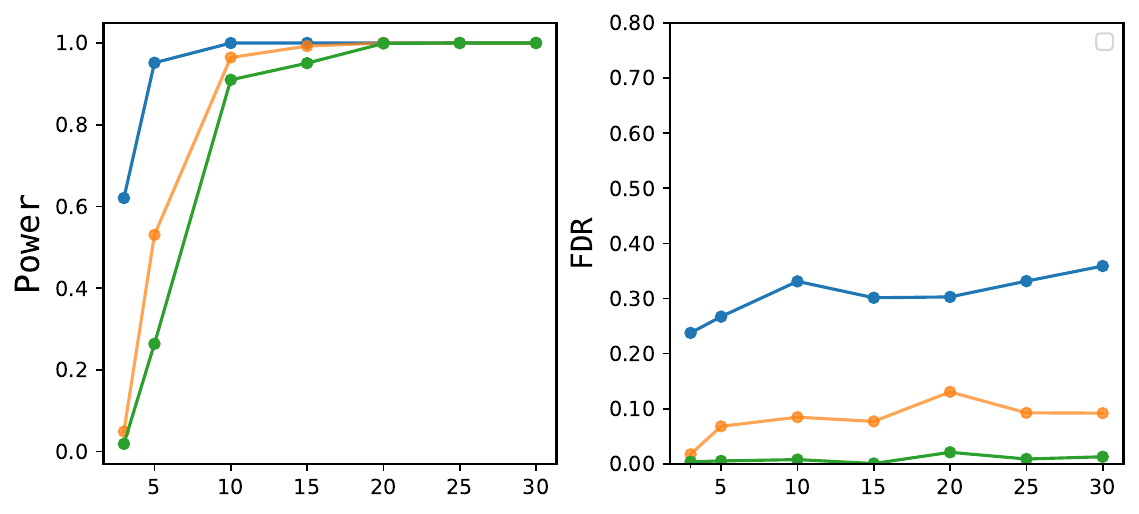}}\quad
  \subfloat[][Sparse Gaussian]{\includegraphics[width=.48\linewidth]{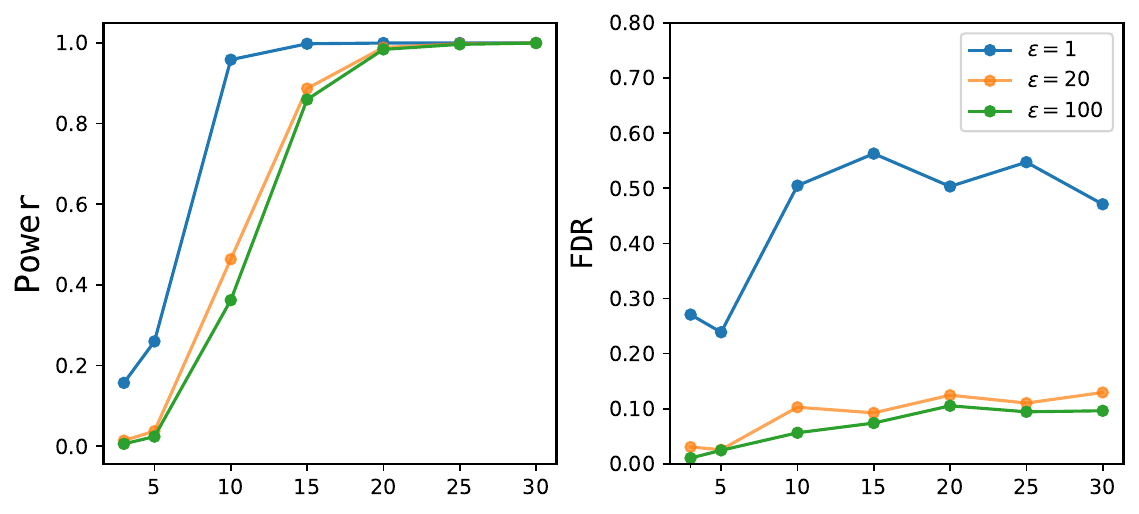}}
  \caption{Average FDR and power computed over 200 independent experiments are shown on the $y$-axis  for each synthetic benchmark. The FDR level is set to 0.1.  The $x$-axis represents the amplitude parameter $\upsilon$.}
    \label{fig:fdr_epsilon}
    \vspace{-6mm}
\end{figure}

In this section, we measure the GoF of knockoffs for each distributional setting considered in Section \ref{subsec:synthetic_setting} w.r.t. different entropic regularization parameters $\varepsilon$. For a fixed covariate distribution $P_X$ and corresponding conditional distribution $P_{\tilde X|X}$ of the knockoffs, we test the following hypothesis: 
\begin{equation*}
H_0^{\text{parital}}: P_{(X, \tilde X)} = P_{(X, \tilde X)_{\text{parital}(B)}},
\end{equation*}
where $B$ is a random subset of {\small $\{1, \dots, d\}$}, such that each $j\in B$ with probability $1/2$ independent of other elements.  We draw $n$ independent observations from $ P_{(X, \tilde X)}$ and $P_{(X, \tilde X)_{\text{parital}(B)}}$ and generate two matrices $\mathbf Z,\mathbf Z' \in \mathbf R^{n\times 2d}$, $n=200, d=100$, respectively. Then, we compute an estimate of MMD to measure the GoF via: 
\begin{align*}
    \mathtt{MMD}(\mathbf Z, \mathbf Z') \triangleq  \frac{1}{n(n-1)}\sum_{i, j = 1}^{n} k(Z_i, Z_j)- \frac{2}{n^2}\sum_{i, j = 1}^{n, n} k( Z_i, Z'_j) + \frac{1}{n(n-1)}\sum_{i, j =1}^{n} k( Z'_i, Z'_j),
    \end{align*}
where $k(\cdot, \cdot)$ is a Gaussian mixture kernel with bandwidth parameter $\sigma =(1, 2, 4, 8, 16, 32, 64, 128)$. A small value of MMD indicates that the knockoffs achieve greater pairwise exchangeability \ref{eq:exchangeability} compared to the knockoffs that yield higher values of MMD.
\begin{figure}[H]
   \centering
  \subfloat[][{\tiny {\bf Multivariate Gaussian}}]{\includegraphics[width=.23\linewidth]{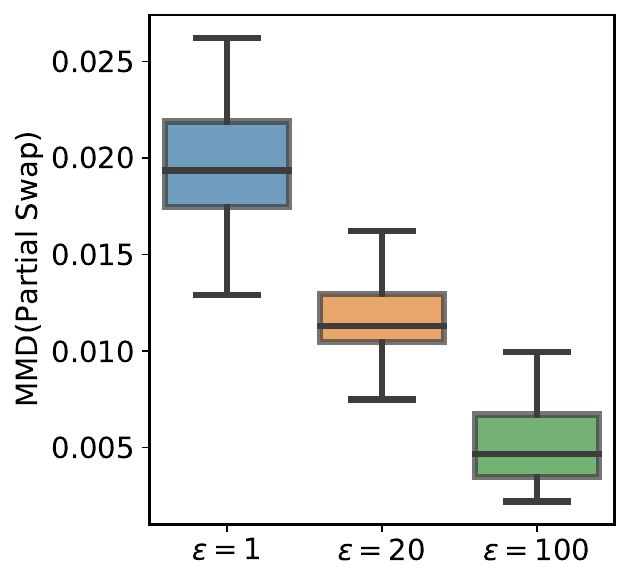}}
  \subfloat[][{\tiny {\bf Gaussian Mixture Model}}]{\includegraphics[width=.23\linewidth]{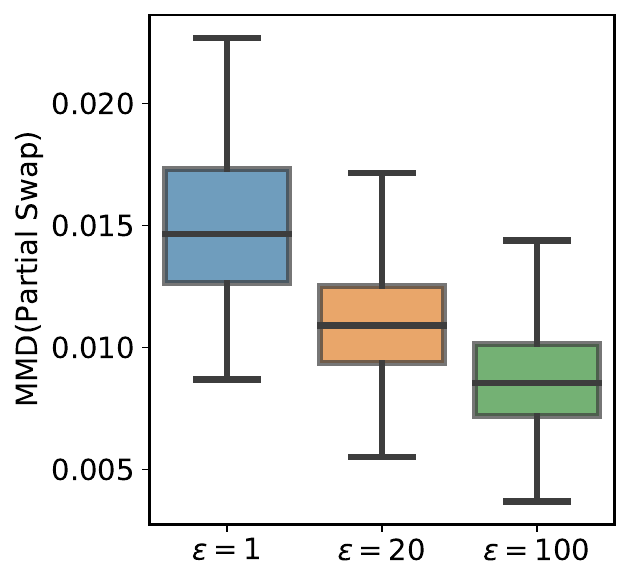}}
   \subfloat[][{\tiny {\bf Multivariate Student's t}}]{\includegraphics[width=.23\linewidth]{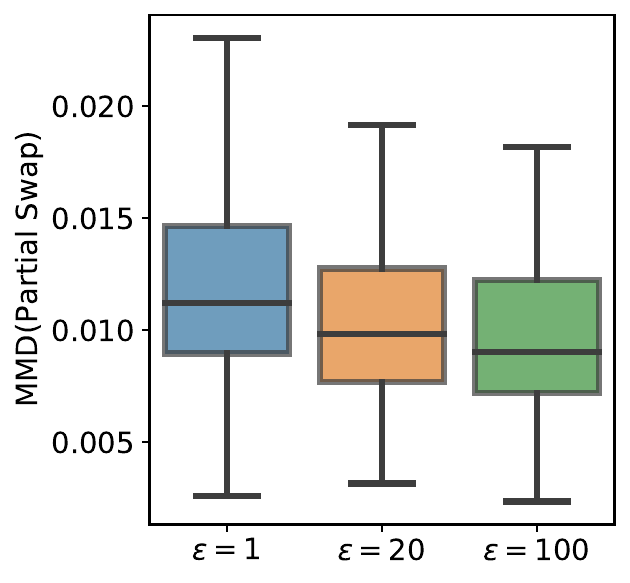}}
   \subfloat[][{\tiny{\bf Sparse Gaussian}}]{\includegraphics[width=.23\linewidth]{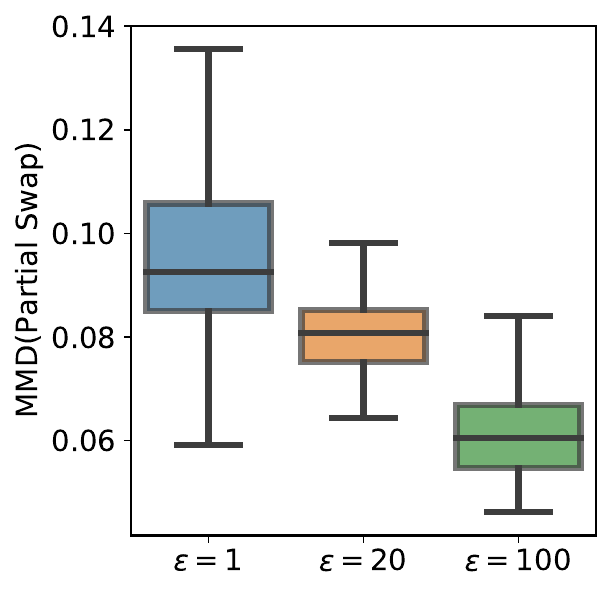}}
   \caption{Boxplot showing the quality of the knockoff generator w.r.t.  $\varepsilon$. Lower MMD indicates better quality knockoffs compared to higher value of MMD.}
   \label{fig:quality_epsilon}
   \vspace{-4mm}
\end{figure}
Figure \ref{fig:quality_epsilon} shows the quality of the knockoffs produced by a generator trained using different $\varepsilon$. For each distributional setting, we observe that knockoffs generated using larger entropic regularizer, (e.g., $\varepsilon = 100$) have better quality compared to using the smaller values (e.g., $\varepsilon= 1, 20$). As a result, we observe that in most cases using larger $\varepsilon$ leads to better FDR control (Figure \ref{fig:fdr_epsilon}). In addition, using larger $\varepsilon$ reduces the computational complexity to a great extent and helps the model  to converge faster.
\end{document}